\documentclass[twoside,11pt]{article}

\usepackage[abbrvbib, preprint, nohyperref]{jmlr2e}

\usepackage{xcolor}

\usepackage{amsmath}
\usepackage{physics}
\usepackage{algorithm}
\usepackage{algpseudocode}

\usepackage{multirow}
\usepackage{wrapfig}
\usepackage{booktabs}
\usepackage{hyperref}
\usepackage[nameinlink, noabbrev, capitalize]{cleveref}
\usepackage{xspace}
\usepackage{tikz}
\usetikzlibrary{shapes.geometric, arrows}
\usepackage{mathtools}
\usepackage{annotate-equations}
\usepackage{caption}
\usepackage{cleveref}

\makeatletter
\renewcommand{\ALG@name}{CP Framework: Experiment Structure}
\makeatother
\usepackage{subcaption}

\usepackage{lastpage}

\ShortHeadings{Uncertainty Quantification of Surrogate Models using Conformal Prediction}{Gopakumar et al.}
\firstpageno{1}

\begin{document}

\title{Uncertainty Quantification of Surrogate Models \\ using Conformal Prediction}

\author{\name Vignesh Gopakumar\thefootnote{*}\email v.gopakumar@ucl.ac.uk \\
       \addr Centre for Artificial Intelligence \\ Department of Computer Science\\
       University College London\\
       London, WC1V 6LJ, UK
       \AND
       \name Ander Gray\thefootnote{*} \email ander.gray@hds.utc.fr \\
       \addr Heudiasyc Laboratory\\
        Universit\'e de Technologie de Compi\`egne\\
        Compi\`egne, 60200, France
       \AND
       \name Joel Oskarsson \email joel.oskarsson@outlook.com \\
       \addr Department of Computer and Information Science\\
       Linköping University\\
       Linköping, 581 83, Sweden
       \AND
       \name Lorenzo Zanisi \email lorenzo.zanisi@ukaea.uk \\
       \addr Computing Division\\
       UK Atomic Energy Authority\\
       Oxford, OX14 3EB, UK
       \AND
       \name Daniel Giles \email d.giles@ucl.ac.uk \\
       \addr Centre for Artificial Intelligence\\ Department of Computer Science\\
       University College London \\
       London, WC1V 6LJ, UK
       \AND
       \name Matt J. Kusner \email m.kusner@ucl.ac.uk \\
       \addr Centre for Artificial Intelligence\\ Department of Computer Science\\
       University College London\\
       London, WC1V 6LJ, UK   
       \AND
       \name Stanislas Pamela \email stanislas.pamela@ukaea.uk \\
       \addr Computing Division\\
       UK Atomic Energy Authority\\
       Oxford, OX14 3EB, UK
       \AND
       \name Marc Peter Deisenroth \email m.deisenroth@ucl.ac.uk \\
       \addr Centre for Artificial Intelligence\\ Department of Computer Science\\
       University College London\\
       London, WC1V 6LJ, UK
       }
       
\def\thefootnote{*}\footnotetext{These authors contributed equally to this work}\def\thefootnote{\arabic{footnote}}


\maketitle

\newpage
\begin{abstract}
Data-driven surrogate models offer fast, inexpensive approximations to complex numerical and experimental systems but typically lack uncertainty quantification, limiting their reliability in safety-critical applications. While Bayesian methods provide uncertainty estimates, they offer no statistical guarantees and struggle with high-dimensional spatio-temporal problems due to computational costs and dependence on prior specification. We present a conformal prediction (CP) framework that provides statistically guaranteed marginal coverage for surrogate models in a model-agnostic manner with near-zero computational cost. Our approach handles high-dimensional spatio-temporal outputs by performing cell-wise calibration while preserving the tensorial structure of predictions. Through extensive empirical evaluation across diverse applications—including partial differential equations, magnetohydrodynamics, weather forecasting, and fusion diagnostics—we demonstrate that CP achieves empirical coverage with valid error bars regardless of model architecture (MLP, U-Net, FNO, ViT, GNN), training regime, or output dimensionality (spanning 32 to over 20 million dimensions). We evaluate three nonconformity scores (conformalised quantile regression, absolute error residual, and standard deviation) for both deterministic and probabilistic models, showing that guaranteed coverage holds even for out-of-distribution predictions where models are deployed on physics regimes different from their training data. Calibration requires only seconds to minutes on standard hardware, with prediction set construction incurring negligible computational overhead. The framework enables rigorous validation of pre-trained surrogate models for downstream applications without retraining, providing actionable uncertainty quantification for decision-making in scientific domains. While CP provides marginal rather than conditional coverage and assumes exchangeability between calibration and test data—limitations we demonstrate empirically through sensitivity analyses—our method circumvents the curse of dimensionality inherent in traditional uncertainty quantification approaches, offering a practical tool for the trustworthy deployment of machine learning in the physical sciences.
\end{abstract}

\begin{keywords}
  Surrogate Models, Uncertainty Quantification, Conformal Prediction, Neural-PDE, Neural-Weather 
\end{keywords}

\section{Introduction}
\label{sec: introduction}

Partial Differential Equations (PDEs) governing physical processes are solved using complex numerical simulation codes. While these codes offer mathematically rigorous solutions, they are limited to discretised domains and require computationally expensive iterative solvers such as finite-volume and finite-element methods. Such simulation codes are central to scientific disciplines in biology \citep{Hospital_Adam2015-bw}, engineering \citep{giudicelli2024moose}, and climate science \citep{cesm2,simintelligence}, but are difficult to deploy for rapid, iterative modelling required while exploring design spaces. Machine learning offers an alternative data-driven route for obtaining quick, inexpensive approximations to numerical simulations \citep{Bertone2019, Karniadakis2021}. Data-driven surrogate models distil spatio-temporal information from simulations into parameterised machine learning models. Due to their efficiency, cost-effectiveness, and relative accuracy, neural networks have become ubiquitous within scientific modelling, with primary importance in tackling large-scale PDEs in climate \citep{lam2022graphcast,pathak2022fourcastnet}, computational fluid dynamics \citep{jiang2020meshfreeflownet,pfaff2021learning}, and nuclear fusion \citep{van_de_Plassche_2020, Gopakumar_2020}. 

However, these surrogate models remain approximations of true physical systems, inheriting multiple layers of uncertainty from both the numerical codes and the PDE formulations themselves. Critically, they often fail to quantify their approximation error relative to the numerical code, producing overconfident outputs regardless of their training domain. This poses two key problems: (a) without confidence assessment, erroneous predictions can lead to severe downstream consequences; (b) high training costs are wasted if predictions lack actionable uncertainty quantification. While several works have attempted uncertainty estimation for surrogate models \citep{GENEVA2020109056, ALHAJERI202234, zou2022neuraluq, Psaros2023}, they fail to provide statistical guarantees over error bars and do not scale to complex scenarios \citep{ABDAR2021243}. Moreover, they require computationally expensive ensemble training \cite{lakshminarayanan2017simple}, extensive sampling \citep{bnns}, or architectural modifications \citep{gal2016dropout}. Validating surrogate model outputs for specific downstream applications remains a pressing challenge.

\textbf{Conformal prediction (CP)}~\citep{vovk2005algorithmic} provides a framework for computing statistically guaranteed error bars over pre-trained and fine-tuned models, i.e. error bars calibrated to provide required coverage. Conformal prediction relies on calibrating model performance across a dataset representative of the desired prediction distribution, then utilising these calibration measures to provide valid error bars for model outputs.

In this paper, we conduct a thorough empirical study demonstrating that conformal prediction provides statistically guaranteed error bars for neural-network-based surrogate models across spatio-temporal domains, even in out-of-training-distribution scenarios. Through experiments of increasing complexity, we show guaranteed marginal coverage irrespective of model choice (deterministic or probabilistic), output dimensionality (up to 20 million dimensions), training data, or physical setting. We explore various conformal prediction methods, comparing their cost, performance, and architectural requirements. Our work provides a rigorous method to assess the usefulness, validity, and applicability of pre-trained and fine-tuned surrogate models for inference and production scenarios.

\subsection*{Applications to Critical Scientific Challenges}
Machine-learning-based surrogate modelling accelerates scientific simulation through computational efficiency and enables data-driven discovery at scale. When modelling complex systems such as computational fluid dynamics, nuclear fusion and weather forecasting, both computational speed and accurate uncertainty estimates are crucial. In safety-critical systems \cite{safetycriticalsystems}, supplementing model predictions with calibrated uncertainty estimations is imperative for improved decision-making in downstream tasks. The CP framework demonstrated here advances uncertainty quantification for complex scientific modelling at scale with industry-level safety-critical applications. Neural-PDE solvers offer quick, inexpensive PDE solutions\citep{Yin_2023}, enabling system understanding and optimal design point identification\citep{li2023geometryinformed, SHUKLAdnoairfoil}. As these models become ubiquitous, verifying prediction accuracy becomes pressingly important. Applications span nuclear fusion for low-carbon energy production \citep{LEREDE2023101144,Degrave2022,PFC,Kates-Harbeck2019,pamela2024neuralpararealdynamicallytrainingneural,carey2024dataefficiencylongterm} to weather forecasting for proactive climate change response \citep{cc_extremeweather,gcm_error_growth,ddw_risk}. Through this work, we propose a model-agnostic method providing calibrated error bounds for all variables, lead times, and spatial locations, requiring no model modifications with negligible computational costs.

\subsection*{Outline}
The paper is structured as follows: \cref{sec: conformal_prediction} introduces the inductive CP framework, spatio-temporal data, associated exchangeability assumptions, and the mathematical extension of CP to spatio-temporal domains. \cref{sec: experiments} presents experiments deploying our CP framework across diverse modelling tasks, from PDEs to climate modelling. \cref{Discussion} discusses the framework's strengths and limitations, and \cref{conclusion} concludes.

\section{Conformal Prediction}
\label{sec: conformal_prediction}

Conformal prediction (CP) \citep{vovk2005algorithmic, shafer2008tutorial} addresses a fundamental question in machine learning: given a dataset $(X_1,Y_1), (X_2,Y_2), \ldots, (X_n, Y_n)$ and a trained model $\hat{f}:\mathcal{X}\to \mathcal{Y}$, how accurate is $\hat{f}$ at predicting the true label $Y_{n+1}$ for a new query point $X_{n+1}$? CP extends point predictions $\hat{y}$ to prediction sets $\mathbb{C}^{\alpha}$ that contain the true label with guaranteed probability:
\begin{equation}
    \label{eq: coverage}
    \mathbb{P}(Y_{n+1}\in \mathbb{C}^{\alpha}) \geq 1 - \alpha.
\end{equation}
This coverage guarantee holds regardless of the model architecture or training procedure, requiring only that calibration samples are exchangeable (a weaker form of i.i.d)\citep{vovk2005algorithmic}.

Several CP variants exist, with inductive conformal prediction \citep{papadopoulos2008inductive} being particularly efficient for neural networks. This approach splits data into a training set (for model training) and a calibration set (for constructing prediction sets $\mathbb{C}^\alpha$). The prediction sets satisfy \cref{eq: coverage} by comparing model outputs to calibration data using a \textit{nonconformity score}, i.e. a metric quantifying prediction error.

\tikzstyle{process} = [rectangle, minimum width=3.0cm, minimum height=1.5cm, text centered, draw=black]
\tikzstyle{arrow} = [thick,->,>=stealth]

\begin{figure}
\centering
\begin{tikzpicture}[remember picture, node distance=5.0cm]
    \node (start) [process, fill=red!30, label={[font=\color{red}]below:Calibration}] {$\hat{s} = |y_c - \tilde{y}_c|$};
    \node (step) [process, fill=blue!30, label={[font=\color{blue}]below:Quantile Estimation}, right of=start] {$\hat{q} = F^{-1}_{\hat{s}}\bigg(\frac{\lceil(n+1)(1-\alpha)\rceil}{n}\bigg)$};
    \node (end) [process, fill=orange!20, label={[font=\color{orange}]below:Prediction Sets}, right of=step] {$\hat{y}_p = \tilde{y}_{p} \pm \hat{q}$};

    \draw [arrow] (start) -- (step);
    \draw [arrow] (step) -- (end);
\end{tikzpicture}
\caption{Inductive CP framework using Absolute Error Residual (AER) nonconformity scores (see \cref{nonconformity scores}): (1) Calibrate by computing nonconformity scores ($\hat{s}$) from calibration predictions ($\tilde{y}_c$) and targets ($y_c$). (2) Estimate the quantile ($\hat{q}$) for desired coverage $(1-\alpha)$ using $n$ calibration samples and the inverse CDF $F_{\hat{s}}^{-1}$. (3) Construct prediction sets by applying $\hat{q}$ to test predictions ($\tilde{y}_p$).}
\label{fig: cp_framework_tikz}
\end{figure}
    
The inductive CP framework operates through three fundamental steps (\cref{fig: cp_framework_tikz}). First, \textit{nonconformity scores} $s(x,y)$ are computed on calibration data, quantifying the disagreement between predictions and ground truth—larger scores indicate worse model performance. Second, the $(1-\alpha)$-quantile $\hat{q}$ of these scores is estimated, providing the threshold for constructing prediction sets. Third, for any test point $x_{\text{test}}$, the prediction set $\mathcal{C}(x_{\text{test}}) = \{y : s(x_{\text{test}}, y) \leq \hat{q}\}$ includes all outputs whose nonconformity scores fall below $\hat{q}$ \citep{gentle_introduction_CP}. Critically, while the coverage guarantee in \cref{eq: coverage} holds universally, regardless of model quality, the \textit{usefulness} of the prediction sets depends entirely on the choice of nonconformity score function. Well-designed scores that accurately rank prediction difficulty yield tight intervals for easy inputs and wider intervals for challenging ones; poorly chosen scores produce uninformative but still valid prediction sets.

\subsection{Conformal Prediction for Spatio-Temporal Data}

While originally developed for single-output predictions \citep{vovk2005algorithmic}, CP has been extended to spatio-temporal domains \citep{sun2022conformal, conformaltimeserires, cp_dynamic_timeseries, CP_Wildfire, ma2024calibrated}. We consider models predicting the evolution of spatio-temporal fields in physical systems as found in PDE modelling, weather forecasting, and fusion diagnostics. Each modelling task is formulated as an initial value problem where calibration and prediction sets consist of input-output pairs characterised by initial conditions and their solutions.

\begin{figure}
    \centering
    \includegraphics[width=0.75\textwidth]{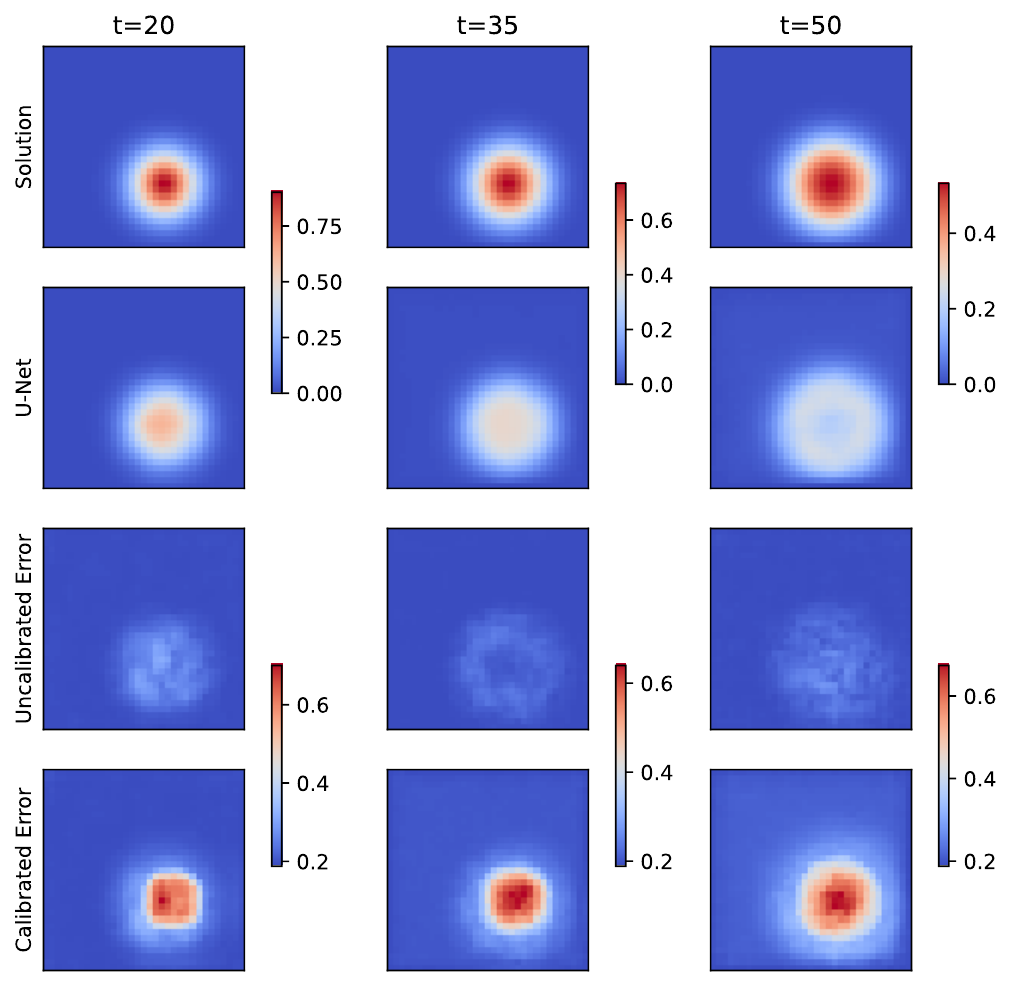}
    \caption{Cell-wise uncertainty calibration using CP for a U-Net modelling the wave equation in an out-of-distribution setting (\cref{sec: wave-unet}). Rows show: ground truth, model prediction, uncalibrated 95\% coverage from MC dropout ($2\sigma$), and calibrated 95\% coverage ($\alpha=0.05$) from CP. Cell-wise calibration provides guaranteed coverage for each spatial location. MC dropout produces unrealistically small uncertainties, while CP error bars correctly identify regions of high uncertainty corresponding to complex dynamics.}
    \label{fig:cell-wise-cp}
\end{figure}

Our framework performs calibration independently for each cell of the spatio-temporal tensor (\cref{fig:cell-wise-cp}), providing marginal coverage guarantees at every spatial and temporal location. This cell-wise approach yields upper and lower bounds for each point without explicitly modelling spatial correlations, instead relying on the neural network to capture spatial dependencies during training. The discretised spatio-temporal grid must remain consistent between calibration and prediction sets. We formalise this approach to provide statistically valid, dimension-independent marginal coverage for high-dimensional outputs, as demonstrated across diverse applications, including multi-physics systems and operational weather models.

\subsubsection{Mathematical Formulation}
\label{subsec: formal_definition}

Consider a model $\hat{f}$ mapping initial temporal sequences of spatial fields $X \in \mathbb{R}^{T_{\text{in}} \times N_x \times N_y \times N_{\text{var}}}$ to future sequences $\tilde{Y} = \hat{f}(X) \in \mathbb{R}^{T_{\text{out}} \times N_x \times N_y \times N_{\text{var}}}$, where $T_{\text{in}}, T_{\text{out}}$ denote input and output time steps, $N_x, N_y$ are spatial dimensions, and $N_{\text{var}}$ is the number of variables. The calibration procedure $\hat{q} = \hat{C}(\tilde{Y}, Y)$ uses model predictions $\tilde{Y}$ and ground truth $Y$ to compute quantiles $\hat{q} \in \mathbb{R}^{T_{\text{out}} \times N_x \times N_y \times N_{\text{var}}}$ in a point-wise manner. These quantiles define lower ($L$) and upper ($U$) bounds forming the prediction set $\mathbb{C}$, with $L$ and $U$ sharing the dimensionality of $\hat{q}$. For a test point $X_{n+1}$ with true label $Y_{n+1}$, the coverage guarantee becomes:
\begin{equation} \label{eq:coverage_tensor}
    \mathbb{E}\bigg[ (Y_{n+1} \geq L) \wedge (Y_{n+1} \leq U) \bigg]\geq 1 - \alpha.
\end{equation}
\Cref{eq:coverage_tensor} holds for each tensor cell given sufficient calibration samples and maintained exchangeability \citep{vovk2012conditional}.

\subsection{Nonconformity Scores}
\label{nonconformity scores}

Nonconformity scores quantify model deviation from ground truth using the calibration set \citep{gentle_introduction_CP}. We employ three methods:

\begin{itemize}
    \item \textbf{Conformalised Quantile Regression (CQR)} \citep{conformalized_quantile_regression}: Train three models to predict the $100 \times \alpha$-th, median, and $100(1-\alpha)$-th percentiles using quantile loss \citep{koenker_2005}. The nonconformity score measures distance to the nearest bound: $s(x,y) = \max\{\underline{f}(x) - y, y - \overline{f}(x)\}$. After computing $\hat{q}$, the prediction set is obtained as $\{\underline{f}(x) - \hat{q}, \overline{f}(x) + \hat{q}\}$.
    
    \item \textbf{Absolute Error Residual (AER)} \citep{error_residual}: Train a single deterministic model using standard loss (e.g., MSE). Compute nonconformity scores as absolute errors: $s(x,y) = |y - \tilde{f}(x)|$. The prediction set becomes $\{\tilde{f}(x) - \hat{q}, \tilde{f}(x) + \hat{q}\}$. This requires no architectural modifications and is computationally efficient.
    
    \item \textbf{Standard Deviation (STD)}: Use probabilistic models outputting mean $\mu(x)$ and standard deviation $\sigma(x)$. The nonconformity score is $s(x,y) = \frac{|y - \mu(x)|}{\sigma(x)}$, yielding prediction sets $\{\mu(x) - \hat{q}\sigma(x), \mu(x) + \hat{q}\sigma(x)\}$. This requires architectural changes (e.g., dropout layers) or modified training (e.g., negative log-likelihood loss). The dependence on the standard deviation of the prediction introduces a weak sense of conditionality. 
\end{itemize}

\begin{figure}[!ht]
    \centering
    \includegraphics[width=0.9\textwidth]{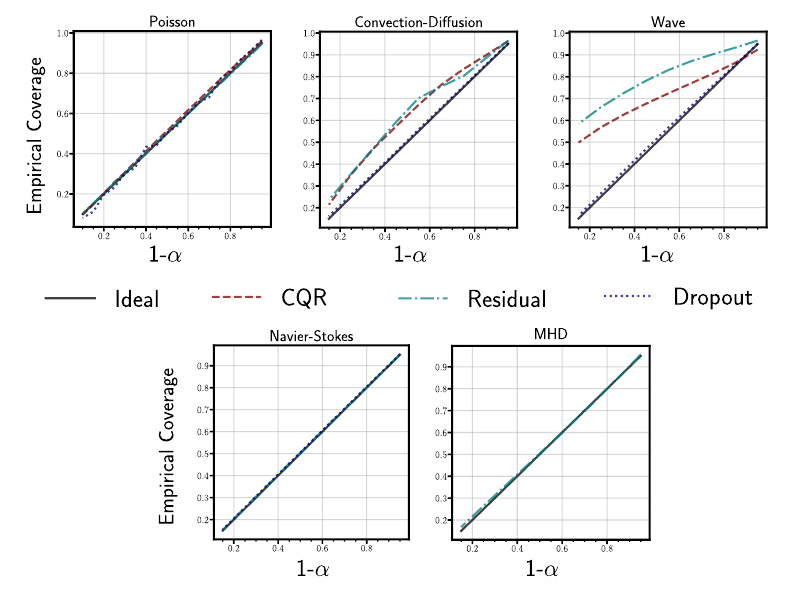}
    \caption{Empirical coverage versus target coverage $(1-\alpha)$ across experiments and nonconformity scores. The diagonal represents ideal coverage. All methods achieve near-perfect coverage across four PDE experiments, validating \cref{eq: coverage}.}
    \label{fig: Validation_Plots}
\end{figure}

\Cref{fig: Validation_Plots} demonstrates that all three nonconformity scores achieve guaranteed coverage across diverse experiments (details in \cref{sec: experiments}). While coverage quality varies slightly between methods, CP ensures validity regardless of the choice. We select nonconformity scores based on practical considerations: architectural constraints, calibration cost, and data availability.

To validate coverage empirically, we compute:
\begin{equation} \label{eq: emp_cov}
    \mathbb{P}(Y_{\text{val}} \in \mathbb{C}^{\alpha}) \approx \frac{1}{n_{\text{val}}} \sum_{i=1}^{n_{\text{val}}} I_{\mathbb{C}^{\alpha}}(Y_{i}),
\end{equation}
where $I_{\mathbb{C}^{\alpha}}$ is the indicator function for the prediction set. Valid coverage requires this to exceed $1 - \alpha$. The empirical coverage obtained for a setting follows a Beta distribution characterised as \citep{vovk2012conditional}:
\begin{equation}\label{eq: cov_dist}
\frac{1}{n_{\text{val}}} \sum_{i=1}^{n_{\text{val}}} I_{\mathbb{C}^{\alpha}}(Y_{i}) \sim \text{Beta}(n_\text{cal} + 1 - l, l),
\end{equation}
where $l = \lfloor(n_{\text{cal} + 1})(1 - \alpha)\rfloor$ and $n_\text{cal}$ is the calibration set size.

\subsection{Computational Complexity of Calibration}
\label{sec: comp_complex}
The conformal prediction calibration procedure has complexity $\mathcal{O}(d \cdot n_{\text{cal}} \log n_{\text{cal}})$, where $d = T_{\text{out}} \times N_x \times N_y \times N_{\text{var}}$ is the output dimensionality and $n_{\text{cal}}$ is the calibration set size. This comprises: (1) computing nonconformity scores $s(x,y)$ for all $n_{\text{cal}}$ samples across $d$ dimensions, requiring $\mathcal{O}(n_{\text{cal}} \cdot d)$ element-wise operations (e.g., $|Y - \tilde{Y}|$ for AER), and (2) sorting scores to estimate quantiles $\hat{q}$ per dimension, requiring $\mathcal{O}(d \cdot n_{\text{cal}} \log n_{\text{cal}})$ operations. Once calibrated, constructing prediction sets for new predictions requires only $\mathcal{O}(d)$ operations by applying the pre-computed quantiles. The procedure requires only forward passes and sorting---no gradient computation or model retraining---and is embarrassingly parallel across dimensions. This computational efficiency is particularly advantageous compared to alternative UQ methods that require ensemble training ($\mathcal{O}(n_{\text{ensemble}} \cdot \text{training cost})$) or extensive Bayesian sampling. As shown in \cref{table: coverage_all_models}, calibration remains practical even for $d > 10^7$ dimensions, with times ranging from $<1$ second (low-dimensional cases) to a few hundred seconds (20M+ dimensions for weather forecasting), demonstrating the near-zero computational cost that makes CP particularly suitable for production deployment of surrogate models.

\subsection{Exchangeability Requirements}
\label{sec: exchangeability}

We treat spatio-temporal surrogate modelling as an initial-value problem (IVP), where models evolve initial states autoregressively or in one-shot mappings. Each input-output pair $(X_i, Y_i)$ from calibration and prediction sets is assumed exchangeable when initial conditions are sampled i.i.d. from the distribution of interest. This assumption requires: (1) consistent spatio-temporal structure (\cref{subsec: formal_definition}) across the inputs and output, (2) identical discretised domains across calibration and prediction, and (3) exchangeability between calibration and test data \citep{gentle_introduction_CP}.

\begin{figure}
    \centering
    \includegraphics[width=\linewidth]{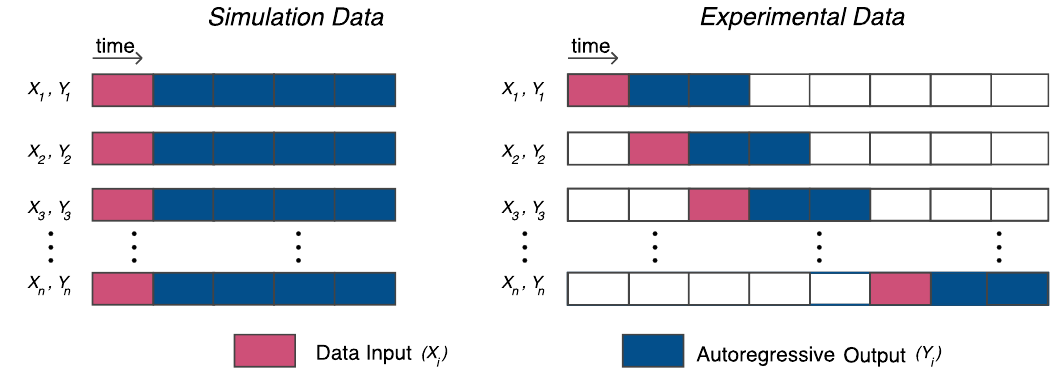}
    \caption{Constructing exchangeable pairs for simulation versus experimental data. \textit{Simulation-based}: Multiple independent runs with varying initial conditions at $t=0$, each forecasting the same time horizon to $t=T$. \textit{Experimental data}: Single long-duration experiment partitioned into multiple IVPs with different starting times $T_i$ from a larger temporal domain, each forecasting the same duration $\Delta T$.}
    \label{fig:time_windowing_diff}
\end{figure}

Our experiments (\cref{sec: experiments}) involve two distinct scenarios for constructing exchangeable pairs (\cref{fig:time_windowing_diff}):

\subsubsection{Simulation-Based Exchangeability}

For surrogate models of numerical simulations (\cref{sec: poisson}--\cref{mpp}), exchangeability is straightforward. The modelling task maps from the simulation start $t=0$ to a fixed future time $t=T$. Each simulation begins at $t=0$ with initial conditions sampled from a distribution $\mathcal{P}_{\text{IC}}$ characterising the parameter space of interest. For example, in the MHD plasma blob experiments (\cref{sec: mhd}), initial conditions vary in blob positions, widths, and amplitudes (\cref{table: data_generation_mhd}), while the temporal evolution window $[0, T]$ remains identical across all simulations. 

This creates a natural exchangeability structure: each pair $(X_i, Y_i)$ represents an independent draw from the joint distribution of initial conditions and their corresponding solutions. Multiple simulations with varied physical parameters provide abundant exchangeable pairs without violating temporal dependencies, as each simulation is an independent realisation of the physical system. The computational cost of generating additional calibration data is primarily limited by simulation expense rather than data availability.

\subsubsection{Experimental Data Exchangeability}

For surrogate models trained on experimental observations (\cref{sec:nwp}, \cref{sec: fno_camera}), exchangeability is constructed through temporal windowing while preserving the IVP structure. Experimental data, such as continuous weather observations or plasma diagnostics, form long time series from which we extract multiple training examples.

The key insight is constructing exchangeable IVPs from a single temporal sequence. Each pair $(X_i, Y_i)$ represents an IVP where: (1) initial condition $X_i$ is extracted at time $t = T_i$ from the larger domain, (2) the model forecasts forward for fixed duration $\Delta T$, (3) target $Y_i$ spans $[T_i, T_i + \Delta T]$, and (4) starting time $T_i$ varies while $\Delta T$ remains constant. Here, these pairs can be treated as exchangeable, as we consider the data distribution to be characterised by a wide range of initial conditions, and the prediction window remains the same. 

Crucially, predictions depend \textit{only} on $X_i$, independent of absolute time $T_i$. A weather forecast initialised at 12:00 January 5th, predicting 48 hours ahead is exchangeable with one initialised at 18:00 January 12th with the same forecast horizon, such that both solve the same IVP: ``given these atmospheric conditions, predict evolution over 48 hours.'' This time windowing transforms a single time series into $N$ exchangeable pairs by sampling initial conditions $T_1, T_2, \ldots, T_N$ from distribution $\mathcal{P}_T$ representing typical states within the observed period. The fixed window $\Delta T$ ensures identical structure across pairs, satisfying \cref{subsec: formal_definition}.

For fusion camera diagnostics (\cref{sec: fno_camera}), we extract 10-frame initial conditions from different time points within discharge shots, each predicting the subsequent 10 frames. The model learns mappings from the current plasma state to the near-future state, agnostic to absolute discharge time. Similarly, weather forecasting (\cref{sec:nwp}) constructs pairs from different initialisation times across months, each representing the same multi-day forecast problem.

\textbf{Important limitation:} Exchangeability can be violated with experimental data when calibration and test distributions differ (e.g., different seasons in weather, different plasma regimes in fusion). We demonstrate this sensitivity in \cref{sec:nwp} and \cref{sec: fno_camera}, showing reduced coverage when exchangeability assumptions break. Users must verify distributional similarity or recalibrate when conditions change substantially.

\section{Experiments}
\label{sec: experiments}

We empirically validate our CP framework across diverse surrogate models trained on spatio-temporal data from physical systems. Our experiments span multiple neural architectures commonly deployed in scientific modelling: Multi-Layer Perceptrons (MLPs) \citep{haykin1994neural}, U-Nets \citep{ronneberger2015unet}, Fourier Neural Operators (FNOs) \citep{li2021fourier}, vision transformers \citep{Yin_2022_CVPR}, and graph neural networks \citep{GNNs2009}. These architectures have proven effective for surrogate modelling in diverse applications, including wind turbine design \citep{LALONDE2021104696}, high-energy physics \citep{Baldi2016}, fusion reactors \citep{Manek_2023}, fluid dynamics \citep{gupta2022multispatiotemporalscale}, carbon capture \citep{Wen_2023}, weather forecasting \citep{pathak2022fourcastnet,lam2022graphcast}, and plasma evolution \citep{GOPAKUMAR2023100464}.

\paragraph{Relationship to Other UQ Methods:} In this work, we focus on empirically demonstrating how inductive CP provides calibrated error bars across diverse spatio-temporal models using various non-conformity scores (as detailed in \cref{nonconformity scores}). Our framework addresses both deterministic models (using AER) and probabilistic models—whether frequentist (CQR) or Bayesian (STD)—showcasing CP's model-agnostic nature and its ability to provide or calibrate guaranteed coverage across these different paradigms.

CP fundamentally differs from alternative UQ approaches such as deep ensembles and Bayesian neural networks by providing finite-sample, distribution-free guarantees on coverage. In contrast, ensemble and Bayesian methods provide asymptotic or model-dependent uncertainty estimates without such guarantees. While these methods can produce useful uncertainty estimates, they often fail to achieve desired coverage without post-hoc calibration. A comprehensive empirical comparison between CP and Bayesian deep learning methods for surrogate modelling—including coverage reliability, computational costs, and calibration quality—is presented in \citet{gopakumar2025calibrated}, where we demonstrate that methods like MC Dropout and deep ensembles frequently require further calibration to achieve valid coverage. The present work establishes CP's applicability and scalability to high-dimensional spatio-temporal problems across multiple scientific domains. 

\paragraph{Computational Setup:} All models were trained on NVIDIA A100 GPUs with 80GB memory. Calibration and prediction sets were evaluated on standard laptop hardware, demonstrating the computational efficiency of our approach. The experiment procedure is outlined in the \cref{alg: cp_structure} below.

\begin{algorithm}[h!]
\renewcommand{\thealgorithm}{}
\caption{CP Framework Structure}
  \begin{algorithmic}[1]
    \State Generate/gather training data (simulation or experimental)
    \State Train surrogate model (or use pre-trained model)
    \State Generate/gather calibration dataset (or use fine-tuning dataset)
    \State Compute nonconformity scores and quantiles via CP framework
    \State Construct prediction sets with guaranteed coverage
    \State Validate coverage on independent test set
  \end{algorithmic}
\label{alg: cp_structure}
\end{algorithm}

\begin{table}[h!]
\caption{Comprehensive coverage results for $\alpha=0.1$ (90\% target coverage). Uncalibrated coverage (unavailable for AER) shows initial estimates before CP calibration. Calibration time is reported on standard laptop hardware. Tightness represents average error bar width in normalised units (Min-Max normalisation between -1 and 1). $^*$ indicates out-of-distribution evaluation where calibration and training distributions differ. PT: pre-trained model; FT: fine-tuned model.}
\label{table: coverage_all_models}
\centering
\scalebox{0.7}{
\begin{tabular}{l|l|l|l|l|l|l|l}
\toprule
Case & Model & Output Dims & Method & Uncalib. (\%) & Calib. (\%) & Cal. Time (s) & Tightness \\
\midrule
& & & CQR & 94.61 & 90.01 & 0.0035 & 0.012\\
\textbf{1D Poisson} & MLP & 32 & AER & - & 90.05 & 0.0030 & 0.002\\
& & & STD & 97.5 & 90.85 & 0.133 & 0.025 \\
\midrule
& & & CQR & 25.53 & 93.05 & 19.70 & 0.314 \\
\textbf{1D Conv-Diff} & U-Net & 2,000 & AER & - & 92.60 & 8.30 & 0.266\\
& & & STD & 88.43 & 90.29 & 88.15 & 0.164 \\
\midrule
& & & CQR & 96.95 & 89.21 & 8.40 & 0.132 \\
& U-Net & 32,670 & AER & - & 94.91 & 3.52 & 0.013\\
\textbf{2D Wave} & & & STD & 4.45 & 90.30 & 39.51 & 0.012 \\\cline{2-8}
& FNO$^*$ & 65,340 & AER & - & 89.24 & 34.18 & 0.330\\
& & & STD & 32.81 & 89.83 & 462.0 & 0.669\\
\midrule
\textbf{2D Navier-Stokes} & FNO$^*$ & 40,960 & AER & - & 90.08 & 4.83 & 0.381\\
& & & STD & 7.52 & 90.27 & 64.75 & 0.448 \\
\midrule
\textbf{2D MHD} & FNO & 1,348,320 & AER & - & 90.18 & 359.12 & 0.039 \\
\midrule
\textbf{2D MHD} & ViT$^*$ (PT) & 313,344 & AER & - & 89.95 & 2980.50 & 0.062 \\
&  ViT$^*$ (FT)&  & AER & - & 89.78 & 2078.50 & 0.015 \\
\midrule
\textbf{2D Camera} & FNO & 2,867,200 & AER & - & 91.28 & 293.62 & 0.131 \\
\midrule
\textbf{2D Weather} & GNN & 20,602,232 & AER & - & 91.19 & 229.23 & 1.13 \\
\textbf{(Limited Area)} & GNN &  & STD & 73.55 & 89.97 & 309.55 & 0.91 \\
\midrule
\textbf{2D Weather} & GNN & 12,777,600  & AER & - & 90.03 & 366.41 & 1.34 \\
\textbf{(Global)} & GNN &  & STD & 71.22 & 89.88 & 400.57 & 1.28 \\
\bottomrule
\end{tabular}
}
\end{table}

Table~\ref{table: coverage_all_models} presents our main empirical findings. Across all experiments, models, and nonconformity scores, we achieve near-perfect calibration to the target 90\% coverage ($\alpha = 0.1$). For methods with initial uncertainty estimates (CQR and STD), we show both uncalibrated and calibrated coverage, demonstrating how CP corrects potentially misleading uncertainty quantification. The tightness metric, computed as average error bar width in the normalised space (linear transformation of the field to lie between -1 and 1), reveals that AER generally provides the tightest fits, with STD as a close second. Critically, these results hold across output dimensions ranging from 32 to over 20 million, demonstrating that our framework overcomes the curse of dimensionality. All models were trained on an Nvidia A100 GPU and evaluated and calibrated over standard laptop hardware. 

\subsection{1D Poisson Equation}
\label{sec: poisson}

The Poisson equation generalises the Laplace equation and models diverse phenomena, including electrostatics, gravitation, and fluid potential fields \citep{Hackbusch2017}. This steady-state elliptic PDE serves as our simplest test case, mapping an initial field distribution to its equilibrium state along a 1D domain $[0,1]$ discretised into 32 uniform points.

\textbf{Dataset:} We generated 7,000 simulations using finite difference methods (py-pde package \citep{py-pde}) by varying the initial field value $u_{\text{init}} \sim \mathcal{U}(0, 4)$, allocated as 5,000 training, 1,000 calibration, and 1,000 validation samples. All datasets were sampled from the same distribution.

\textbf{Models and Training:} We trained separate MLPs (3 layers, 64 neurons per layer) for each nonconformity method: (i) three models for CQR modeling the $5^{\text{th}}$, $50^{\text{th}}$, and $95^{\text{th}}$ quantiles using quantile loss \citep{koenker_2005}; (ii) one deterministic model for AER using L1 loss; and (iii) one probabilistic model with dropout layers for STD. Training used the Adam optimiser \citep{adam} with initial learning rate 0.005 (halved every 100 epochs) for up to 1,000 epochs. Further details about the
physics, data generation strategies and model training can be found in \cref{appendix_poisson}.

\begin{figure}[ht]
    \centering
    \includegraphics[width=\textwidth]{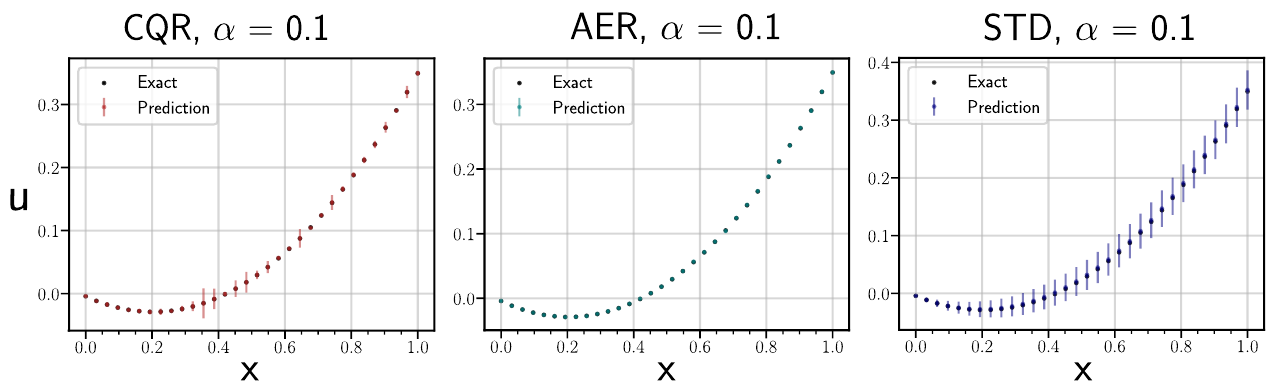}
    \caption{Calibrated prediction sets ($\alpha=0.1$, 90\% coverage) for the 1D Poisson equation using three nonconformity scores (CQR, AER, STD). The simple dynamics enable near-perfect model fit, resulting in tight, well-calibrated error bars. Ground truth (black line), model prediction (blue line), and shaded regions depict the prediction sets.}
    \label{fig: cp_comparison_poisson}
\end{figure} 

\textbf{Results:} Figure~\ref{fig: cp_comparison_poisson} visualizes the $\alpha=0.1$ prediction sets for all three methods. The MLP learns this simple mapping with high accuracy, yielding tight uncertainty bounds. All methods achieve the guaranteed coverage (Table~\ref{table: coverage_all_models}), with AER providing the tightest fit (0.002 normalised units) and minimal calibration time (3 ms). The probabilistic STD method achieves comparable coverage but with slightly wider bounds (0.025 normalised units) and longer calibration time (133 ms) due to Monte Carlo dropout sampling.

\subsection{1D Convection-Diffusion Equation}
\label{conv_diff}

We advance to a spatio-temporal system governed by the convection-diffusion equation, which combines parabolic and hyperbolic PDE characteristics to model transport phenomena across diverse applications \citep{Chandrasekhar}. This equation describes how a fluid density field evolves under the competing effects of diffusion (smoothing) and convection (transport). We consider a spatially-varying diffusion coefficient and Gaussian initial conditions parametrised by mean and variance, with the system evolving over a 1D spatial domain $x \in [0,10]$ and time interval $t \in [0, 0.1]$.

\textbf{Dataset:} We generated 5,000 simulations using a forward-time centred-space finite difference scheme across 200 spatial points and 100 time steps. The training set (3,000 samples) was generated via Latin hypercube sampling over physically relevant ranges of diffusion coefficients, convection velocities, and initial condition parameters. To test out-of-distribution robustness, calibration and validation sets (1,000 each) were sampled from a shifted parameter regime with reduced diffusion and enhanced convection. This distribution shift mimics real-world deployment scenarios where test conditions may differ from training.

\textbf{Model and Training:} A 1D U-Net with 4 encoder-decoder levels maps the first 10 time steps (downsampled from 100) to the next 10 steps. The architecture uses batch normalisation and tanh activations. For STD, we added dropout layers (rate 0.1) after each encoder-decoder block. Training followed the same optimiser schedule as the Poisson case, using quantile loss for CQR and MSE loss otherwise. Further details about the
physics, data generation strategies and model training can be found in \cref{appendix_CD}.

\begin{figure}[ht]
    \centering
    \includegraphics[width=\textwidth]{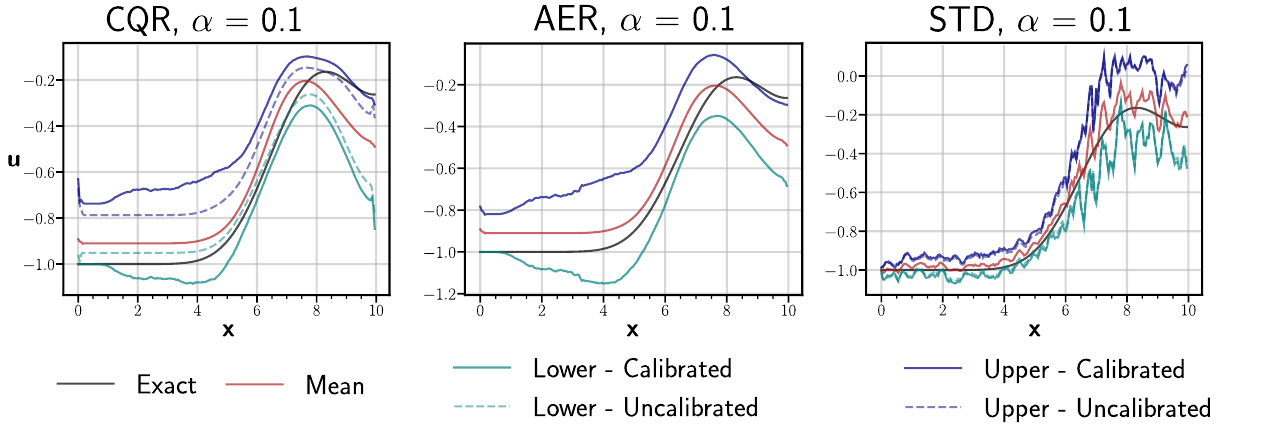}
    \caption{Calibrated prediction sets ($\alpha = 0.1$, 90\% coverage) for 1D convection-diffusion evaluated on out-of-distribution data at final time $t = 0.1$. CP successfully calibrates initially insufficient coverage (CQR: 25.53\% $\rightarrow$ 93.05\%) or refines sampled uncertainty (STD: 88.43\% $\rightarrow$ 90.29\%) to guarantee target coverage. Ground truth (black), prediction (blue), calibrated bounds (red shaded), and uncalibrated bounds (gray dashed) where applicable.}
    \label{fig: cp_comparison_convdiff}
\end{figure}

\textbf{Results:} Despite testing on a physical regime with 2$\times$ higher convection and half the diffusion compared to training data (Figure~\ref{fig: cp_comparison_convdiff}), CP provides valid guaranteed coverage (Table~\ref{table: coverage_all_models}). The uncalibrated CQR severely underestimates uncertainty (25.53\% coverage), while CP calibration increases this to 93.05\%. For STD, the uncalibrated 88.43\% coverage is refined to 90.29\%. Figure~\ref{fig: Validation_Plots} confirms guaranteed coverage across all $\alpha$ levels for both training-distribution and out-of-distribution settings. While CQR and AER produce conservative bounds (potentially due to model over-fitting on this relatively simple task), all methods maintain coverage guarantees. This experiment demonstrates a critical capability: \textbf{CP provides valid uncertainty quantification even when deployed outside the training distribution}, provided exchangeability holds between calibration and prediction regimes.

\subsection{2D Wave Equation}
\label{wave}

The 2D wave equation models wave propagation in acoustics, optics, and quantum mechanics \citep{thermodynamics_textbook}. We simulate Gaussian wave packets evolving on a $33 \times 33$ spatial grid over 80 time steps. A training dataset of 500 simulations is generated by varying the amplitude and position of the initial Gaussian. We train both a U-Net (feed-forward) and an FNO (autoregressive) to model the temporal evolution. Calibration and validation each use 100 additional simulations respectively. Full physics details and numerical methods are in \cref{appendix_wave}.

\subsubsection{U-Net}
\label{sec: wave-unet}

The U-Net performs a single feed-forward mapping from 20 input time steps to 30 output time steps, producing outputs of shape $[30, 33, 33]$. Coverage is estimated cell-wise across the output tensor following \cref{subsec: formal_definition}.

\begin{figure}[h!]
    \centering
    \includegraphics[width=\textwidth]{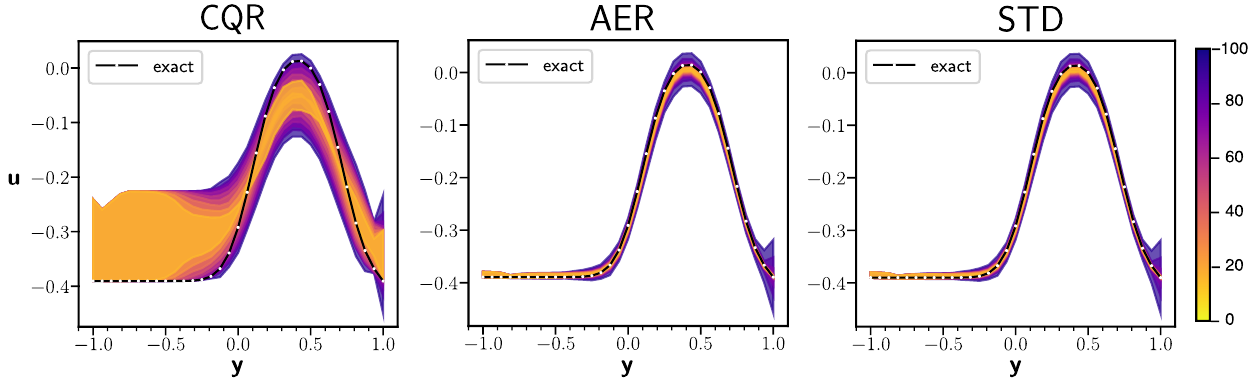} 
    \caption{Prediction sets for the 2D wave equation using three nonconformity scores. Colour bars show coverage levels (in \%, corresponding to $1-\alpha$). Each panel displays a spatial slice along the $y$-axis at the final time step. The ground truth (black with white markers) lies within the calibrated bounds, validating \cref{eq: coverage}.}
    \label{fig: cp_comparison_wave}
\end{figure}

\Cref{fig: cp_comparison_wave} shows spatial slices of prediction sets at multiple $\alpha$ levels. All three methods achieve valid coverage (\cref{fig: Validation_Plots}, \cref{table: coverage_all_models}), with AER being computationally cheapest. Both CQR and AER produce conservative (wide) intervals, as expected from the inequality in \cref{eq: coverage}. AER and STD provide tighter fits than CQR while maintaining coverage guarantees.

\paragraph{Out-of-Distribution Testing.} To test robustness, we generate new calibration and validation datasets by solving the wave equation with half the wave velocity used during training. This tests CP's effectiveness when the calibration regime differs from the training distribution. As shown in \cref{fig:cell-wise-cp}, uncalibrated MC dropout fails to capture modelling errors in this out-of-distribution regime. In contrast, CP provides statistically guaranteed bounds regardless of the model's training conditions. This demonstrates CP's value for deploying pre-trained surrogates in new physical regimes without costly retraining.

\subsubsection{FNO}

We train an FNO in an autoregressive framework: the model takes 20 initial time steps, predicts the next 10 steps, then recursively unrolls to produce 60 total output steps (shape $[60, 33, 33]$). CP is performed over the entire rolled-out output. Since FNOs perform best with relative $L^p$ loss, we omit CQR for this architecture.

\begin{figure}[h!]
    \centering
    \subfloat[In-distribution]{\includegraphics[width=0.65\textwidth]{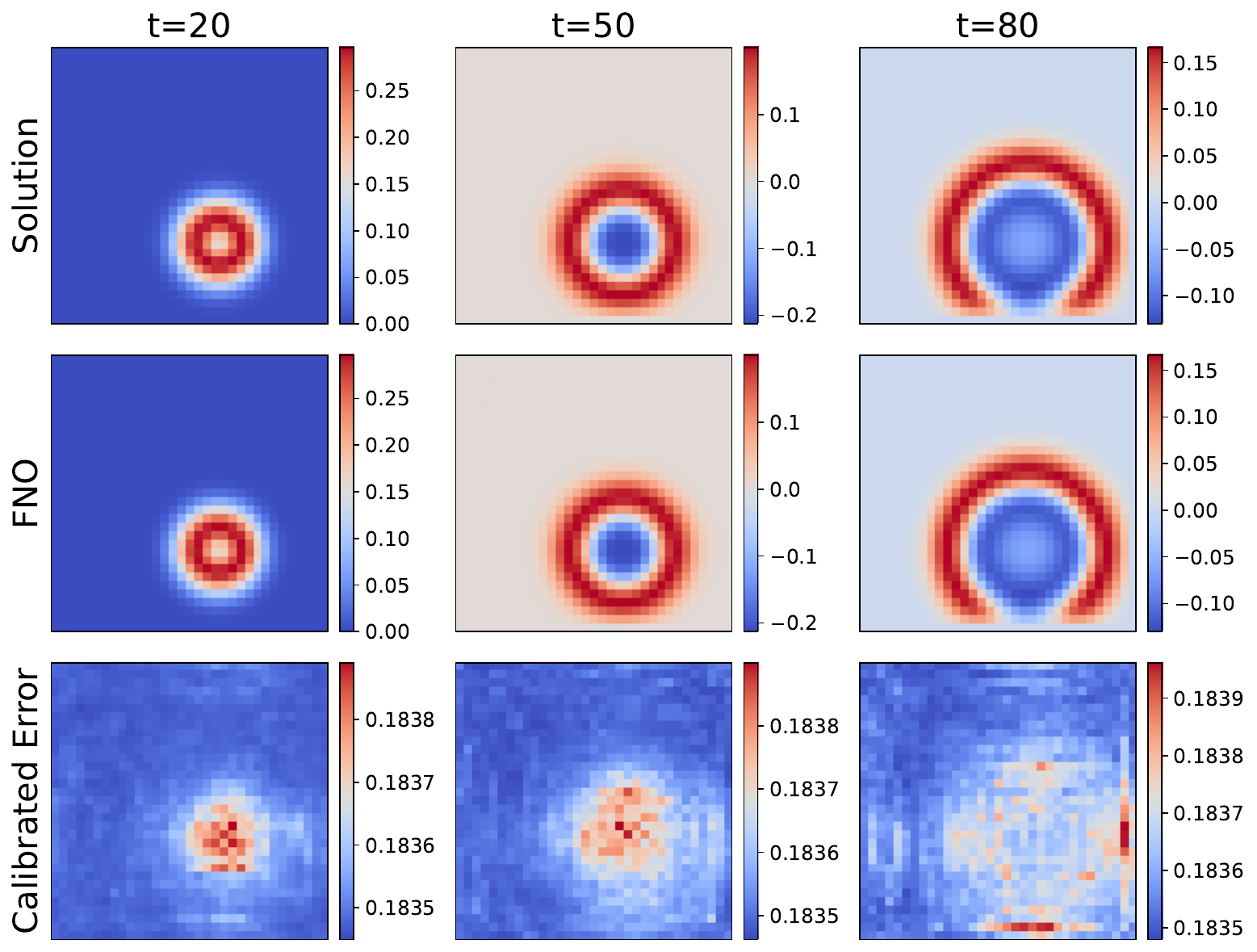}}
    \label{fig:wave_normalspeed}
    \hspace{10mm}
    \subfloat[Out-of-distribution]{\includegraphics[width=0.65\textwidth]{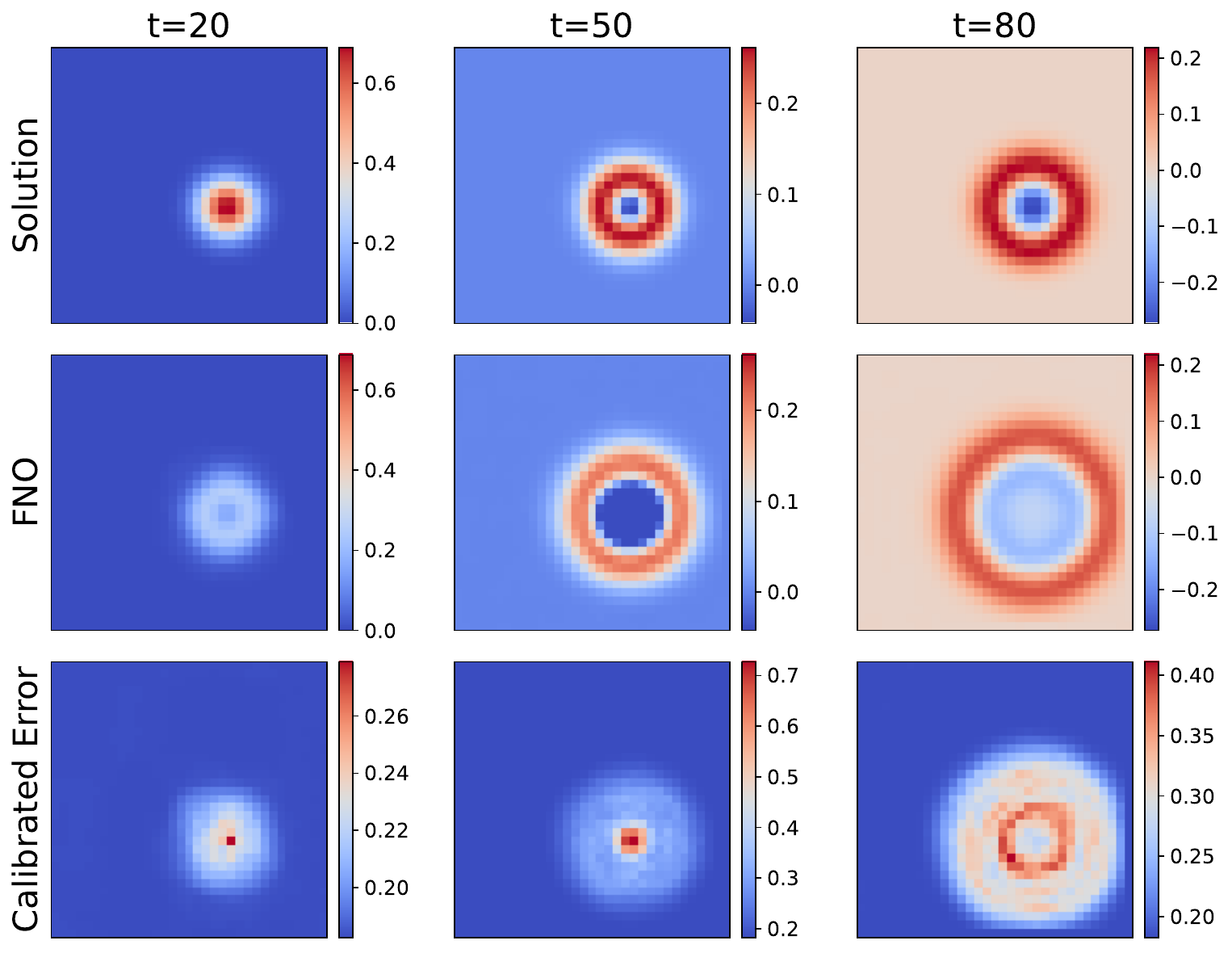}}
    \label{fig:wave-halfspeed}
    \caption{CP performance for autoregressive FNO on the wave equation. Top row: ground truth; middle row: FNO predictions; bottom row: calibrated error bars (90\% coverage). Valid coverage is maintained both within (a) and outside (b) the training distribution, demonstrating CP's robustness to distribution shift.}
    \label{fig: wave_res_fno}
\end{figure}

The FNO is tested in both in-distribution and out-of-distribution (half-speed) settings, matching the U-Net experiments. \Cref{fig: wave_res_fno} shows that CP provides valid error bars under both conditions. The key requirement is exchangeability between calibration and prediction regimes, not similarity to training data. This enables valuable UQ even for previously unseen solution families. The FNO achieves tighter coverage than the U-Net (\cref{fig: Validation_Plots}), likely due to its operator learning formulation. Additional coverage plots are in \cref{fig: fno_halfspeed} (\cref{appendix_wave}).

\subsection{2D Navier--Stokes Equations}

The incompressible 2D Navier--Stokes equations describe viscous fluid dynamics, modelling conservation of mass and momentum. Their complexity and strong nonlinearity necessitate computational fluid dynamics (CFD) solvers. Neural-PDE methods offer efficient alternatives at scale \citep{Azizzadenesheli2024}. Following \citet{li2021fourier}, we train an FNO to model vorticity evolution. The model is trained on simulations with viscosity $\nu=10^{-3}$, then calibrated and tested on data with $\nu=10^{-4}$ (out-of-distribution). The FNO maps 10 input time steps to the next 10 output steps. Physics details and training specifications are in \cref{appendix_ns}.

To enable STD-based CP, we modify the FNO architecture by adding dropout layers, creating a probabilistic operator. The resulting model outputs both mean predictions and uncertainty estimates via MC dropout sampling.

\begin{figure}[h!]
    \centering
    \includegraphics[width=0.75\textwidth]{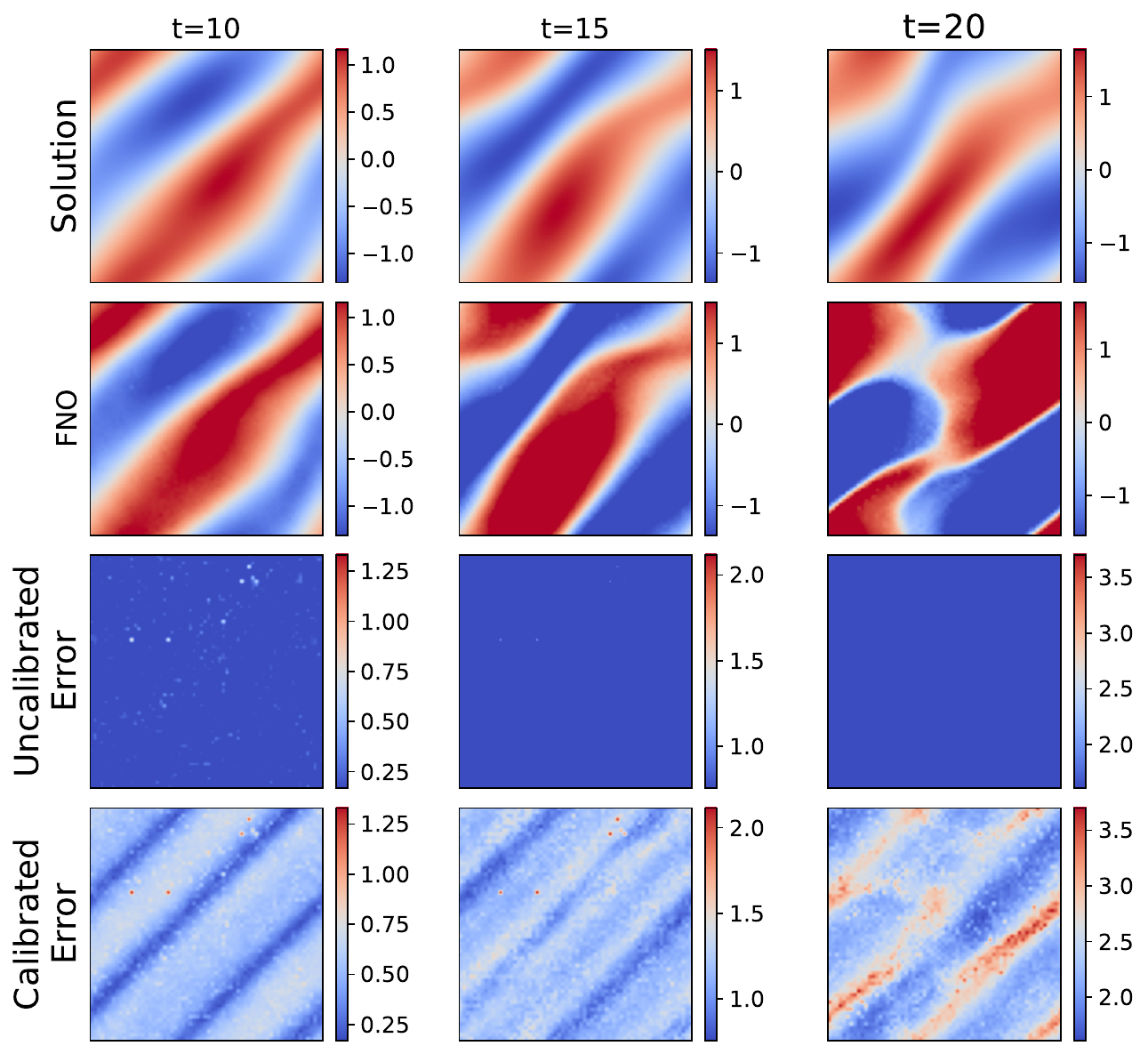}
    \caption{Calibrating probabilistic FNO predictions for out-of-distribution Navier--Stokes data. Rows show: ground truth, FNO predictions, uncalibrated MC dropout uncertainty (standard deviation), and calibrated CP error bars (67\% coverage). The uncalibrated dropout underestimates errors, but CP corrects this to achieve the target coverage.}
    \label{fig:ns_fno_heatmap}
\end{figure}

\Cref{fig:ns_fno_heatmap} demonstrates CP's ability to calibrate probabilistic models. The uncalibrated dropout-based uncertainty provides only 7.52\% coverage, severely underestimating prediction errors. CP adjusts these intervals to achieve the target 67\% coverage (we show 67\% instead of the standard 90\% to better visualise the calibration effect at moderate coverage levels). Coverage validation across all $\alpha$ levels is shown in \cref{fig: Validation_Plots}. This helps illustrate CP's dual utility: providing guarantees for deterministic models (via AER) and calibrating uncalibrated probabilistic models (via STD).

\subsection{2D Magnetohydrodynamics}
\label{sec: mhd}

Magnetohydrodynamics (MHD) couples the Navier--Stokes equations with Maxwell's equations to model plasma evolution in fusion devices such as tokamaks \citep{bellan2006fundamentals}. We consider a reduced-MHD system \citep{Hoelzl2021jorek} describing multiple plasma blobs in a non-uniform temperature field. The system evolves three coupled variables---density ($\rho$), electrostatic potential ($\Phi$), and temperature ($T$)---on a $106 \times 106$ toroidal grid (coordinates $R$, $Z$). This represents a challenging multi-variable, multi-physics problem. Full physics equations are in \cref{appendix_mhd}.

We use 2000 simulations from the JOREK code \citep{Hoelzl2021jorek}, split into 1000 for training, 500 for calibration, and 500 for validation. Each simulation varies the initial conditions (blob positions, widths, amplitudes) while keeping the physics parameters fixed. The dataset and pre-trained model are taken from \citet{Gopakumar_2024}.

A multi-variable FNO learns the coupled dynamics of all three fields simultaneously. The model autoregressively predicts from 10 input time steps to 5 output steps, recursively continuing to the 50th time step. CP is performed over the complete spatio-temporal domain, yielding prediction sets for each cell in the 4D output tensor (time $\times$ $R$ $\times$ $Z$ $\times$ variables). Given the model's scale (9.4M parameters, 1.35M output dimensions), we use only the AER method for computational efficiency.

\begin{figure}[h!]
    \centering
    \includegraphics[width=1.0\textwidth]{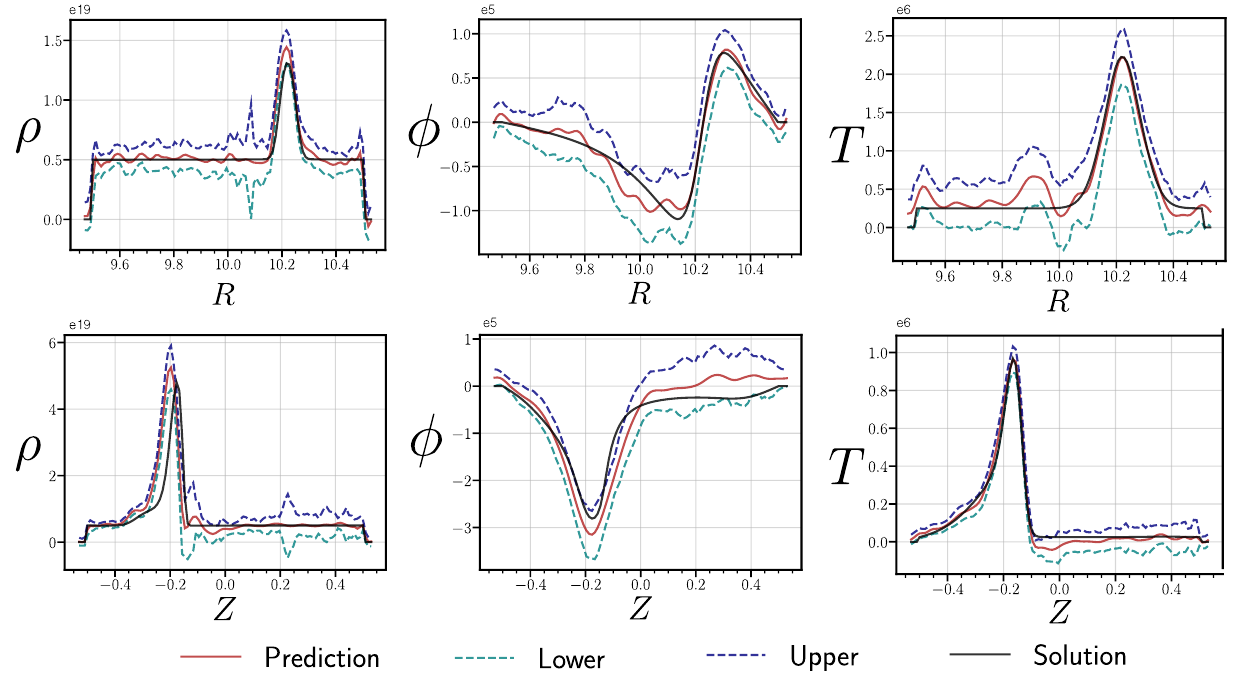}
    \caption{Spatial profiles at timestep 20 showing 90\% coverage ($\alpha = 0.1$) for multi-variable MHD predictions. Columns: density $\rho$ (left), potential $\Phi$ (center), temperature $T$ (right). Rows show profiles along the $R$-axis (top) and $Z$-axis (bottom). For each variable, we display the numerical solution, FNO prediction, and calibrated upper/lower bounds. CP accurately captures the sharp features of plasma blobs across all three coupled variables.}
    \label{fig: cp_mhd}
\end{figure}

\Cref{fig: cp_mhd} shows spatial slices through the prediction sets for all three variables at a single time step. The calibrated error bars successfully bound the sharp density, potential, and temperature peaks characterising the plasma blobs. Coverage validation (\cref{fig: Validation_Plots}) confirms guaranteed marginal coverage across all 1,348,320 output dimensions ($50 \times 106 \times 106 \times 3$), demonstrating CP's scalability and immunity to the curse of dimensionality. The bounds provide interpretable confidence estimates for each variable's spatial distribution, enabling assessment of the surrogate's reliability across the operational domain.

\subsection{Foundational Physics Models}
\label{mpp}

Foundation models pre-trained on diverse PDE datasets \citep{bommasani2022opportunities, mccabe2023multiple, alkin2024upt, hao2024dpot, rahman2024pretraining} have emerged as a promising approach for multi-task scientific modelling. These models employ transformer-based architectures with attention mechanisms across spatio-temporal domains, learning shared representations of differential operators (e.g., diffusion, convection) common across PDE families. This enables them to capture global behaviours during pre-training, leaving task-specific local features for fine-tuning \citep{alkin2024upt}.

As these models scale and deploy across safety-critical applications, UQ becomes essential. CP offers an efficient validation framework: for fine-tuned models, the existing fine-tuning data can serve as calibration data, eliminating the need for additional simulations. This is justified because fine-tuning aims to align the model with a specific target distribution, making performance within that distribution the primary concern.

We apply CP to the Multi-Physics Pre-trained Adaptive Vision Transformer (MPP-AViT) of \citet{mccabe2023multiple}. This model uses shared embeddings and normalisation across variables, with an AViT backbone \citep{yin2022avit} that sequentially attends over space and time. The model autoregressively predicts the next time step given current field values. We refer to \citet{mccabe2023multiple} for full architectural and training details.

\subsubsection{Pre-trained Model: Zero-Shot Learning}

We test the largest pre-trained model (MPP-AViT-L, 409M parameters) on MHD density evolution---a physics regime not seen during training. The model was pre-trained on shallow-water, diffusion-reaction, and Navier--Stokes equations, learning to model densities, velocities, and pressures. We extract only density fields from our MHD dataset (\cref{sec: mhd}) for inference. The model takes 16 input time steps and autoregressively predicts a single step forward until the 50th time step.

\begin{figure}[h!]
    \centering
    \includegraphics[width=0.9\textwidth]{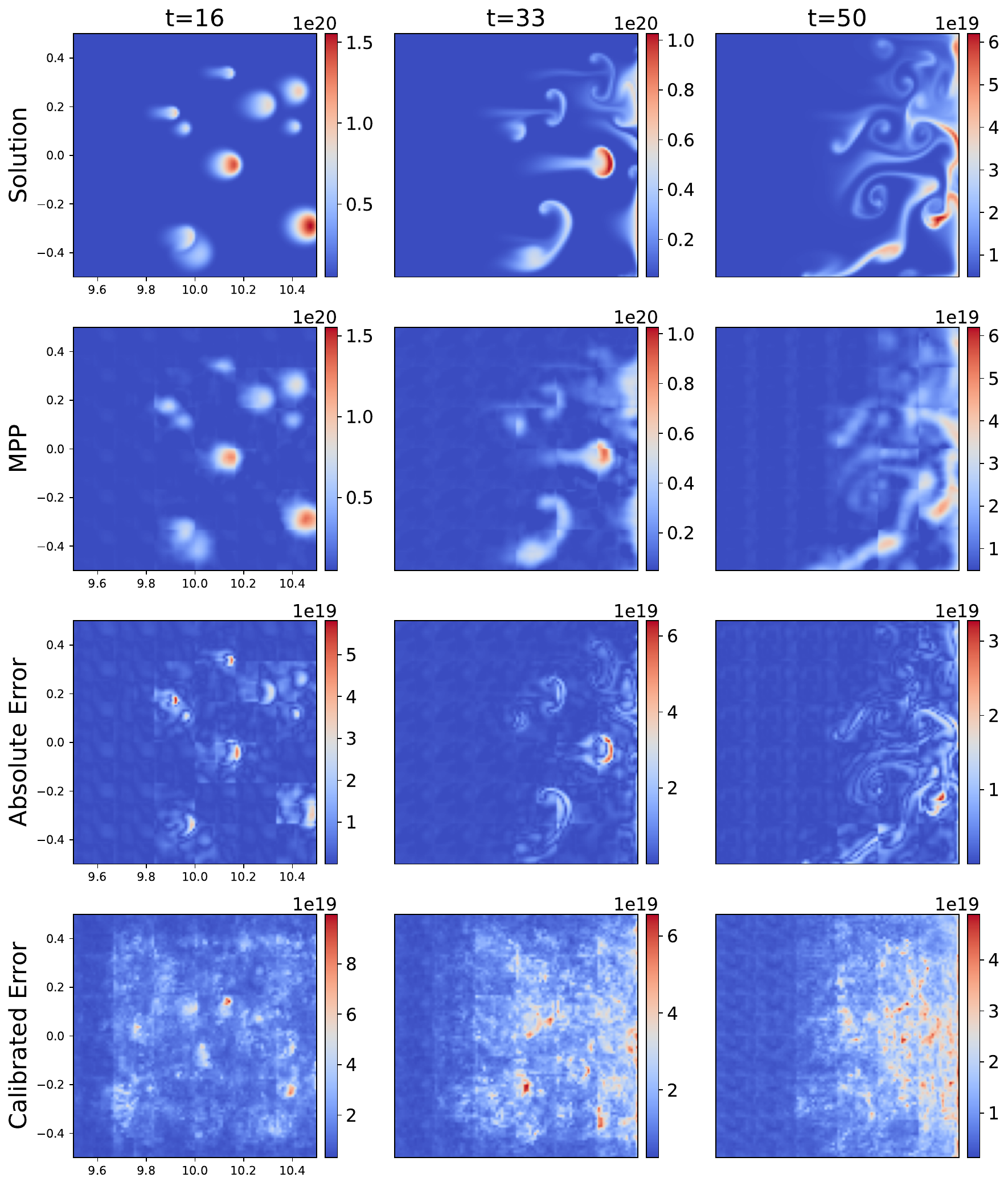}
    \caption{Zero-shot MPP-AViT performance on out-of-distribution MHD density evolution. Rows show: ground truth, model predictions, absolute error, and calibrated error bar width (95\% coverage). Despite no training on MHD physics, the model captures major blob features. CP quantifies the prediction uncertainty with guaranteed coverage.}
    \label{fig:mhd_mpp_pretrain}
\end{figure}

\Cref{fig:mhd_mpp_pretrain} shows that despite zero exposure to MHD during training, the model captures the major features of plasma blob evolution (radial outward motion). However, finer details are lost, and patch-based artefacts appear, a known limitation of the architecture \citep{mccabe2023multiple}. Using 1000 calibration data points and the AER method, CP provides valid 95\% coverage bounds (\cref{MPP_appendix}). Importantly, for this deterministic model, the calibrated errors ($\hat{q}$) are input-independent constants determined solely by the calibration data, representing global rather than instance-specific uncertainty.

\subsubsection{Fine-tuned Model}

We fine-tune the smaller MPP-AViT-Ti variant on MHD density fields using 75\% of the data for training/calibration and 25\% for validation. Here, the training data serves a dual purpose as calibration data, justified because fine-tuning targets a specific distribution, and we only care about performance within that regime.

\begin{figure}[h!]
    \centering
    \includegraphics[width=0.9\textwidth]{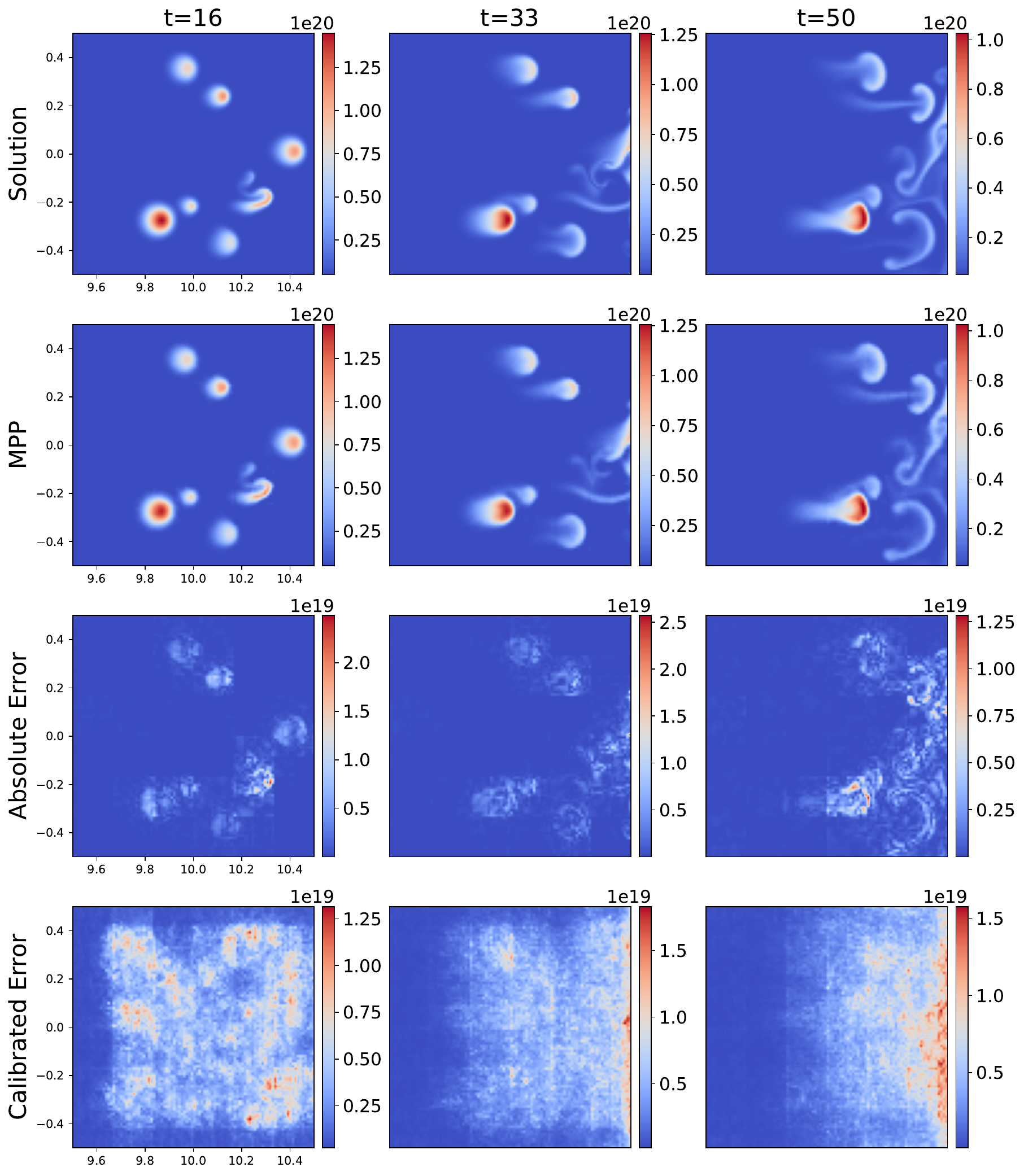}
    \caption{Fine-tuned MPP-AViT performance on MHD density. Rows show: ground truth, predictions, absolute error, and calibrated error bar width (95\% coverage). Fine-tuning dramatically improves accuracy compared to zero-shot (\cref{fig:mhd_mpp_pretrain}). For this well-fitting deterministic model, calibrated errors capture global uncertainties in regions of high dynamics.}
    \label{fig:mhd_mpp_finetune}
\end{figure}

\Cref{fig:mhd_mpp_finetune} demonstrates substantial improvement over zero-shot performance. The fine-tuned model accurately captures both major and fine-scale density features. CP provides guaranteed coverage (\cref{MPP_appendix}), though the error bars reflect global rather than local uncertainties due to the model's deterministic nature. For well-fitted deterministic models like this, CP identifies regions of generally high dynamics across the dataset rather than instance-specific failure modes. Probabilistic models using the STD method would provide input-dependent bounds that adapt to each specific prediction.

\subsection{Neural Weather Prediction}
\label{sec:nwp}
In addition to surrogate models of systems described by explicit PDEs, the proposed methodology is also applicable to more general machine learning models describing physical processes.
To demonstrate this, we here study the use of CP for data-driven weather forecasting models.
Traditional weather forecasting models typically combine PDEs describing large-scale interactions and parametrisations describing subgrid-scale physical processes \citep{atmospheric_modeling_book}.
Data-driven machine learning models approximate this whole process with a single neural network model.
This allows for orders of magnitude faster forecasting speed and, when training on data incorporating observations, also more accurate forecasts \citep{ddw_risk, panguweather, pathak2022fourcastnet, lam2022graphcast}.

Due to the chaotic nature of the weather system, capturing uncertainty in weather forecasts has long been an important consideration both in research and operations.
Such probabilistic modelling has typically been achieved by ensemble forecasting, where perturbations are used to produce samples of possible forecast trajectories \citep{fundamentals_of_nwp}.
Existing data-driven models are still largely deterministic \citep{weatherbench2}.
There are attempts to produce ensemble forecasts using machine learning models by perturbing initial states \citep{pathak2022fourcastnet, fuxi}, training multiple models \citep{calibration_of_large_neurwp} or generative modelling \citep{swin_vrnn, price2024gencast, neural_lam}.
Fundamentally, ensemble forecasting always requires a computational cost proportional to the number of ensemble members, i.e. the number of forecasts made via perturbations.
In contrast, CP offers a cheap method to immediately quantify forecast uncertainty for a time, position, and variable of interest.
This uncertainty can be used by meteorologists interpreting the forecast, conveyed to decision-makers reacting to extreme weather events or directly presented to end-users looking up the forecast for the coming week.
As CP enables UQ for a single forecast output by the model, it is directly applicable to existing deterministic machine learning models.
A limitation of scalar uncertainty estimates is that there are no samples of the distribution over the atmospheric state.
In some scenarios, it can be valuable to inspect such samples to gain an understanding of how different weather scenarios are unfolding.

\newcommand{\lammodel}{Graph-FM\xspace}
\newcommand{\lammodelmse}{\lammodel (MSE)\xspace}
\newcommand{\lammodelnll}{\lammodel (NLL)\xspace}
\newcommand{\armapping}{g}
\subsubsection{Model}
We apply CP to the \lammodel model of \cite{neural_lam}.
\lammodel is a graph-based neural weather prediction model \citep{keisler, lam2022graphcast}, where a hierarchical Graph Neural Network (GNN) is utilised for producing the forecast.
Let $X^t$ denote the full weather state at time step $t$, including multiple atmospheric variables modelled for all grid cells in some discretisation.
Examples of such atmospheric variables are temperature, wind, geopotential and solar radiation.
The GNN $\armapping$ in \lammodel represents the single time step prediction
\begin{equation}
    \label{eq:forecasting_ar}
    X^{t+1} = \armapping\left(X^{t-1:t}, F^{t+1}\right)
\end{equation}
where $F^{t+1}$ are known forcing inputs that should not be predicted.
Taking the two past states as inputs to $\armapping$ allows the model to make use of both magnitude and first derivative information.
\Cref{eq:forecasting_ar} can be applied iteratively to roll out a complete forecast of $T$ time steps.
The full forecasting model can thus be viewed as a mapping from initial weather states $X^{-1:0}$ and forcing $F^{1:T}$ to a forecast $X^{1:T}$.
The forecast $X^{1:T}$ is a tensor of shape $T \times N_x \times N_y \times N_\text{var.}$ where $N_\text{var.}$ is the number of atmospheric variables modelled.
We consider two versions of \lammodel, trained with different loss functions:
\begin{itemize}
    \item \textbf{\lammodelmse:} \lammodel trained with a weighted MSE loss.
    This model outputs only a single prediction, to be interpreted as the mean of the weather state.
    \item \textbf{\lammodelnll:} A version of \lammodel that outputs both the mean and standard deviation for each time, variable and grid cell.
    This model was trained with a Negative Log-Likelihood (NLL) loss, assuming a diagonal Gaussian predictive distribution (also referred to as the uncertainty loss \citep{fengwu}).
    Apart from the change of loss function, the training setup was identical.
\end{itemize}
For the \lammodelmse we compute non-conformity scores using the AER strategy.
As the \lammodelnll is probabilistic, we use STD non-conformity scores.
Note that these are computed based on the standard deviations directly output from the model, rather than from sample estimates based on MC dropout.

\subsubsection{Limited Area Forecasting}
In this first experiment, we apply CP to a limited area version of \lammodel.
Forecasts are here produced for a limited area covering the Nordic region.
These \lammodel models were trained on the limited area dataset from \cite{neural_lam}, consisting of forecasts from the MEPS system \citep{meps}.
One such forecast includes $N_\text{var.} = 17$ variables modelled on a $N_x \times N_y = 238 \times 268$ grid over $T = 19 \times 3h$ time steps (up to 57-hour lead time).
When \lammodel is used in a limited area configuration, it produces weather forecasts for a specific sub-area of the globe.
To achieve this, boundary conditions along the edges of the forecasting area are given as important forcing inputs.
The exact models used have 4 graph processing layers and use 64-dimensional latent representations.
We refer to \cite{neural_lam} for further details about the model and data.

We use forecasts started during September 2021\footnote{For calibration, we specifically use forecasts started during the dates 2021-09-04 -- 2021-09-30. The model was trained on forecasts started during the last days of August, which are rolled out over the first days of September. To avoid strong correlations to the training data, we use only forecasts from September 4 onwards for calibration.} As our calibration data set forecasts that started during September 2022 as test data.
By using the same month for calibration and testing, we minimise the effect of distributional shifts due to seasonal effects.
Having access to calibration data from the same month, collected the previous year, is a reasonable assumption in practical settings.

\begin{figure}[tb]
    \centering
    \begin{subfloat}
        \centering
        \includegraphics[width=\linewidth]{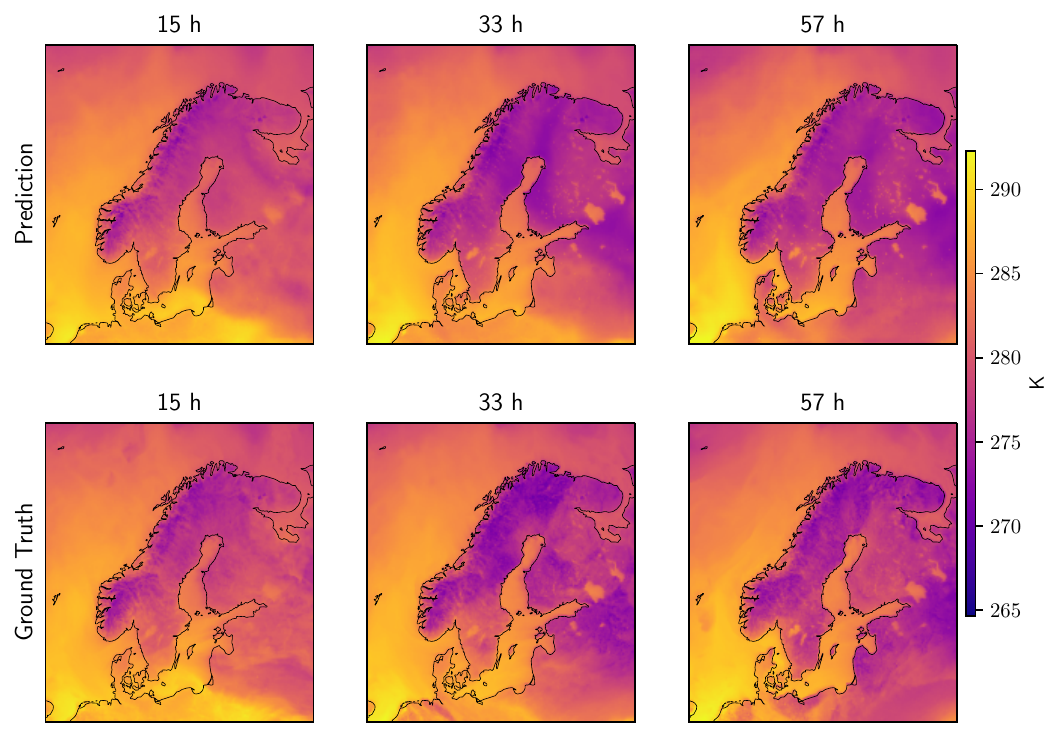}
    \end{subfloat}
    \begin{subfloat}
        \centering
        \hspace{0.02\linewidth}%
        \includegraphics[width=0.95\linewidth]{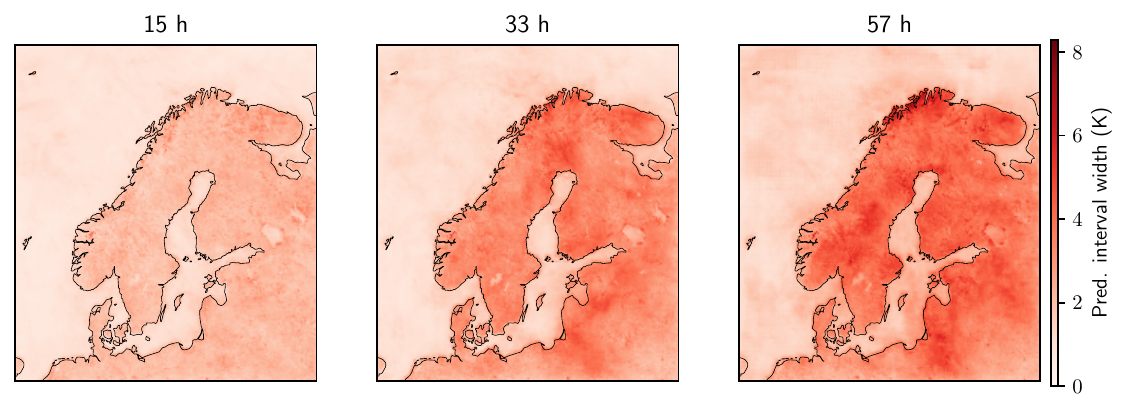}
    \end{subfloat}
    \caption{Prediction (top), Ground Truth (middle) and width of the error bars (bottom) at $\alpha = 0.05$ for predicting the temperature 2m above ground (\texttt{2t}) using \lammodelmse.}
    \label{fig:temp_2m}
\end{figure}

\cref{fig:temp_2m} shows the ground truth, predicted forecast and the conformalised error intervals for temperature 2m above ground. 
Considering the autoregressive nature of \lammodel, the error accumulates and grows further in time, which is accurately captured by the CP framework (refer \cref{fig:neurwp_iwidth_z}).

\begin{figure}[tbp]
    \centering
    \begin{subfigure}{0.5\textwidth}
        \centering
        \includegraphics[width=\linewidth]{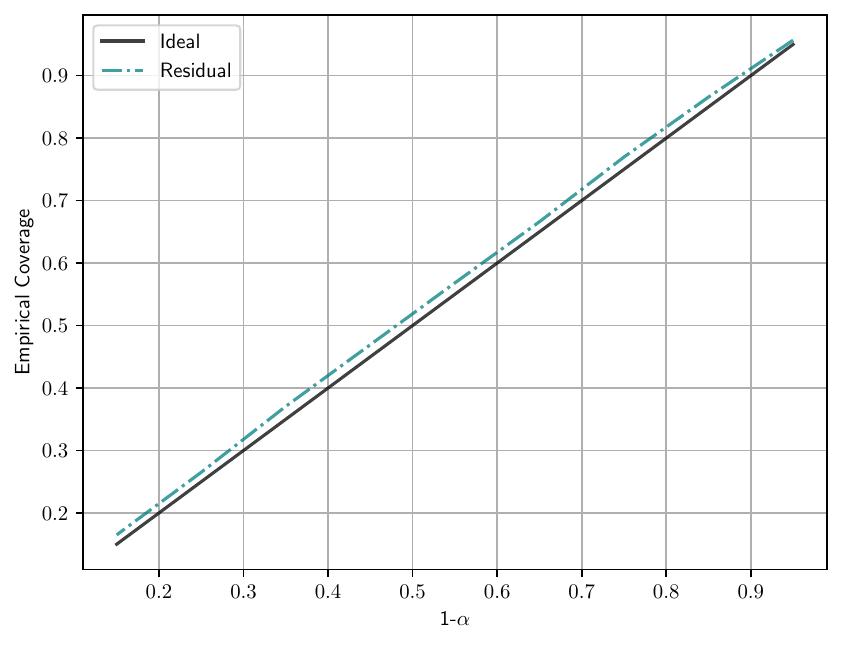}
        \caption{\lammodelmse}
    \end{subfigure}%
    \begin{subfigure}{0.5\textwidth}
        \centering
        \includegraphics[width=\linewidth]{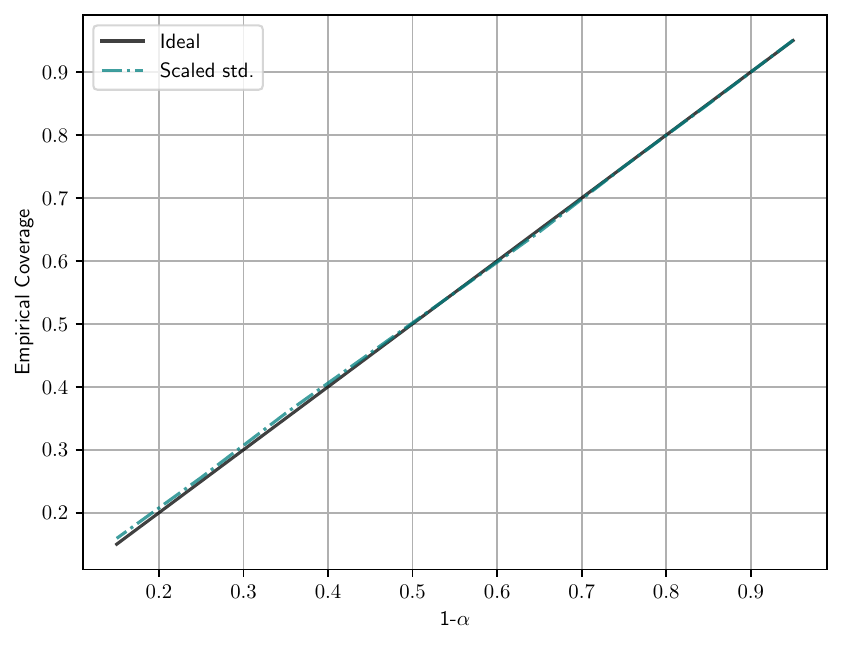}
        \caption{\lammodelnll}
    \end{subfigure}
    \caption{Empirical Coverage for weather forecasting models.}
    \label{fig:neurwp_emp_cov}
\end{figure}

\begin{figure}[tbp] 
    \centering
    \begin{subfigure}{\textwidth}
        \centering
        \includegraphics[width=\linewidth]{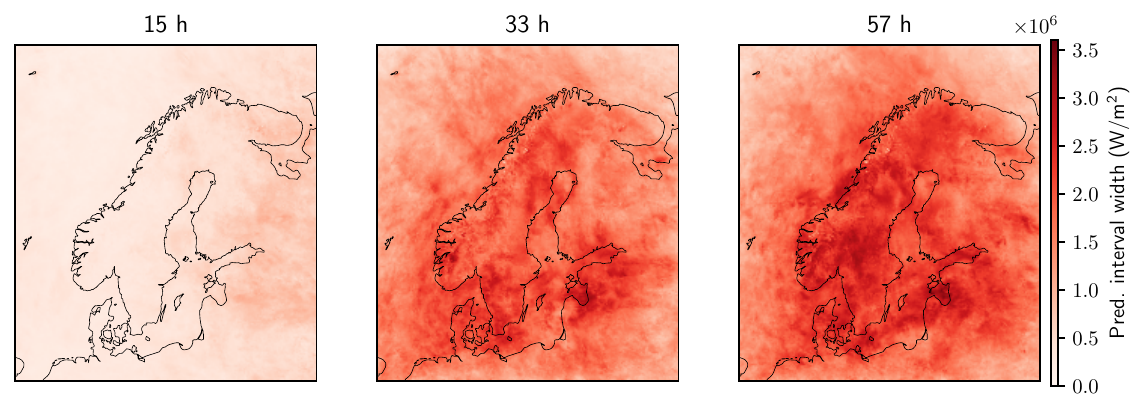}
        \caption{\lammodelmse}
        \label{fig:neurwp_iwidth_nswrs_mse}
    \end{subfigure}
    \begin{subfigure}{\textwidth}
        \centering
        \includegraphics[width=\linewidth]{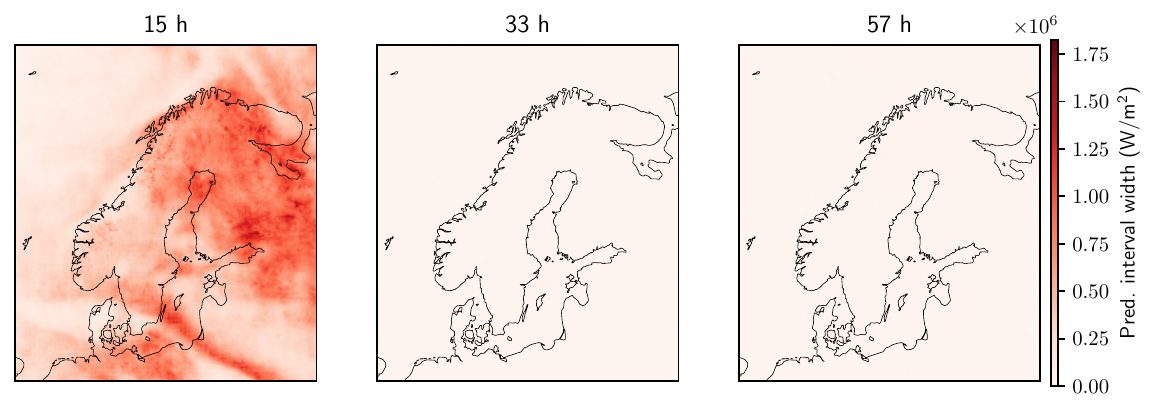}
        \caption{\lammodelnll}
        \label{fig:neurwp_iwidth_nswrs_nll}
    \end{subfigure}
    \caption{Width of predictive interval at $\alpha = 0.05$ for shortwave solar radiation (\texttt{nswrs}) in an example forecasts. Note that for \lammodelmse these widths are constant after calibration, while for \lammodelnll they depend on the predicted standard deviations for the specific forecast. This is most noticeable by \lammodelnll having very tight bounds during the night (lead times 33 h and 57 h), when the short-wave solar radiation is close to 0 and easy to predict.}
    \label{fig:neurwp_iwidth_nswrs}
\end{figure}

\begin{figure}[tbp]
    \begin{subfigure}{\textwidth}
        \centering
        \includegraphics[width=\linewidth]{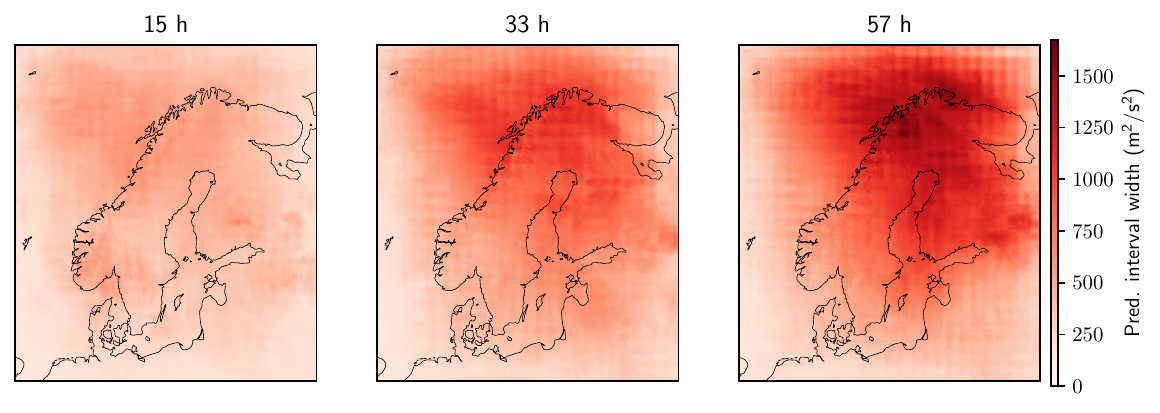}
        \caption{\lammodelmse}
        \label{fig:neurwp_iwidth_z_mse}
    \end{subfigure}
    \begin{subfigure}{\textwidth}
        \centering
        \includegraphics[width=\linewidth]{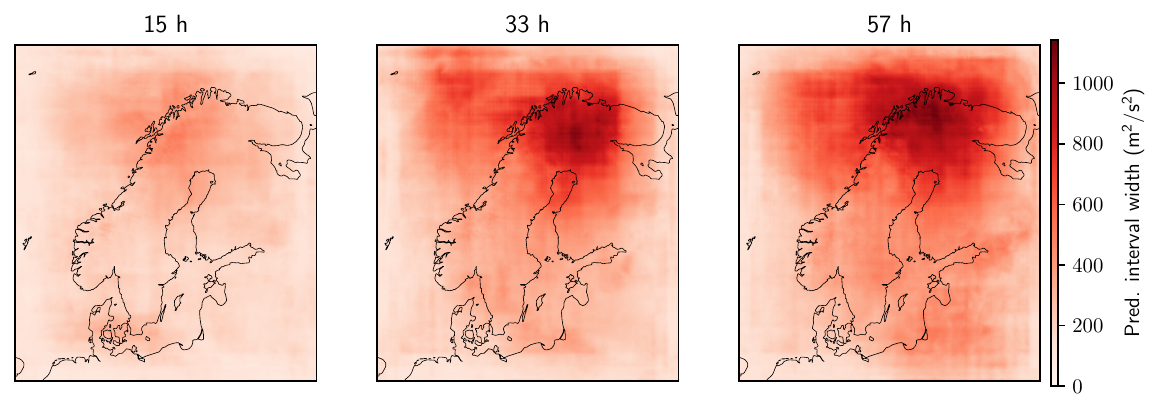}
        \caption{\lammodelnll}
        \label{fig:neurwp_iwidth_z_nll}
    \end{subfigure}
    \caption{
    Width of predictive interval at $\alpha = 0.05$ for geopotential at 500 hPa (\texttt{z500}). 
    Both models show a certain spatial pattern, especially for longer lead times.
    This pattern can be connected to how the GNN in \lammodel is defined over the forecasting area.
    }
    \label{fig:neurwp_iwidth_z}
\end{figure}

\Cref{fig:neurwp_emp_cov} shows the empirical coverage for the test set.
With CP, we can achieve calibrated uncertainty estimates for both versions of \lammodel.
Of great interest in the weather forecasting setting is the uncertainty for specific future time points.
We visualise this by plotting the width of the error bars for all spatial locations at different lead times in an example forecast.
Such plots for shortwave solar radiation are shown in \cref{fig:neurwp_iwidth_nswrs} and for geopotential in \cref{fig:neurwp_iwidth_z}.

\Cref{fig:neurwp_iwidth_nswrs} highlights an important difference between the two methods for computing non-conformity scores. 
As the shortwave solar radiation is close to 0 during the night, it is easy for the model to predict.
During the day, this is far more challenging.
With the AER non-conformity scores, used for \lammodelmse in \cref{fig:neurwp_iwidth_nswrs_mse}, the width of the predictive intervals is determined during calibration, and does not change depending on the forecast from the model.
As a specific lead time can fall both during the  day and night, depending on the initialisation time, CP will give large error bars also during the night.
This can be compared to the results for \lammodelnll in \cref{fig:neurwp_iwidth_nswrs_nll}, using STD non-conformity scores.
In this case, the bounds are very tight for lead times during the night (33~h and 57~h).
It can also be noted that for \lammodelnll at lead time 15, we see clear spatial features appearing in the error bars themselves.
This corresponds to higher forecast uncertainty in areas of rapid change.
The conditional dependency that emerges while using STD in \lammodelnll thus has desirable properties, but this relies on having a model that outputs (potentially uncalibrated) standard deviations.



\subsubsection{Global Forecasting}
We next experiment with CP for global weather forecasting.
The models used are again \lammodelmse and \lammodelnll, but here applied on the full globe.
These models are trained on a version of the ERA5 reanalysis dataset \citep{era5} using a 1.5\textdegree{} latitude-longitude grid.
The global models have 8 graph processing layers and use 256-dimensional latent representations.
Each forecast includes 5 surface variables and 6 atmospheric variables, each modelled at 13 different vertical pressure levels in the atmosphere.
Due to the large number of variables forecast (83 in total), we here only perform CP for a subset of these. 
This subset includes all surface variables and the atmospheric variables at pressure level 700 hPa.
This results in a total of $N_\text{var.} = 11$ variables, modelled on a $N_x \times N_y = 240 \times 121$ grid over $T = 40$ time steps (up to 10 days lead time with 6 h time steps).
We note that a strength of the CP framework is that uncertainty quantification can be performed per variable, alleviating memory issues during calibration.
Therefore, the procedure could trivially be extended to the full set of variables, as long as the full forecasts are stored.
We again refer to \cite{neural_lam} for more details about the global models and data configuration.

For the global experiment, we use full years of forecasts for calibration and evaluation, all starting from ERA5 as initial conditions.
Forecasts at 00 and 12 UTC each day of 2018 are used as the calibration set, and a similar set of forecasts for 2019 is used as the test set\footnote{We remove forecasts started during the last 10 days of 2018 from the calibration set to avoid strong correlations with the test data at the start of 2019. Note that forecasts in the test set started during the last days of 2019 will extend into time points in 2020.}.
The ground truth is given by ERA5 at each forecasted time point.
Using a full year for calibration allows for capturing the model performance across all different seasons.
This allows for calibrating the model once, and then using the computed $\hat{q}$ values for the full next year of forecasts.
However, any distributional shift due to climate variations from one year to the next remains.
We note that in practic,e this does not seem to cause any major issue for achieving the desired coverage.

\begin{figure}[tb]
    \centering
    \begin{subfloat}
        \centering
        \includegraphics[width=\linewidth]{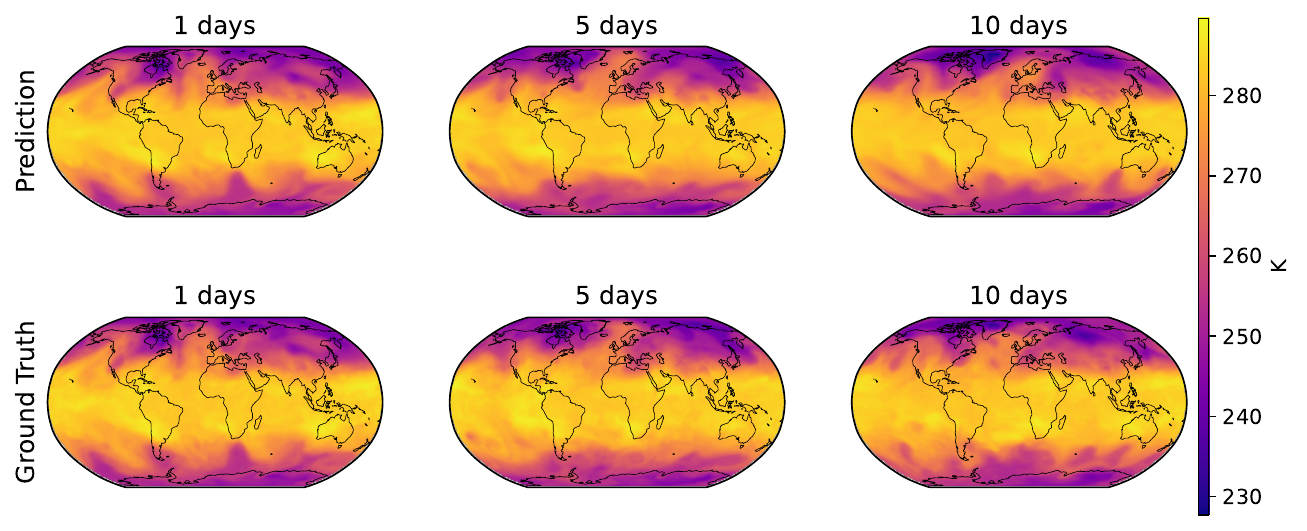}
    \end{subfloat}
    \begin{subfloat}
        \centering
        \hspace{0.04\linewidth}%
        \includegraphics[width=0.87\linewidth]{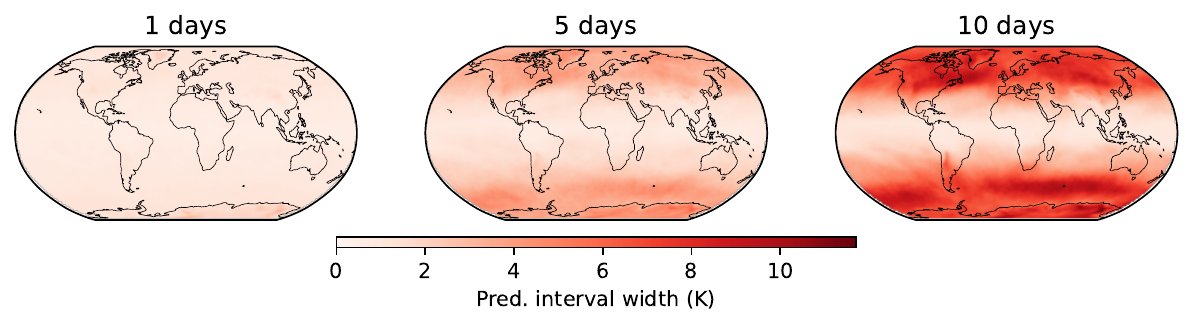}%
        \hspace{0.09\linewidth}%
    \end{subfloat}
    \caption{Prediction (top), Ground Truth (middle) and width of the error bars (bottom) at $\alpha = 0.15$ for predicting the temperature at 700 hPa (\texttt{t700}) using \lammodelmse.}
    \label{fig:global_example_pred_t700}
\end{figure}

\Cref{fig:global_example_pred_t700} shows an example prediction from \lammodelmse and corresponding error bars.
Global forecasting up to 10 days is a more challenging task than the limited area modelling up to 57 h.
We see that at 10 days the model prediction fails to capture much of the patterns in the ground truth data.
Importantly, this is accurately captured in the error bars, which increase with the lead time to high values at 10 days.

\begin{figure}[tbp]
    \begin{subfigure}{\textwidth}
        \centering
        \includegraphics[width=\linewidth]{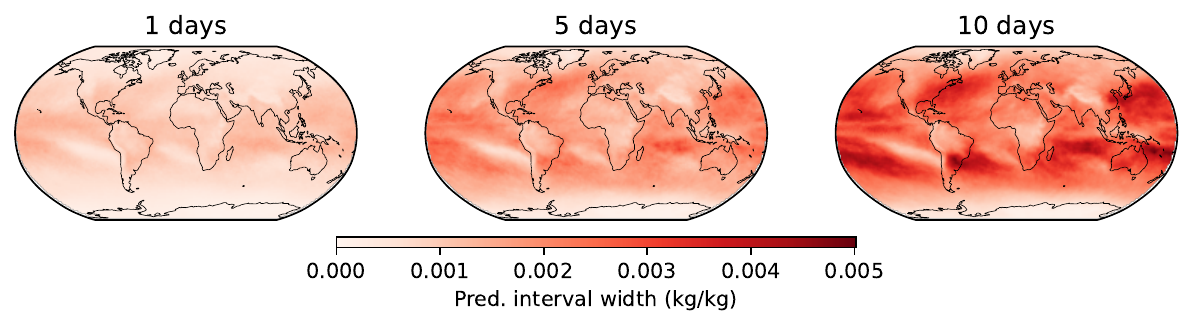}
        \caption{\lammodelmse}
        \label{fig:global_neurwp_iwidth_q_mse}
    \end{subfigure}
    \begin{subfigure}{\textwidth}
        \centering
        \includegraphics[width=\linewidth]{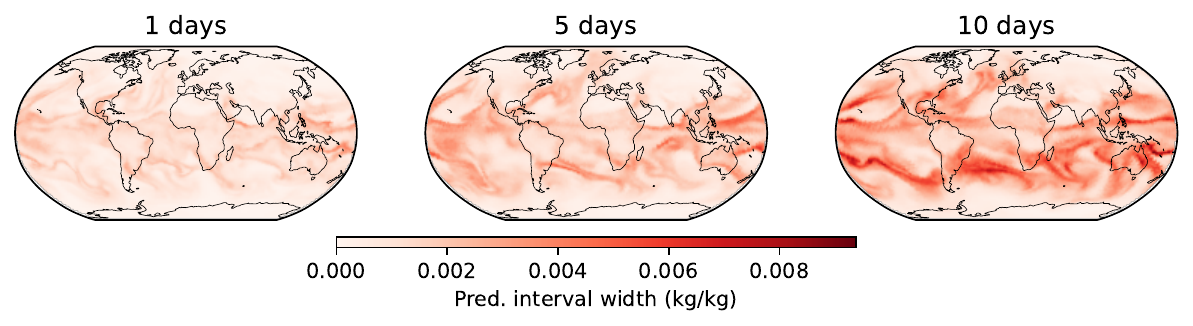}
        \caption{\lammodelnll}
        \label{fig:global_neurwp_iwidth_q_nll}
    \end{subfigure}
    \caption{
    Width of predictive interval at $\alpha = 0.15$ for specific humidity at 700 hPa (\texttt{q700}).
    }
    \label{fig:global_neurwp_iwidth_q}
\end{figure}

\begin{figure}[tbp]
    \begin{subfigure}{\textwidth}
        \centering
        \includegraphics[width=\linewidth]{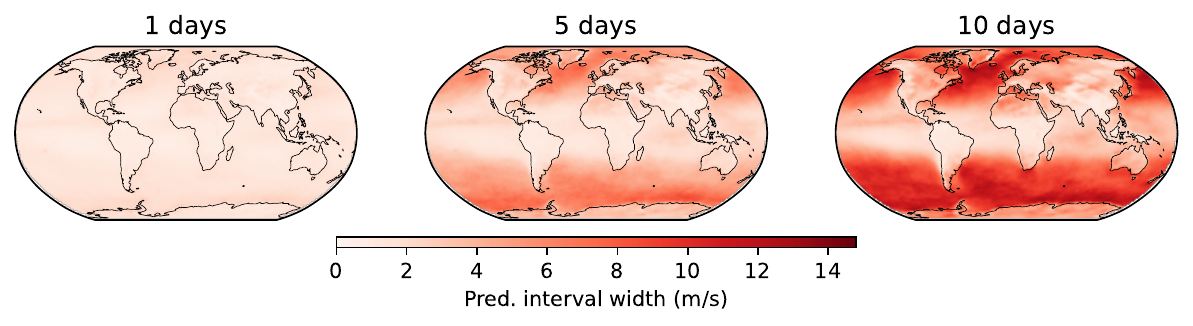}
        \caption{\lammodelmse}
        \label{fig:global_neurwp_iwidth_u_mse}
    \end{subfigure}
    \begin{subfigure}{\textwidth}
        \centering
        \includegraphics[width=\linewidth]{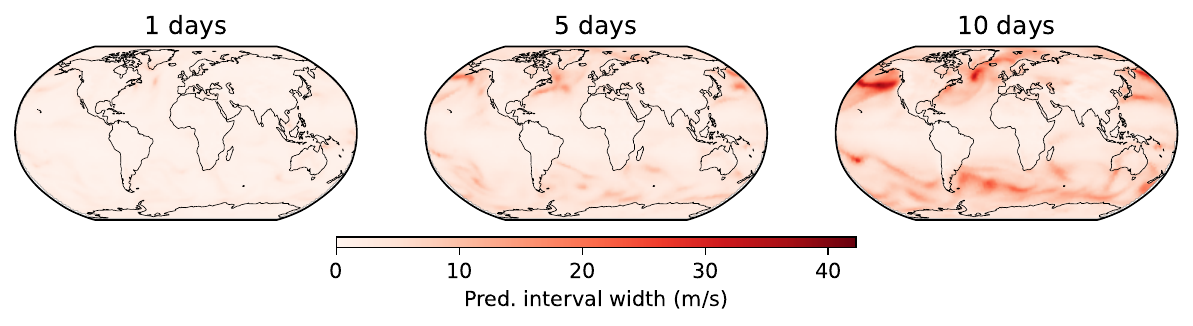}
        \caption{\lammodelnll}
        \label{fig:global_neurwp_iwidth_u_nll}
    \end{subfigure}
    \caption{
    Width of predictive interval at $\alpha = 0.15$ for $u$-component of wind at 10 m above ground (\texttt{10u}). 
    }
    \label{fig:global_neurwp_iwidth_u}
\end{figure}

In \cref{fig:global_neurwp_iwidth_q,fig:global_neurwp_iwidth_u} we plot the width of the error bar for specific humidity and wind.
Similar to the limited area case, we note for \lammodelnll the error bars corresponds to patterns in the forecast itself, due to the use of predicted standard deviations from the model.
For \lammodelmse the plots instead highlight the regions where predictions are more challenging in general, across all forecasts.
Additional plots from the weather forecasting experiments are given in \cref{sec:weather_appendix}.

As for all experiments, we include results for the weather forecasting models in \cref{table: coverage_all_models}.
In both the global and limited area setting CP successfully produces calibrated error bars.
For the \lammodelnll model the original standard deviations output by the model are too low, leading to invalid error bars and insufficient coverage. 
After applying CP however the error bars are well calibrated.
We generally see that the \lammodelnll has tighter error bars than \lammodelmse.
This can be attributed to these being input-dependent, specific to each forecast from the model.


\subsubsection{Discussion on Exchangeability}

\begin{figure}[t]
    \centering
    \includegraphics[width=\linewidth]{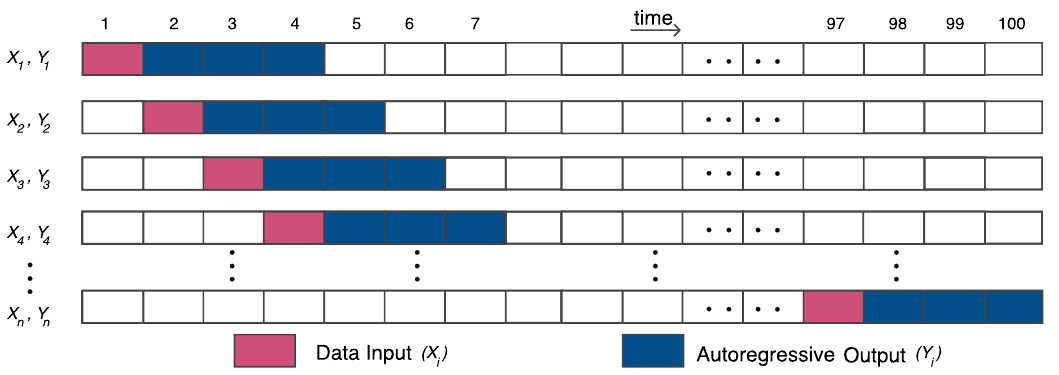}
    \caption{Constructing exchangeable input-output pairs from a time series dataset. The calibration dataset is arbitrarily characterised by a spatio-temporal dataset spanning 100 time instances, where the model takes in the field at a day and autoregressively predicts for the next 3 time instances. Since the model represents an initial boundary value problem, we are able to break each input-output ($X_i,Y_i$) as an exchangeable data point, essentially sampled from a distribution spanning the entire month.}
    \label{fig:time_window_ibvp}
\end{figure}

Traditionally, the CP framework is limited in application to time-series modelling as it fails the exchangeability assumption. Previous research has looked into fixing this violation of exchangeability by accounting for the distribution shift using weighted conformal techniques  \citep{tibshirani2019conformal}, but becomes limited in application in multi-variate settings. Other work has explored CP for multi-variate time series forecasting, where each time-series is treated as an exchangeable observation \citep{conformaltimeserires}. Within the weather modelling tasks outlined in this section, we maintain exchangeability by treating each modelling task as an Initial Boundary Value Problem (IBVP, boundary given by the forcing terms in \cref{eq:forecasting_ar}). As given in \cref{eq:forecasting_ar}, the model takes in the initial conditions, $\mathcal X \in \mathbb{R}^{T_{\text{in}=1} \times N_x \times N_y \times N_{var}} $ and is auto-regressively rolled out $T_{out}$ steps to obtain the output $\mathcal Y \in \mathbb{R}^{T_{\text{out}} \times N_x \times N_y \times N_{var}} $. Being dependent on the initial conditions alone and being rolled out for a fixed number of steps, each input-output pair as mathematically outlined in \cref{subsec: formal_definition} and visually represented in \cref{fig:time_window_ibvp} can be treated as an exchangeable pair. We are allowed to make this assumption on exchangeability since the model is agnostic to the temporal nature of the dataset beyond the autoregressive roll-out of each forward prediction, typical of an initial boundary value problem i.e. the neural weather forecast starting from 18:00 $3^{rd}$ January is independent of the forecast made using the neural weather models starting at 12:00 $1^{st}$ January. Here the calibration dataset is seen as samples from an extremely large distribution which effectively characterises the entirety of the entire month/year under consideration. Thus, by combining our preservation of spatio-temporal structure as outlined in \cref{subsec: formal_definition} and by treating the neural weather models as initial boundary value problems, we are able to maintain exchangeability across the calibration datasets, allowing us to perform CP. 

Though the above description discusses about exchangeability across the calibration dataset, it does not extend across to the prediction set. For each of the experiments within the limited area and global weather forecasting, we assume that the climate does not vary significantly across the years under consideration for the calibration and prediction sets. 


\newpage
\subsection{Camera Diagnostic on a Tokamak}
\label{sec: fno_camera}

The Mega-Ampere Spherical Tokamak (MAST) at the UK Atomic Energy Authority was equipped with fast Photron camera diagnostics to capture plasma evolution in the visible spectrum in real-time. These cameras have been instrumental in understanding plasma phenomena \citep{Kirk2006}, providing statistical insights into plasma turbulence \citep{Walkden2022} and disruptions \citep{Ham2022}.

Building on our previous work \citep{Gopakumar_2024}, we apply the CP framework to an FNO trained to forecast plasma evolution from camera imagery. The model takes 10 consecutive camera frames as input and predicts the subsequent 10 frames. Using the AER nonconformity score, we demonstrate that CP provides statistically valid error bars for these predictions. Details on the camera data, FNO architecture, and training are available in \citep{Gopakumar_2024}.

\subsubsection{Exchangeability and Coverage Validation}

Although the FNO predicts plasma evolution over an entire shot duration, we structure the problem as an initial value problem where each forecast depends solely on its initial 10 frames. As illustrated in \cref{fig:time_window_ibvp}, this allows us to treat each input-output pair as exchangeable: the model predicts the entire spatio-temporal output tensor simultaneously rather than sequentially, preserving the temporal structure.

The validity of CP for this application rests on three key properties: (1) only the calibration dataset and prediction $X_{n+1}$ must be exchangeable, (2) we predict error bars over the complete time interval simultaneously, not autoregressively, and (3) we assume minimal distributional shift between calibration and prediction shots. This final assumption requires that calibration and prediction shots exhibit similar plasma discharge profiles and device conditions.

\begin{figure}
    \centering
    \includegraphics[width=0.7\linewidth]{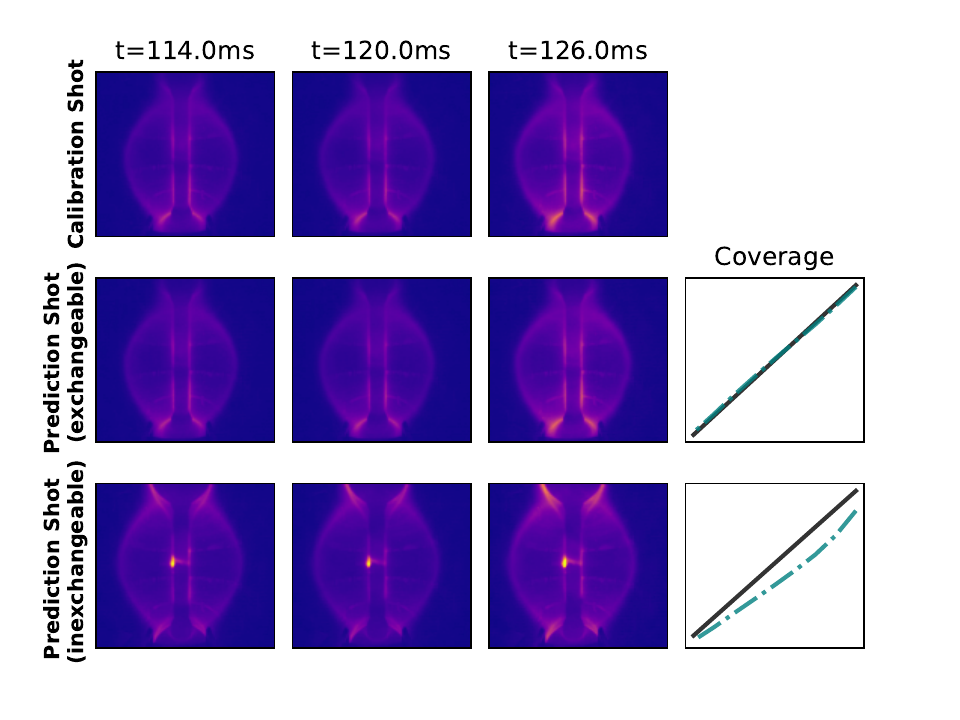}
    \caption{Coverage comparison for exchangeable (similar plasma profiles) versus non-exchangeable (dissimilar plasma profiles) prediction shots relative to the calibration set. Violations of exchangeability lead to insufficient coverage.}
    \label{fig:camera_coverage_diff_shots}
\end{figure}

To assess sensitivity to exchangeability violations, we compare coverage across shots with similar versus dissimilar plasma profiles relative to the calibration set. \Cref{fig:camera_coverage_diff_shots} demonstrates that predictions on shots with substantially different characteristics suffer from insufficient coverage, highlighting the importance of the exchangeability assumption across the calibration and prediciton set. The calibration shot is indicative of a L-mode of confinement within the plasma, where as the inexchangeable prediciton shot shows H-mode of confinement characterising a vastly different physics regime \citep{Howlett_2023}. When this assumption holds, CP provides exact coverage as shown in \cref{appendix_camera}. \Cref{fig:fno_camera} visualises the calibrated error bars for plasma evolution forecasts across the tokamak central solenoid, demonstrating 50\% coverage ($\alpha=0.5$).

\begin{figure}
    \centering
    \includegraphics[width=0.7\linewidth]{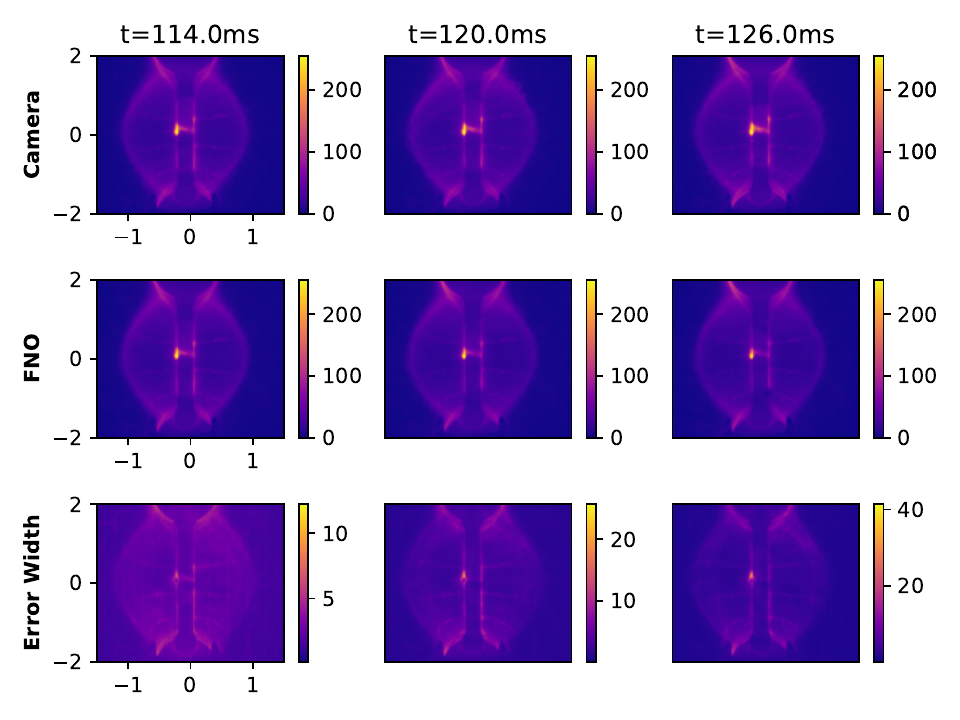}
    \caption{Camera observations (top), FNO predictions (middle), and width of prediction intervals obtained using CP with $\alpha=0.5$ corresponding to 50\% coverage (bottom).}
    \label{fig:fno_camera}
\end{figure}




\section{Discussion}
\label{Discussion}
We have demonstrated that conformal prediction provides a practical, theoretically grounded approach to uncertainty quantification for surrogate models across diverse spatio-temporal applications. Through comprehensive empirical evaluation spanning PDEs, fusion diagnostics, and weather forecasting, CP delivers statistically guaranteed marginal coverage regardless of model architecture, training regime, or output dimensionality. However, understanding both its capabilities and inherent limitations is essential for responsible application in scientific domains. Below, we discuss CP's key strengths for practical deployment, followed by an honest assessment of its limitations and implications for real-world use cases

\subsection{Strengths}

In safety-critical applications such as fusion reactor design, climate modelling, and engineering optimisation, surrogate models must provide credible uncertainty estimates alongside their predictions \citep{Begoli2019}. Conformal prediction addresses this need by offering statistical guarantees for uncertainty quantification with several key advantages:

\noindent\textbf{Statistical Guarantees.} CP provides provable marginal coverage (\cref{eq: coverage}) regardless of model architecture, training regime, or output dimensionality. This validity is particularly crucial when surrogate models transition from research to production environments where retraining opportunities are limited.

\noindent\textbf{Model-Agnostic and Scalable.} The framework requires no architectural modifications or knowledge of training procedures, enabling application to pre-trained models. Our experiments demonstrate guaranteed coverage across outputs spanning up to 20 million dimensions (weather forecasting) with near-zero calibration costs, effectively circumventing the curse of dimensionality.

\noindent\textbf{Computational Efficiency.} Unlike ensemble methods or Bayesian approaches requiring extensive sampling, CP calibration is computationally trivial (\cref{sec: comp_complex}). Calibration times range from seconds for low-dimensional problems to minutes for high-dimensional applications, performed on standard hardware without specialised computational resources.

\noindent\textbf{Practical Utility.} CP enables rigorous validation of a surrogate model's usefulness for specific downstream applications, providing actionable uncertainty estimates for decision-making in risk-averse scenarios.

\subsection{Limitations}

While CP offers substantial benefits, several inherent limitations must be acknowledged for responsible application in scientific domains. We discuss these limitations, their practical implications, and potential mitigation strategies.

\noindent\textbf{Marginal vs. Conditional Coverage.} The coverage guarantee in \cref{eq: coverage} provides \textit{marginal coverage}—validity averaged over all predictions—rather than the more desirable \textit{conditional coverage}:
\begin{equation}
    \mathbb{P}(Y_{n+1}\in \mathbb{C}^{\alpha} | X_{n+1}) \geq 1 - \alpha.
\end{equation}
In practice, this means that while 90\% of predictions will be covered on average, any individual prediction may have substantially different actual coverage. Although conditional coverage cannot be guaranteed in general, approximations exist \citep{vovk2012conditional}.

Our formulation guarantees coverage cell-wise across the spatio-temporal tensor but does not provide joint coverage across the entire prediction domain. Extensions to joint coverage exist \citep{diquigiovanni2021conformal, messoudiCopula2021, messoudiEllipsoidal2022} but fail to scale with dimensionality, limiting their applicability to the high-dimensional problems considered here.

\textit{Practical Impact:} In risk-critical scenarios requiring reliable bounds for specific predictions (e.g., fusion disruption avoidance, extreme weather events), marginal coverage may be insufficient. Users must understand that individual predictions, particularly in distribution tails, may not achieve the target coverage. Our experiments on exchangeability violations (\cref{sec: fno_camera}, \cref{fig:camera_coverage_diff_shots}) demonstrate this sensitivity.

\textit{Mitigation:} Covariate shifting methods \citep{tibshirani2019conformal} can improve conditional coverage when probability densities of calibration and deployment distributions are known or estimable. However, density estimation becomes unreliable in high dimensions \citep{dataset_shift_book}. For critical applications, we recommend conservative $\alpha$ values and validation on held-out data similar to deployment conditions.

\noindent\textbf{Data Requirements and Exchangeability.} CP requires calibration data that is exchangeable with the prediction regime. The quality of coverage guarantees follows a beta distribution (\cref{eq: cov_dist}) governed by calibration set size. Our empirical study (\cref{calibration_size}) shows that $n_{\text{cal}} \geq 1000$ typically provides reliable coverage.

\textit{Practical Impact:} For experimental data (weather, fusion diagnostics), obtaining sufficient exchangeable calibration data can be challenging. The exchangeability assumption is particularly delicate for time-series data where distribution shifts are common. As demonstrated in \cref{fig:camera_coverage_diff_shots}, violations lead to invalid coverage. Weather experiments (\cref{sec:nwp}) assume minimal climate variation between calibration (2021) and prediction (2022) years—an assumption that may not hold for long-term climate change scenarios or extreme events outside the calibration distribution.

\textit{Mitigation:} When calibration data is expensive (e.g., fusion experiments), fine-tuning scenarios offer a natural solution: training data doubles as calibration data since both represent the target distribution. For time-series applications, careful validation of exchangeability assumptions and sensitivity analyses (as performed for camera diagnostics) is essential.

\noindent\textbf{Prediction Sets vs. Distributions.} CP provides prediction \textit{sets} (intervals) rather than full probability distributions. While Bayesian methods offer distributions that can be propagated via Monte Carlo sampling or used for risk calculations, CP intervals lack this flexibility.

Recent work \citep{cella2022validity} provides imprecise probabilistic interpretations of CP, enabling uncertainty propagation \citep{balch2012mathematical, hose2021universal}. However, these methods are not yet widely adopted. It is worth noting that while Bayesian posteriors appear more informative, the false-confidence theorem \citep{martin2019false} shows that precise probability models can assign high confidence to low-probability events.

\noindent\textbf{Input Independence in Deterministic Models.} For deterministic models using AER nonconformity scores, the error bars ($\sim \hat{q}$) are fixed during calibration and do not vary with inputs. As shown in \cref{fig:mhd_mpp_finetune}, this produces global error estimates rather than input-specific bounds. This limitation is less severe for probabilistic models (STD scores) where prediction sets scale with model-predicted standard deviations (\cref{fig:cell-wise-cp}), providing a weak conditionining on the input. 

\textit{Practical Impact:} Deterministic CP may produce overly conservative bounds in easy regions and insufficient bounds in challenging regions. This is particularly evident in foundation models trained on diverse data, where local complexity varies substantially.

\textit{Mitigation:} When possible, use probabilistic models or ensemble approaches that provide initial uncertainty estimates for STD-based CP. Alternatively, recent work on normalised CP \citep{normalisedCP2021} offers more input-adaptive bounds, though at the cost of increased computational complexity or architectural modifications.

\noindent\textbf{Spatial and Temporal Correlations.} Our cell-wise calibration treats each spatio-temporal point independently, ignoring correlations between adjacent cells. This is a significant limitation for PDE-based systems where solutions exhibit strong spatial and temporal dependencies.

\textit{Practical Impact:} While we achieve marginal coverage at each cell, the framework does not capture or leverage the correlation structure inherent in physical systems. This may lead to overly conservative joint coverage across regions or miss spatially coherent error patterns.

\textit{Mitigation:} We implicitly rely on the surrogate model to learn spatial dependencies during training. Future extensions could incorporate spatial or temporal correlation structures into the CP framework, leading to joint coverage as we have explored in \citet{gopakumar2025calibrated}. 

\subsection{Broader Context}

Despite these limitations, CP offers a valuable and practical tool for UQ in scientific machine learning. The guaranteed marginal coverage, model-agnostic nature, and computational efficiency make it particularly suitable for validating pre-trained surrogate models before deployment. When combined with careful validation of assumptions (particularly exchangeability), CP provides actionable uncertainty quantification that can guide decision-making in safety-critical applications.

For optimal results, we recommend: (1) validating exchangeability assumptions through sensitivity analyses, (2) using conservative $\alpha$ values for risk-critical applications, (3) preferring probabilistic models or ensembles when input-dependent bounds are crucial, and (4) maintaining awareness of the marginal nature of coverage guarantees when interpreting individual predictions.

\section{Conclusion}
\label{conclusion}

This paper presents a comprehensive empirical study demonstrating that conformal prediction provides statistically guaranteed uncertainty quantification for surrogate models across diverse scientific applications. By maintaining exchangeability of spatio-temporal data through preservation of tensorial structure, we achieve valid error bars satisfying \cref{eq: coverage} for outputs spanning up to 20 million dimensions.

\noindent\textbf{Key Contributions.} Our work establishes that CP can be applied to any pre-trained or fine-tuned surrogate model—regardless of architecture (MLP, U-Net, FNO, ViT, GNN), training regime, or output dimensionality—to obtain guaranteed marginal coverage with near-zero computational cost. We benchmark three nonconformity scores (CQR, AER, STD) across both deterministic and probabilistic models, demonstrating consistent coverage across applications ranging from fundamental PDEs to operational weather forecasting and fusion diagnostics. 

Critically, we show that CP provides valid prediction sets even for out-of-distribution scenarios where models are deployed on physics regimes different from their training distribution (wave equation at half-speed, Navier-Stokes at different viscosity, pre-trained foundation models on new physics). This capability is essential for validating surrogate model utility in production environments where retraining is infeasible.

\noindent\textbf{Practical Impact.} For scientific machine learning practitioners, our framework offers a rigorous method to assess whether a pre-trained model is suitable for a specific downstream application. The guaranteed coverage enables confident deployment of surrogate models in safety-critical contexts—from fusion reactor control to extreme weather response—where uncertainty quantification is imperative but computational budgets preclude ensemble methods or extensive Bayesian inference.

\noindent\textbf{Scope and Limitations.} While we focus on spatio-temporal data, our methodology extends to any models producing fixed tensorial outputs with exchangeable calibration and prediction regimes. However, users must carefully validate exchangeability assumptions, particularly for time-series and experimental data where distribution shifts are common. As discussed in \cref{Discussion}, the marginal nature of coverage guarantees, cell-wise independence assumptions, and input independence in deterministic models represent important limitations that practitioners should consider when applying CP to their specific problems.

\noindent\textbf{Future Directions.} Extensions to conditional coverage, incorporation of spatial-temporal correlation structures, and methods for handling systematic exchangeability violations remain important open problems. The integration of CP with recent advances in foundation models for scientific computing presents particularly promising opportunities for scalable, trustworthy uncertainty quantification.

\noindent\textbf{Reproducibility.}  All code, data generation scripts, and trained models are publicly available at \url{https://github.com/gitvicky/Spatio-Temporal-UQ}.

\section{Acknowledgements}
\label{acknowledgements}
The authors would like to thank Anima Anandkumar and Zongyi Li from Caltech for their help with defining neural operators and extending them to complex physics cases. The authors would also like to thank Michael McCabe at the Flatiron Institute for his help in setting up the Multi-Physics Pretrained foundation physics model and extending it to a Fusion-relevant database. 
This work was performed using resources provided by the \href{(www.csd3.cam.ac.uk)}{Cambridge Service for Data Driven Discovery (CSD3) operated by the University of Cambridge Research Computing Service}, provided by Dell EMC and Intel using Tier-2 funding from the Engineering and Physical Sciences Research Council (capital grant EP/T022159/1), and \href{www.dirac.ac.uk}{DiRAC funding from the Science and Technology Facilities Council}.
This work was supported under project ID a122 as part of the Swiss AI Initiative, through a grant from the ETH Domain and computational resources provided by the Swiss National Supercomputing Centre (CSCS) under the Alps infrastructure. It was also supported by the Excellence Center at Linköping--Lund in Information Technology (ELLIIT).
Computations were enabled by the Berzelius resource at the National Supercomputer Centre, provided by the Knut and Alice Wallenberg Foundation. This work has been (part-) funded by the EPSRC Energy Programme [grant number EP/W006839/1].  To obtain further information on the data and models underlying this paper, please contact PublicationsManager@ukaea.uk*. 

\newpage
\bibliography{bibliography}

\appendix
\newpage
\section{Poisson Equation}
\label{appendix_poisson}
The Poisson Equation in one-dimension takes the form: 
\begin{equation}
\label{eq: poisson_1d}      
    \pdv[2]{u}{x} = \rho,   \quad  x  \in [0,1],
\end{equation}
where $u$ defines the field value, $x$ the spatial domain, and $\rho$ the density of the source. 

The Poisson equation is solved with a finite difference scheme using the \textit{py-pde} python package \citep{py-pde}. \Cref{eq: poisson_1d} is constructed as an initial-value problem, where a scalar uniform field is initialised across the domain and evolved until convergence. A dataset comprising different instances of the 1D Poisson equation is constructed by sampling for different initial values uniformly from within the domain:  $u_{\text{init}}  \in [0, 4) $. 

A total of 7,000 data points are generated, where 5,000 are used to train an MLP with 3 layers and 64 neurons in each layer, 1,000 are used to perform the calibration required to estimate the nonconformity scores and another 1,000 for validation. Being a steady-state problem, the MLP learns how a scalar field evolves under the influence of the Laplacian, mapping from the initial to the final state of evolution. The network learns to map the initial condition to the final steady-state solution. 

Each MLP is trained to take in the scalar initial field along the 32-point spatial domain to output the final field at the steady state. 
For the case of STD, the architecture is modified with 1D dropout layers. Each model is trained for up to 1000 epochs using the Adam optimiser \cite{adam} with a step-decaying learning rate. The learning rate is initially set to 0.005 and scheduled to decrease by half after every 100 epochs. The model was trained using a quantile loss for the case of CQR and MSE loss in all other cases. 

\newpage
\section{Convection-Diffusion Equation}
\label{appendix_CD}
\subsection{Physics}
Consider a modified version of the one-dimensional convection-diffusion equation used to model the transport of a fluid:
\begin{align}
\label{eq: conv_diff}
&\pdv{u}{t} = D\pdv[2]{u}{x} + u\pdv{D}{x} - c\pdv{u}{x}, \quad x \ \in\ [0,10],\; t\ \in\ [0, 0.1] \\
&u(x, t=0) = \exp\left(-\tfrac{(x-\mu^2)}{2\sigma^2}\right).
\end{align}
Here $u$ defines the density of the fluid, $x$ the spatial coordinate,  $t$ the temporal coordinate,  $D$ the diffusion coefficient, and $c$ the convection velocity. The initial condition is parameterised by $\mu$ and $\sigma^2$, representing the mean and variance of a Gaussian distribution. The system is bounded by a no-flux boundary condition.

The numerical solution for the above equation is built using a Newtonian solver with a forward time centered space implementation in Python.  We construct a dataset by  Latin hypercube sampling across parameters $D, c, \mu, \sigma$. Each parameter is sampled from within the domain given in \cref{table: data_generation_convdiff} to generate 3,000 simulation points, each with its own initial condition, diffusion coefficient and convection velocity. We generate another 2,000 data points, 1,000 each for the calibration and procuring of the prediction sets. These datasets are built by sampling across a different domain of the diffusion coefficient and convection velocity, different from that used for training; see \cref{table: data_generation_convdiff_2} for details. We use a one-dimensional U-Net to model the evolution of the convection-diffusion equation. The U-Net learns to perform the mapping from the first 10 time instances to the next 10 time instances, learning across the different field parameters and initial conditions. A more detailed physics description and the training set-up of the model can be found in \cref{appendix_CD}. 

As discussed in section \ref{conv_diff}, the dataset is built by solving the one-dimensional Convection Diffusion equation numerically. The physics of the equation, given by the various coefficients, is sampled from a certain range as given in \cref{table: data_generation_convdiff}. Each datapoint, as in each simulation, is generated with different Diffusion coefficients and wave velocities as described in section \ref{conv_diff}. Each simulation is run for 100 time iterations with a $\Delta t = 0.0005$ across a spatial domain spanning [0,10], uniformly discretised into 200 spatial units in the x-axis. Once the simulations are run and the dataset is generated, we downsample the temporal discretisation from 100 to 20 by taking every 5th time step. The sampling parameters governing the PDE solutions used for the training are given in \cref{table: data_generation_convdiff} and that used for the calibration and prediction is given in \cref{table: data_generation_convdiff_2}. 


\begin{table}[h!]
\caption{ Domain range and sampling strategies across the coefficients and initial condition parameters for building the training dataset for the 1D convection-diffusion equation. } 
\label{table: data_generation_convdiff}
  \centering
  \begin{tabular}{lll}
  \toprule
  Parameter & Domain & Type \\
    \midrule
        Diffusion Coefficient $(\alpha)$ & $[\sin(\frac{x}{\pi}), \sin(\frac{x}{2\pi})]$ & Continuous  \\
    Convection velocity $(\beta)$ & $[0.1, 0.5]$ & Continuous \\
    Mean $(\mu)$ & $[1.0, 8.0]$ & Continuous \\
    Variance $(\gamma)$ & $[0.25, 0.75]$ & Continuous \\
    \bottomrule
    \end{tabular}
\end{table}
\subsection{Model and Training}

We train a U-Net to map the spatio-temporal evolution of the field variable, taking in the first 20 time instances (\textit{T\_in}) to the next 10 time instances (\textit{T\_out}). For the case of the Convection-Diffusion Equation, we don't deploy an auto-regressive structure but perform a mapping from the initial distribution to the later distribution. The U-Net architecture can be found in \cref{table: unet_arch_cd}. For the case of STD, the architecture is modified with 1D dropout layers following each encoder and decoder of the U-Net. Though the values governing the evolution of Convection-Diffusion are relatively small, for better representation, we normalise the value with a linear range scaling, allowing the field values to lie between -1 and 1. Each model is trained for up to 500 epochs using the Adam optimiser \cite{adam} with a step-decaying learning rate. The learning rate is initially set to 0.005 and scheduled to decrease by half after every 100 epochs. The model was trained using a quantile loss for the case of CQR and an MSE loss in all other cases. 

\begin{table}[h!]
\caption{ Domain range and sampling strategies across the coefficients and initial condition parameters for building the calibration and prediction datasets for the 1D convection-diffusion equation. } 
\label{table: data_generation_convdiff_2}
  \centering
  \begin{tabular}{lll}
  \toprule
  Parameter & Domain & Type \\
    \midrule
    Diffusion Coefficient $(\alpha)$ & $[\sin(\frac{x}{2\pi}), \sin(\frac{x}{4\pi})]$ & Continuous  \\
    Convection velocity $(\beta)$ & $[0.5, 1,0]$ & Continuous \\
    Mean $(\mu)$ & $[1.0, 8.0]$ & Continuous \\
    Variance $(\gamma)$ & $[0.25, 0.75]$ & Continuous \\
    \bottomrule
    \end{tabular}
\end{table}

\begin{table}[ht]
  \caption{Architecture of the 1D U-Net deployed for modelling 1D Convection-Diffusion Equation}
  \label{table: unet_arch_cd}
  \centering
  \begin{tabular}{lll}
    \toprule
    Part     & Layer     &  Output Shape \\
    \midrule
    Input & - & (50, 20, 200) \\    
    Encoder 1 & \texttt{Conv1d/BatchNorm1d/Tanh} & (50, 32, 200) \\
    Pool 1 & \texttt{MaxPool1d} & (50, 32, 200)\\
    Encoder 2 & \texttt{Conv1d/BatchNorm1d/Tanh} & (50, 64, 100) \\
    Pool 2 & \texttt{MaxPool1d} & (50, 64, 100)\\
    Encoder 3 & \texttt{Conv1d/BatchNorm1d/Tanh} & (50, 128, 50) \\
    Pool 3 & \texttt{MaxPool1d} & (50, 128, 50)\\
    Encoder 4 & \texttt{Conv1d/BatchNorm1d/Tanh} & (50, 256, 25) \\
    Pool 4 & \texttt{MaxPool1d} & (50, 256, 25\\
    Bottleneck & \texttt{Conv1d/BatchNorm1d/Tanh} & (50, 512, 12) \\
    Decoder 4 & \texttt{ConvTranspose1d/Encoder 4} & (50, 256, 25) \\
    Decoder 3 & \texttt{ConvTranspose1d/Encoder 3} & (50, 128, 50) \\
    Decoder 2 & \texttt{ConvTranspose1d/Encoder 2} & (50, 64, 100) \\
    Decoder 1 & \texttt{ConvTranspose1d/Encoder 1} & (50, 32, 200) \\
    Rescale  & \texttt{Conv1d} & (50, 10, 200) \\
    \bottomrule
  \end{tabular}
\end{table}

\clearpage
\newpage
\section{Wave Equation}
\label{appendix_wave}

\subsection{Physics}
Consider the two-dimensional wave equation
\begin{align}
\label{eq: wave}
    &\pdv[2]{u}{t} = c^2 \bigg(\pdv[2]{u}{x} + \pdv[2]{u}{y}\bigg) = 0 , \quad x,y \in [-1,1],\ t \in [0, 1]\\
    &u(x,y,t=0) = \exp \left(-\alpha((x-\beta)^2 + (y-\gamma)^2)\right) \\
    &\pdv{u(x,y,t=0)}{t} = 0, \qquad u(x,y,t) = 0, \quad x,y \ \in\ \partial\Omega,\ t\in [0, 1],
\end{align}
where $u$ defines the field variable, $c$ the wave velocity, $x$ and $y$ the spatial coordinates, $t$ the temporal coordinates. $\alpha$, $\beta$ and  $\gamma$ are variables that parameterise the initial condition of the PDE setup. There exists an additional constraint to the PDE setup that initialises the velocity of the wave to 0. The system is bounded periodically within the mentioned domain.

The solution for the wave equation is obtained by deploying a spectral solver that uses a leapfrog method for time discretisation and a Chebyshev spectral method on a tensor product grid for spatial discretisation \citep{GOPAKUMAR2023100464}. The dataset is built by performing a Latin hypercube scan across the defined domain for the parameters $\alpha, \beta, \gamma$,  which accounts for the amplitude and the location of the Gaussian peak, sampled differently for each simulation. We generate 2,500 simulation points, each one with its own initial condition and use 500 for training, 1,000 each for calibration and procuring the prediction sets. We train a 2D U-Net and an FNO to learn the evolution of wave dynamics. 

The physics of the equation, given by the various coefficients, is held constant across the dataset generation throughout, as given in  \cref{eq: wave}. Each data point, as in each simulation, is generated with a different initial condition as described above. The parameters of the initial conditions are sampled from within the domain as given in \cref{table: data_generation_wave}. Each simulation is run for 150-time iterations with a $\Delta t = 0.00667$ across a spatial domain spanning $[-1,1]^2$,  uniformly discretised into 33 spatial units in the x and y axes. Once the simulations are completed and the dataset is generated, we select the first 80 time instances of the evolution of each simulation to be used for training. 

\begin{table}[h!]
\caption{ Domain range and sampling strategies across the initial condition parameters for the 2D Wave Equation. A 2D Gaussian peak with a given amplitude and position within the domain is sampled using a Latin hypercube. } 
\label{table: data_generation_wave}
  \centering
  \begin{tabular}{lll}
  \toprule
  Parameter & Domain & Type \\
    \midrule
    Amplitude $(\alpha)$ & $[10, 50]$ & Continuous  \\
    X position $(\beta)$ & $[0.1, 0.5]$ & Continuous \\
    Y position $(\gamma)$ & $[0.1, 0.5]$ & Continuous \\
    \bottomrule
    \end{tabular}
\end{table}

\subsection{Model and Training}

\begin{wrapfigure}{r}{0.35\textwidth}
  \centering
    \includegraphics[width=\hsize]{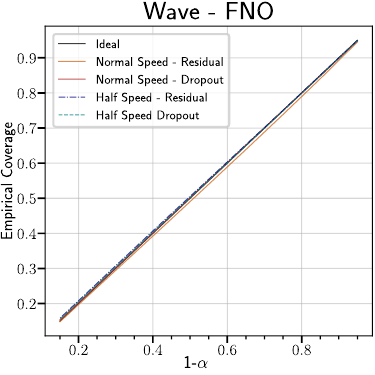}
  \caption{Coverage observed using AER and STD across datasets with normal wave speed (in-distribution) and half speed (out-distribution).}
\label{fig: fno_halfspeed}
\end{wrapfigure}

We train U-Nets and FNOs to map the spatio-temporal evolution of the field variables. For the U-Nets, the network takes in the first 20 time instances (\textit{T\_in}) to map the next 30 time instances (\textit{step}). The U-net performs a feed-forward mapping without any autoregressive roll-outs. For the FNO we deploy an auto-regressive structure that performs time rollouts, allowing us to map the initial time steps in a recursive manner up until the desired time instance (\textit{T\_out}). Each model autoregressively models the evolution of the field variable up until the $80^{th}$ time instance. The U-Net architecture can be found in \cref{table: unet_arch_wave} and the FNO in  \cref{table: fno_arch_wave}. For the case of STD, the architecture is modified with 2D dropout layers following each encoder and decoder of the U-Net and after each Fourier layer within the FNO. We employ a linear range normalisation scheme, placing the field values between -1 and 1. Each model is trained for up to 500 epochs using the Adam optimiser \citep{adam} with a step decaying learning rate. The learning rate is initially set to 0.005 and scheduled to decrease by half after every 100 epochs. The model was trained using a quantile loss in the case of CQR and an MSE loss in the other cases. 

\begin{figure}[!ht]
    \centering
    \includegraphics[scale=0.5]{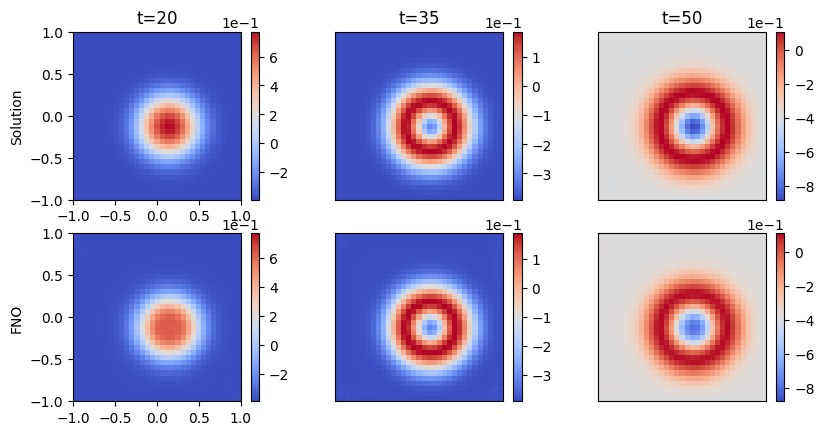}
    \caption{Waves: Temporal evolution of the field associated with the wave equation modelled using the numerical spectral solver (top of the figure) and that of the U-Net (bottom of the figure). The spatial domain is given in Cartesian geometry.}
    \label{fig: wave_comparison}
\end{figure}

\begin{figure}[!ht]
    \centering
    \includegraphics[scale=0.5]{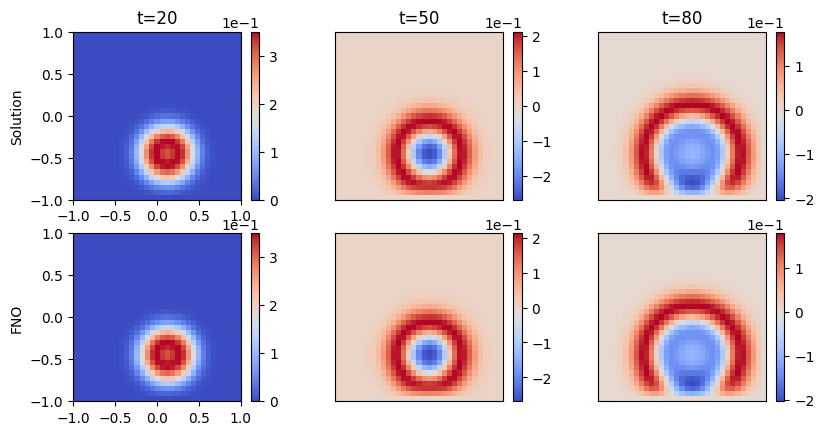}
    \caption{Waves: Temporal evolution of the field associated with the wave equation modelled using the numerical spectral solver (top of the figure) and that of the FNO (bottom of the figure). The spatial domain is given in Cartesian geometry.}
    \label{fig: wave_comparison_fno}
\end{figure}

\begin{table}[ht]
  \caption{Architecture of the 2D U-Net deployed for the 2D Wave Equation}
  \label{table: unet_arch_wave}
  \centering
  \begin{tabular}{lll}
    \toprule
    Part     & Layer     &  Output Shape \\
    \midrule
    Input & - & (50, 20, 33, 33) \\    
    Encoder 1 & \texttt{Conv2d/BatchNorm2d/Tanh} & (50, 32, 33, 33) \\
    Pool 1 & \texttt{MaxPool2d} & (50, 32, 33, 33)\\
    Encoder 2 & \texttt{Conv2d/BatchNorm2d/Tanh} & (50, 64, 16, 16) \\
    Pool 2 & \texttt{MaxPool2d} & (50, 64, 16, 16)\\
    Bottleneck & \texttt{Conv2d/BatchNorm2d/Tanh} & (50, 128, 8, 8) \\
    Decoder 2 & \texttt{ConvTranspose2D/Encoder 2} & (50, 64, 16, 16) \\
    Decoder 1 & \texttt{ConvTranspose2D/Encoder 1} & (50, 32, 33, 33) \\
    Rescale  & \texttt{Conv2D} & (50, 10, 33, 33) \\
    \bottomrule
  \end{tabular}
\end{table}

\begin{table}[h!]
\caption{\label{table: fno_arch_wave}Architecture of the Individual FNO deployed for modelling the Wave Equation} 
  \centering
  \begin{tabular}{lll}
    \toprule
    Part     & Layer     &  Output Shape \\
    \midrule
    Input & - & (50, 33, 33, 22) \\    
    Lifting & \texttt{Linear} & (50, 33, 33, 32) \\
    Fourier 1 & \texttt{Fourier2d/Conv2d/Add/GELU} & (50, 32, 33, 33)\\
    Fourier 2 & \texttt{Fourier2d/Conv2d/Add/GELU} & (50, 32, 33, 33)\\
    Fourier 3 & \texttt{Fourier2d/Conv2d/Add/GELU} & (50, 32  33, 33)\\
    Fourier 4 & \texttt{Fourier2d/Conv2d/Add/GELU} & (50, 32, 33, 33)\\
    Fourier 5 & \texttt{Fourier2d/Conv2d/Add/GELU} & (50, 32, 33, 33)\\
    Fourier 6 & \texttt{Fourier2d/Conv2d/Add/GELU} & (50, 32, 33, 33)\\
    Projection 1 & \texttt{Linear} & (50, 33, 33, 128) \\
    Projection 2 & \texttt{Linear} & (50, 33, 33, 10) \\
    \bottomrule
    \end{tabular}
\end{table}

\clearpage
\newpage
\section{Navier-Stokes Equations for Vorticity}
\label{appendix_ns}

\subsection{Physics}
The Navier-Stokes scenario that we are interested in modelling is taken from the exact formulation in \cite{li2021fourier}, where the viscosity of the incompressible fluid in 2D is expressed as:
\begin{align}
\label{eq: ns}
    \pdv{w}{t} + u \nabla w  &= \nu  \nabla^2 w + f, &\quad x \in (0,1) , \; y\in (0,1) , \; t\in (0,T)\\
    \nabla u &= 0, &\quad x \in (0,1) , \; y\in (0,1) , \; t\in (0,T)\\
    w &= w_0, &\quad x \in (0,1) , \; y\in (0,1) , \; t=0,
\end{align}
where $u$ is the velocity field and vorticity is the curl of the velocity field $w = \nabla \cross u$. The domain is split across the spatial domain characterised by $x,y$ and the temporal domain $t$. The initial vorticity is given by the field $w_0$. The forcing function is given by $f$ and is a function of the spatial domain in $x,y$. We utilise two datasets from \cite{li2021fourier} that are built by solving the above equations with viscosities $\nu = 1e-3$ and $\nu=1e-4$ under different initial vorticity distributions. For further information on the physics and the data generation, refer \cite{li2021fourier}. 

\subsection{Model and Training}

We train an FNO to map the spatio-temporal evolution of the vorticity, taking in the first 10 time instances (\textit{T\_in}) to the next 10 time instances (\textit{step}). For the case of the Navier-Stokes equations, we deploy a feed-forward mapping from the initial 10 time steps to the next 10 time steps. The architecture for the FNO can be found in  \cref{table: fno_arch_ns}. We deploy a Min-Max normalisation strategy, allowing the field values to lie between -1 and 1. Each model is trained for up to 500 epochs using the Adam optimiser \citep{adam} with a step decaying learning rate. The learning rate is initially set to 0.005 and scheduled to decrease by half after every 100 epochs. The model was trained using a relative LP loss. Considering the efficiency and simplicity, conformal prediction for the Navier-Stokes case was conducted using AER and STD as a nonconformity metric as given in the \cref{nonconformity scores}. 

\begin{table}[h!]
\caption{\label{table: fno_arch_ns}Architecture of the Individual FNO deployed for modelling the Navier-Stokes Equation} 
  \centering
  \begin{tabular}{lll}
    \toprule
    Part     & Layer     &  Output Shape \\
    \midrule
    Input & - & (20, 64, 64, 12) \\    
    Lifting & \texttt{Linear} & (50, 64, 64, 32) \\
    Fourier 1 & \texttt{Fourier2d/Conv2d/Add/GELU} & (20, 16, 64, 64)\\
    Fourier 2 & \texttt{Fourier2d/Conv2d/Add/GELU} & (20, 16, 64, 64)\\
    Fourier 3 & \texttt{Fourier2d/Conv2d/Add/GELU} & (20, 16  64, 64)\\
    Fourier 4 & \texttt{Fourier2d/Conv2d/Add/GELU} & (20, 16, 64, 64)\\
    Fourier 5 & \texttt{Fourier2d/Conv2d/Add/GELU} & (20, 16, 64, 64)\\
    Fourier 6 & \texttt{Fourier2d/Conv2d/Add/GELU} & (20, 16, 64, 64)\\

    Projection 1 & \texttt{Linear} & (20, 64, 64, 128) \\
    Projection 2 & \texttt{Linear} & (20, 64, 64, 10) \\
    \bottomrule
    \end{tabular}
\end{table}

\subsubsection{Prediction}

\begin{figure}[h!]
        \centering
        \begin{subfigure}[b]{0.75\textwidth}
            \centering 
            \includegraphics[width=\textwidth]{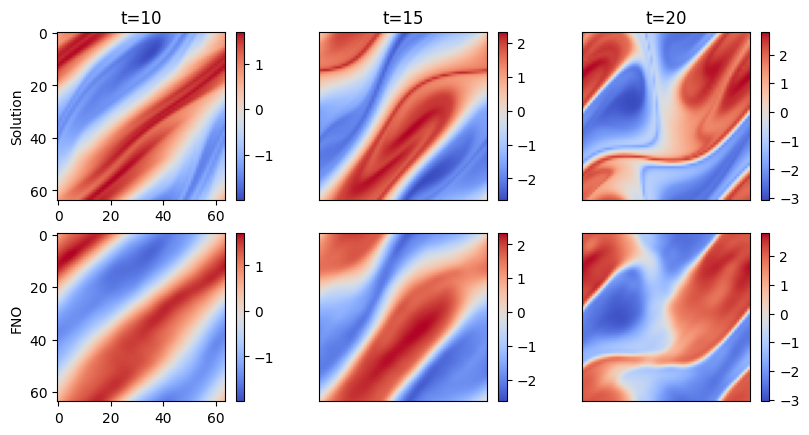}
            \label{fig: ns_1e-3}
            \caption{Training Distribution}
        \end{subfigure}
        \hfill
        \begin{subfigure}[b]{0.75\textwidth}
            \centering 
            \includegraphics[width=\textwidth]{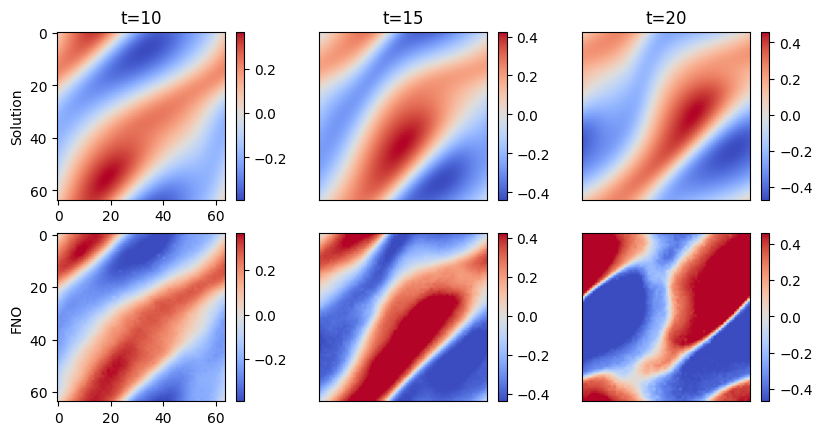}
            \label{fig: ns_1e-4}
            \caption{Out-of-distribution}
        \end{subfigure}
    \caption{Navier--Stokes: Temporal evolution of the vorticity associated with the Navier-Stokes equations. In (a), we compare the FNO performance against that of the numerical solver within the training distribution ($\nu=1e-3$). In (b), we demonstrate the performance of the same FNO on out-of-distribution data upon which we perform Conformal Prediction ($\nu=1e-4$) modelled using the numerical solver (top of the figure) and that of the FNO (bottom of the figure). The spatial domain is given in Cartesian geometry.}
    \label{fig: NS}
\end{figure}

\clearpage
\newpage
\section{Magnetohydrodynamics of Plasma Blobs}
\label{appendix_mhd}

\subsection{Physics}
The Reduced-MHD equations that we are interested in modelling can be described as:

\begin{equation}
      \frac{\partial\rho}{\partial t} =
	- \nabla \cdot \left( \rho \vec{v} \right)
	+ D \nabla^2 \rho 
 \label{CONTINUITY}
\end{equation}

\begin{equation}
    \rho \frac{\partial \vec{v}}{\partial t} =
    - \rho \vec{v} \cdot \nabla \vec{v}
    - \nabla p + \mu \nabla^2\vec{v} 
    \label{MOMENTUM}
\end{equation}

\begin{equation}
 \frac{\partial p}{\partial t} =
    - \vec{v}\cdot\nabla p - \gamma p \nabla \cdot \vec{v} + \kappa \nabla^2 T    
    \label{ENERGY}
\end{equation}

where $\rho$ is the density, $p$ the pressure, $T$ the temperature, and $\vec{v}$ the velocity. $D$ is the diffusion coefficient, $\mu$ the viscosity, and $\kappa$ the thermal conductivity. The ratio of specific heats $\gamma$ is taken to be that of a monatomic gas, $\frac{5}{3}$.

\Cref{CONTINUITY} depicts the continuity equation, modelling the evolution of density subject to diffusion, convection and the electrostatic potential.  \Cref{MOMENTUM} represents the conservation of momentum within the field. \Cref{ENERGY} models the conservation of energy, characterised by the pressure, temperature and velocity. 

Within each simulation, multiple-density blobs with varying positions, width and amplitude are initialised in a low-density background. In the absence of a plasma current to hold the density blob in place, the pressure gradient term in the momentum equation generates a buoyancy effect, causing the blob to move outwards. The system under consideration is characterised by a highly correlated multi-variable setting as given above. Within each simulation, we evolve the blobs to migrate radially outward until they reach the wall, where the Dirichlet boundary conditions engage to allow for convection and diffusion. Refer to \citep{Gopakumar_2024} for more detailed information about the setup. 

\begin{table}[h!]
\label{table: data_generation_mhd}
\caption{ Domain range and sampling strategies across the initial condition parameters.} 
  \centering
  \begin{tabular}{lll}
  \toprule
  Parameter & Distribution & Type \\
    \midrule
    Width & $U[0.02,0.1]$ & Continuous  \\
    Number of Blobs & $U[1,10]$ & Discrete \\
    R - Position of Blobs & $U[9.4,10.4]$ & Continuous \\
    Z - Position of Blobs & $U[-0.4,+0.4]$ & Continuous \\
    Amplitude of Density of Blobs & $U[0.5, 2.0]$ & Continuous \\
    Amplitude of Temperature of Blobs & $U[0.5, 3.0]$ & Continuous \\

    \bottomrule
    \end{tabular}
\end{table}

\subsection{Model and Training}

We train a multi-variable FNO to map the spatio-temporal evolution of the field variable, taking in the first 10 time instances (\textit{T\_in}) to the next 5 time instances (\textit{step}). For the case of the MHD equations, we deploy an auto-regressive structure that performs a time rollout, allowing us to map the initial time steps in a recursive manner up until the desired time instance (\textit{T\_out}). Each model autoregressively models the evolution of the field variable up until the $50^{th}$ time instance. The architecture for the multi-variable FNO can be found in  \cref{table: mv_fno_arch}. We deploy a two-fold normalisation strategy considering the nature of the dataset. The physical field information represented within the MHD cases is in different scales, with densities ranging from 0 to $1e20$ and temperatures ranging up to $1e6$. Since we are considering the gradual diffusion of an inhomogeneous density blob(s), the data distribution within the spatial domain is severely imbalanced. Taking these aspects of the training data into consideration, a physics normalisation is performed initially, where the field values are scaled down by dividing by the prominent field value. This is followed up by a linear range scaling, allowing the field values to lie between -1 and 1. Each model is trained for up to 500 epochs using the Adam optimiser \citep{adam} with a step decaying learning rate. The learning rate is initially set to 0.005 and scheduled to decrease by half after every 100 epochs. The model was trained using a relative LP loss. Considering the efficiency and simplicity, conformal prediction for the MHD case was only conducted using the AER as a nonconformity metric, as given in the  \cref{nonconformity scores}. 

\begin{table}[h!]
\caption{\label{table: mv_fno_arch}Architecture of the Multi-variable FNO deployed for modelling Reduced MHD.} 
  \centering
  \begin{tabular}{lll}
    \toprule
    Part     & Layer     &  Output Shape \\
    \midrule
    Input & - & (10, 3, 106, 106, 12) \\    
    Lifting & \texttt{Linear} & (10, 3, 106, 106, 32) \\
    Fourier 1 & \texttt{Fourier2d/Conv3d/Add/GELU} & (10, 3, 32, 106, 106)\\
    Fourier 2 & \texttt{Fourier2d/Conv3d/Add/GELU} & (10, 3, 32, 106, 106)\\
    Fourier 3 & \texttt{Fourier2d/Conv3d/Add/GELU} & (10, 3, 32, 106, 106)\\
    Fourier 4 & \texttt{Fourier2d/Conv3d/Add/GELU} & (10, 3, 32, 106, 106)\\
    Fourier 5 & \texttt{Fourier2d/Conv3d/Add/GELU} & (10, 3, 32, 106, 106)\\
    Fourier 6 & \texttt{Fourier2d/Conv3d/Add/GELU} & (10, 3, 32, 106, 106)\\

    Projection 1 & \texttt{Linear} & (10, 3, 106, 106, 128) \\
    Projection 2 & \texttt{Linear} & (10, 3, 106, 106, 5) \\

    \bottomrule
    \end{tabular}
\end{table}

\subsection{Prediction}

\begin{figure}[h!]
        \centering
        \begin{subfigure}[b]{0.8\textwidth}
            \centering 
            \includegraphics[width=\textwidth]{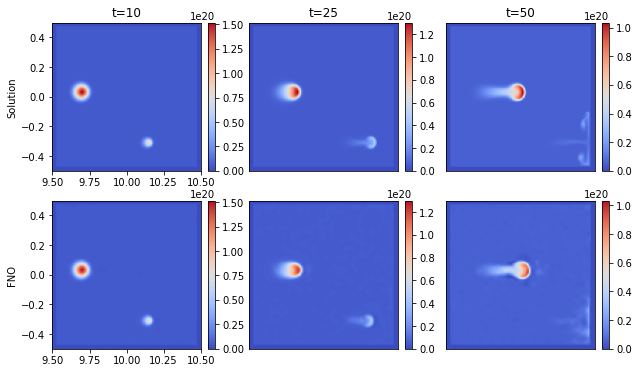}
            \label{fig: mhd_rho}
            \caption{Density}
        \end{subfigure}
        \hfill
        \begin{subfigure}[b]{0.8\textwidth}
            \centering 
            \includegraphics[width=\textwidth]{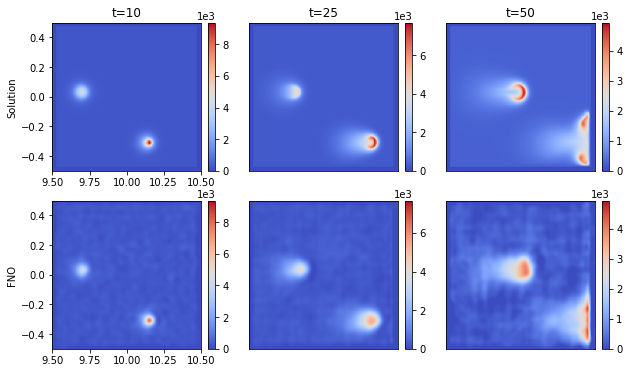}
            \label{fig: mhd_T}
            \caption{Temperature}
        \end{subfigure}
    \caption{Multiple Blobs: Temporal evolution of (a) the density and (b) the temperature variables describing the plasma evolution as obtained using the JOREK code (top of each image) and that of the multi-variable FNO (bottom of each figure). The spatial domain is given in toroidal geometry characterised by R in the x-axis and Z in the y-axis.}
    \label{fig: multi_mhd}
\end{figure}

\clearpage
\newpage
\section{Empirical Coverage of Foundation Physics Models}
\label{MPP_appendix}

\begin{figure}[!ht]
    \centering
    \begin{subfigure}{0.5\textwidth}
        \centering
        \includegraphics[width=\linewidth]{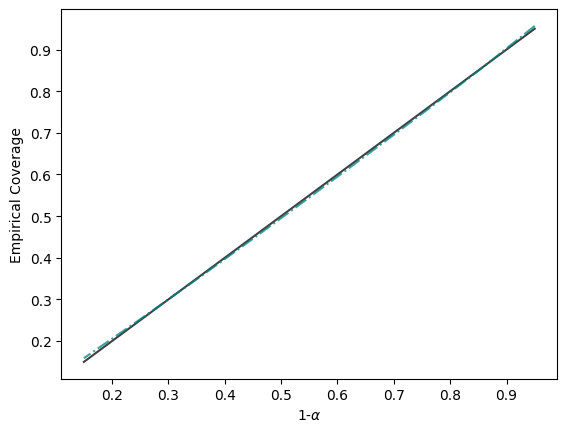}
        \caption{Pre-trained Model}
    \end{subfigure}%
    \begin{subfigure}{0.5\textwidth}
        \centering
        \includegraphics[width=\linewidth]{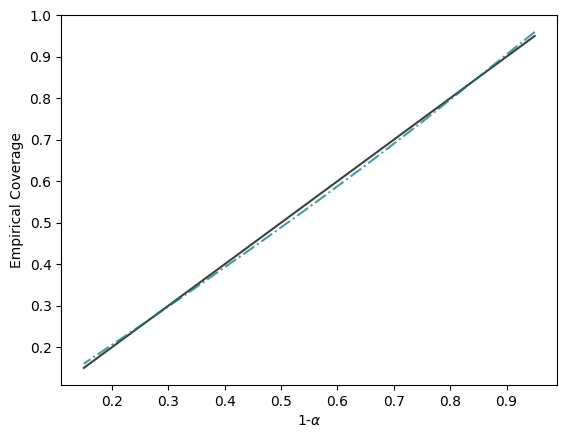}
        \caption{Fine-tuned Model}
    \end{subfigure}
    \caption{Empirical Coverage for Foundation Physics Models.}
    \label{fig:coverage_foundation}
\end{figure}

\clearpage
\newpage
\section{Impact of Calibration Size}
\label{calibration_size}
Across almost all the experiments in \cref{sec: experiments}, we use a calibration dataset of the size 1000 exchangeable simulations. The only exception is for the multi-variable FNO for MHD, for which we use 100 simulations, as there was limited data available. We chose the number 1000 as the baseline size of the calibration dataset since in \citep{gentle_introduction_CP} they demonstrate that choosing $n_{cal} =1000$ calibration points leads to a coverage that is typically between 0.88 and 0.92 for $\alpha = 0.1$. Since the size and nature of the calibration set are a source of finite sample variability, it requires analysis across each problem to which we deploy CP. 

Ideally, \cref{eq: coverage} holds for a calibration dataset of any size $n_{cal}$. The coverage guaranteed by CP conditionally on this calibration dataset is essentially a random quantity. Thus, depending on the choice of the calibration dataset, the coverage would fluctuate around $1-\alpha$. The distribution of the coverage as a function of the size of the calibration size is governed by a Beta distribution as given in \cref{eq: cov_dist}. 

\begin{figure}[h!]
    \centering
    \includegraphics[width=0.9\textwidth]{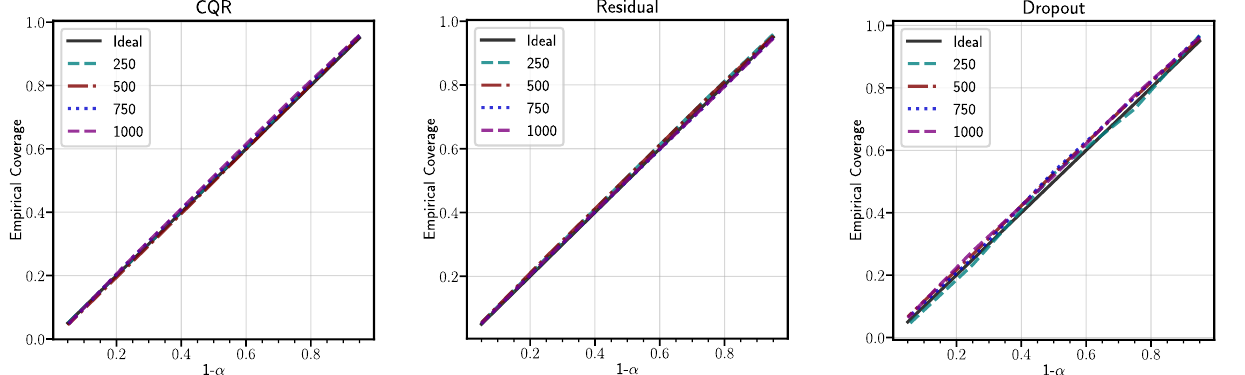}
    \caption{Impact of the size of the calibration dataset on the coverage obtained for Conformal Prediction across various nonconformity scores for the \textbf{Poisson Equation}. Irrespective of the chosen size of the calibration dataset, we obtain guaranteed coverage. }
    \label{fig: calib_size_poisson}
\end{figure}

\begin{figure}[h!]
    \centering
    \includegraphics[width=0.9\textwidth]{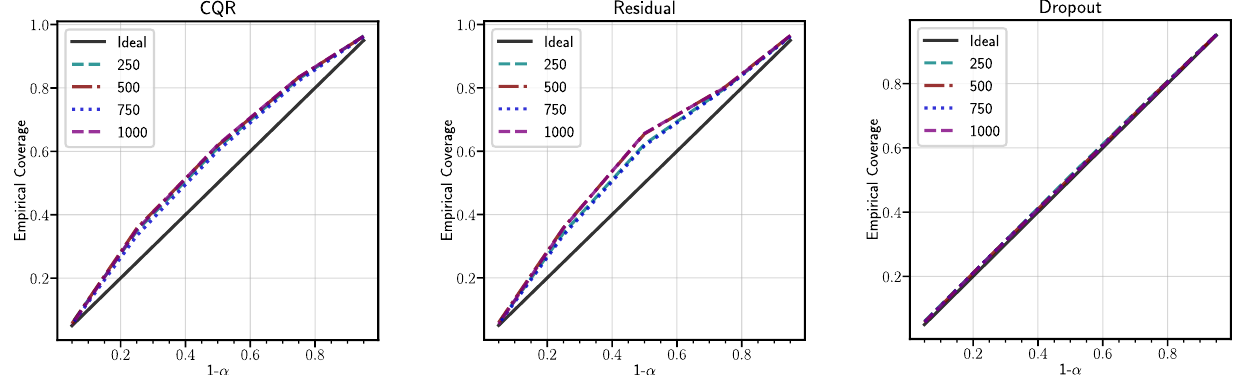}
    \caption{Impact of the size of the calibration dataset on the coverage obtained for Conformal Prediction across various nonconformity scores for the \textbf{Convection-Diffusion Equation}. Irrespective of the chosen size of the calibration dataset, we obtain guaranteed coverage; however, with larger $n_{cal}$ we obtain marginally better coverage. Irrespective of the chosen size of the calibration dataset, we obtain guaranteed coverage.  }
    \label{fig: calib_size_CD}
\end{figure}

\begin{figure}[h!]
    \centering
    \includegraphics[width=0.9\textwidth]{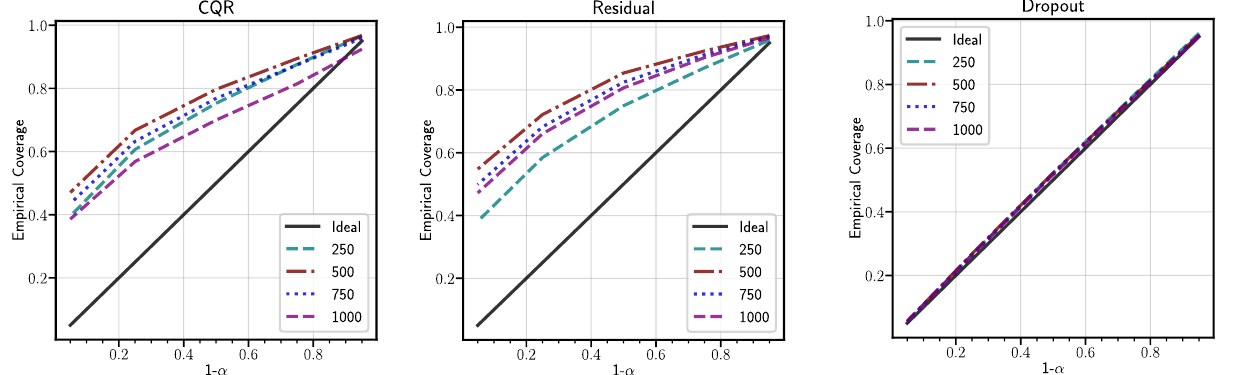}
    \caption{Impact of the size of the calibration dataset on the coverage obtained for Conformal Prediction across various nonconformity scores for the \textbf{Wave Equation modelled using U-Net}. Irrespective of the chosen size of the calibration dataset, we obtain guaranteed coverage. }
    \label{fig: calib_size_wave_unet}
\end{figure}

\begin{figure}[h!]
    \centering
    \includegraphics[width=0.9\textwidth]{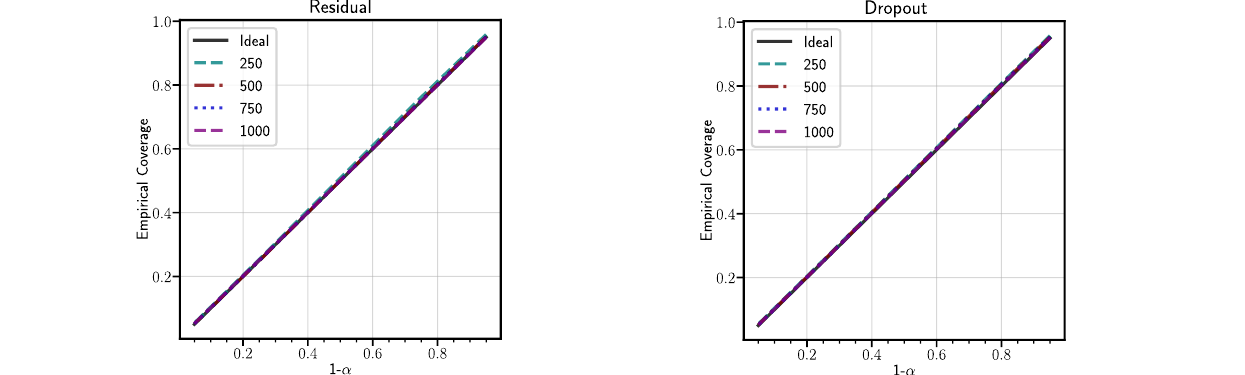}
    \caption{Impact of the size of the calibration dataset on the coverage obtained for Conformal Prediction across various nonconformity scores for the \textbf{Wave Equation modelled using FNO}. Irrespective of the chosen size of the calibration dataset, we obtain guaranteed coverage. }
    \label{fig: calib_size_wave_fno}
\end{figure}

We conduct an empirical study exploring the impact the size of the calibration dataset has on providing guaranteed coverage within our experiments. We iterate over $n_{cal} = 250, 500, 750, 1000$ for the Poisson (\cref{fig: calib_size_poisson}), Convection-Diffusion (\cref{fig: calib_size_CD}) and the Wave Equation (\cref{fig: calib_size_wave_unet}, \ref{fig: calib_size_wave_fno}). Though the coverage obtained is from a Beta distribution governed by $n_{cal}$ and changes with each sampling from that, our experiments are restricted to a single sample of $n_{cal}$ data points from that distribution. The study of the impact of the calibration dataset is done across all the various nonconformity scores as well. 


From \cref{fig: calib_size_poisson} to \cref{fig: calib_size_wave_fno}, we explore the impact of the size of the calibration dataset on the coverage obtained for the various nonconformity scores. We notice that across our experiments, we obtain guaranteed coverage, irrespective of the chosen nonconformity score or the size of the calibration dataset ($n_{cal}$). Though the size of the calibration dataset has no impact on the guarantee, as can be witnessed in the above figures, the tightness of the error bars is governed by the size of the calibration dataset.

\section{Neural Weather Prediction Models}
\label{sec:weather_appendix}

\subsection{Physics and Data}

\subsubsection{Limited Area Forecasting - MEPS Dataset}

The MetCoOp Ensemble Prediction System (MEPS) provides operational weather forecasts for the Nordic region \citep{meps}. Our dataset comprises historical MEPS forecasts from April 2021 to March 2023, initialised at 00 and 12 UTC daily with 5 ensemble members per initialisation ($\ sim$10 forecasts per day).

\paragraph{Spatial Configuration:} The domain uses a Lambert conformal conic projection, downsampled from 2.5 km to 10 km resolution for computational efficiency, yielding a grid of $N_x \times N_y = 238 \times 268$ nodes.

\paragraph{Temporal Configuration:} Forecasts extend 66 hours with 1-hour steps. Models predict at 3-hour intervals using the last two states as input, resulting in an effective 57-hour evaluation horizon (19 time steps).

\paragraph{Variables:} The dataset includes $N_\text{var} = 17$ variables: surface conditions (ground/sea level pressure, 2m temperature and humidity), radiation fluxes (net longwave/shortwave), lowest atmospheric level (~12.5m: temperature, humidity, wind components), pressure levels (temperature at 500/850 hPa, wind at 850 hPa, geopotential at 500/1000 hPa), and integrated water vapor column. All variables except solar radiation are instantaneous.

\paragraph{Data Splits:} Training uses months 1-15 (April 2021-June 2022), validation uses months 16-18, calibration uses September 2021 forecasts (from 2021-09-04 onwards), and testing uses September 2022 forecasts. Using the same month for calibration and testing minimises seasonal distributional shifts.

\subsubsection{Global Forecasting - ERA5 Dataset}

Global models are trained on ERA5 reanalysis data \citep{era5} at 1.5° resolution ($N_x \times N_y = 240 \times 121$ nodes). Forecasts extend 10 days with 6-hour steps ($T = 40$ time steps). The full dataset contains 83 variables (5 surface variables and 6 atmospheric variables at 13 pressure levels). For computational efficiency, CP experiments use $N_\text{var} = 11$ variables: all 5 surface variables and the 6 atmospheric variables at 700 hPa. Training uses multiple years of historical data, calibration uses the full year 2018, and testing uses 2019.

\subsection{Model Architecture and Training}

\subsubsection{Graph Neural Network Architecture}

The \lammodel architecture \citep{neural_lam} employs a hierarchical GNN with an encode-process-decode structure. Grid nodes represent the spatial discretisation, while mesh nodes form a coarser multi-resolution hierarchy ($L=4$ levels for a limited area, $L=8$ for global). Encoding maps grid states to the mesh, processing performs message passing across mesh levels (4 or 8 layers), and decoding projects back to the grid. All representations use 64-dimensional embeddings (limited area) or 256-dimensional embeddings (global).

\subsubsection{Training Procedure}

Models undergo two-stage training: (1) single-step prediction ($X^{t+1}$ from $X^{t-1:t}$), then (2) rollout fine-tuning with 4-step autoregressive sequences. Two model variants are trained:

\textbf{\lammodelmse:} Uses weighted MSE loss for deterministic predictions:
\begin{equation}
    \mathcal{L}_{\text{MSE}} = \frac{1}{N_\text{rollout}} \sum_{t} \frac{1}{|G|} \sum_{v \in G} \sum_{i=1}^{N_\text{var}} \lambda_i \omega_i \left(\hat{X}^t_{v,i} - X^t_{v,i}\right)^2
\end{equation}
where $\lambda_i$ is the inverse variance of time differences, $\omega_i$ weights the vertical level, and $G$ is the set of non-boundary grid nodes.

\textbf{\lammodelnll:} Uses NLL loss with diagonal Gaussians, outputting mean $\mu$ and standard deviation $\sigma$:
\begin{equation}
    \mathcal{L}_{\text{NLL}} = \frac{1}{N_\text{rollout}} \sum_{t} \frac{1}{|G|} \sum_{v \in G} \sum_{i=1}^{N_\text{var}} \left[\log \sigma^t_{v,i} + \frac{(X^t_{v,i} - \mu^t_{v,i})^2}{2(\sigma^t_{v,i})^2}\right]
\end{equation}

Training uses AdamW (learning rate 0.001), batch size 8 (limited area), 500 epochs for single-step and 200 for rollout fine-tuning, requiring 3-4 days on an NVIDIA A100 GPU (80 GB). All variables are normalised to zero mean and unit variance using training statistics.

\subsection{Inference and Autoregressive Forecasting}

Given initial states $X^{-1:0}$ and forcing inputs $F^{1:T}$, models autoregressively predict $X^{t+1} = \hat{f}(X^{t-1:t}, F^{t+1})$. Limited area models use lateral boundary forcing: predictions within a 10-grid-cell boundary are replaced with ground truth at each step. Forcing inputs include solar radiation at the top-of-atmosphere, diurnal and annual cycle encodings (sine/cosine transforms), and open water fraction. A 57-hour limited area forecast requires only 1.5 seconds on a single A100 GPU.

\subsection{Conformal Prediction Application}

\subsubsection{Nonconformity Scores}

\textbf{For \lammodelmse (Absolute Error Residual):}
\begin{equation}
    s(X, Y) = |Y - \hat{f}(X)|, \quad \mathbb{C}^\alpha(X) = [\hat{f}(X) - \hat{q}, \hat{f}(X) + \hat{q}]
\end{equation}

\textbf{For \lammodelnll (Standard Deviation):}
\begin{equation}
    s(X, Y) = \frac{|Y - \mu(X)|}{\sigma(X)}, \quad \mathbb{C}^\alpha(X) = [\mu(X) - \hat{q} \cdot \sigma(X), \mu(X) + \hat{q} \cdot \sigma(X)]
\end{equation}

This calibrates the model's potentially miscalibrated uncertainty estimates.

\subsubsection{Calibration Procedure}

CP is performed independently for each spatio-temporal cell. Limited area models calibrate $19 \times 238 \times 268 \times 17 = 20{,}602{,}232$ cells using ~270 September 2021 forecasts; global models use ~730 forecasts from 2018. For each cell, the $(1-\alpha)$-quantile is computed:
\begin{equation}
    \hat{q} = F^{-1}_{\hat{s}}\left(\frac{\lceil(n+1)(1-\alpha)\rceil}{n}\right)
\end{equation}

Calibration is computationally efficient: 229-310 seconds for limited area models and 366-401 seconds for global models, including score computation, quantile calculation, and coverage validation.

\subsection{Exchangeability Considerations}

\subsubsection{Weather Forecasting as an Initial Boundary Value Problem}

Applying CP to time series typically violates exchangeability due to temporal dependencies. However, treating each forecast as an independent Initial Boundary Value Problem (IBVP) preserves exchangeability: each forecast is fully determined by its initial conditions $X^{-1:0}$ and forcing $F^{1:T}$, making forecasts initialised at sufficiently separated times independent realisations from the atmospheric state space. This holds even when forecasts cover overlapping time periods, since each prediction conditions only on its own initial state.

CP is applied to the entire spatio-temporal output tensor $Y \in \mathbb{R}^{T_\text{out} \times N_x \times N_y \times N_\text{var}}$ simultaneously, preserving the physical forecast structure.

\subsubsection{Calibration-to-Test Exchangeability}

The coverage guarantee requires test forecasts to be exchangeable with calibration forecasts. For limited area models, this assumes September 2022 patterns are exchangeable with September 2021 patterns; for global models, 2019 states are exchangeable with 2018 states. These assumptions are reasonable given year-to-year climate consistency, validated by the excellent empirical coverage observed. Potential violations include anomalous events not represented in calibration data, systematic climate shifts, or changes in model distribution, warranting periodic recalibration with recent data.

\begin{figure}[tbp]
    \centering
    \begin{subfigure}{0.5\textwidth}
        \centering
        \includegraphics[width=\linewidth]{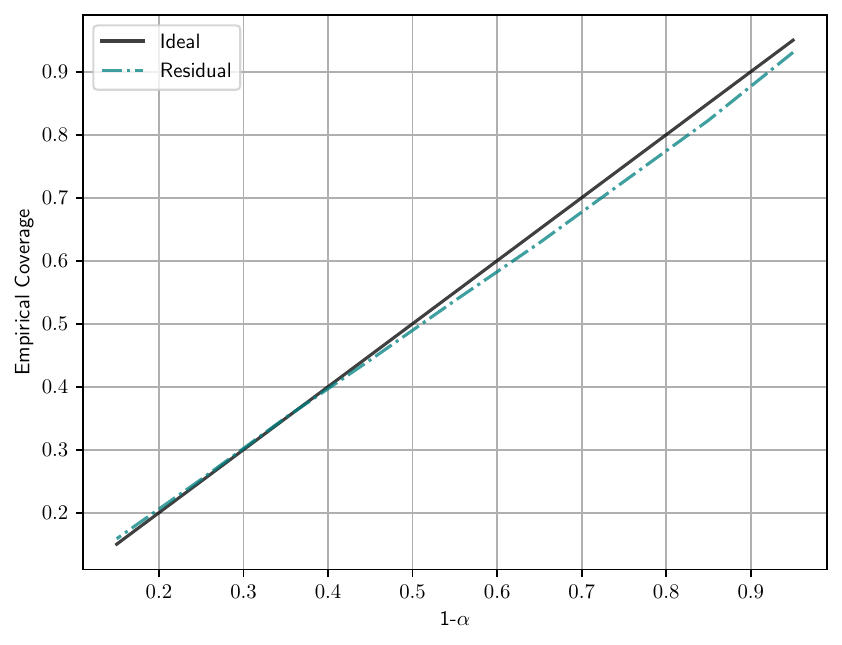}
        \caption{\lammodelmse}
    \end{subfigure}%
    \begin{subfigure}{0.5\textwidth}
        \centering
        \includegraphics[width=\linewidth]{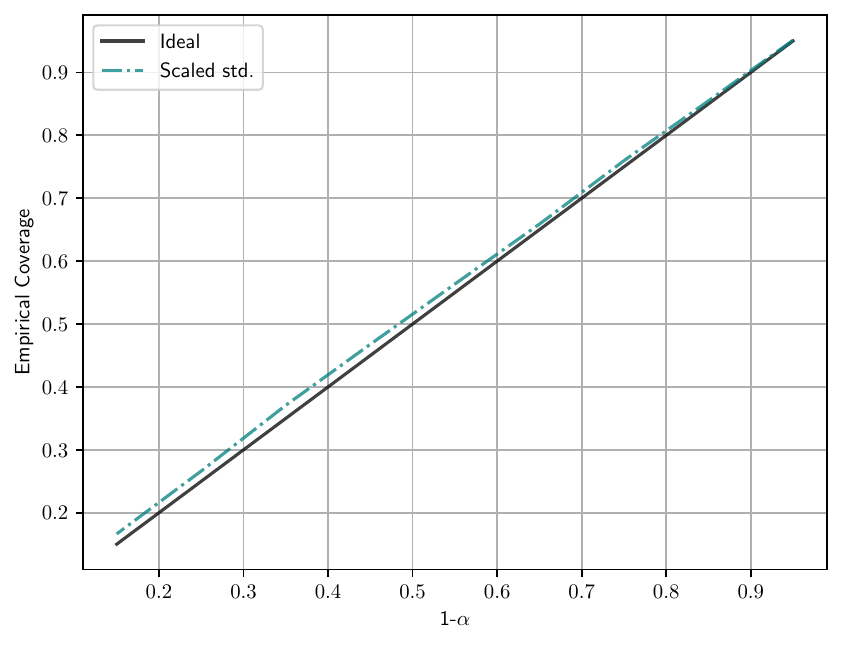}
        \caption{\lammodelnll}
    \end{subfigure}
    \caption{Empirical Coverage for limited area weather forecasting models, when calibrated on September and evaluating coverage on January.}
    \label{fig:neurwp_ood_emp_cov}
\end{figure}

\begin{figure}[tbp] 
    \centering
    \begin{subfigure}{\textwidth}
        \centering
        \includegraphics[width=\linewidth]{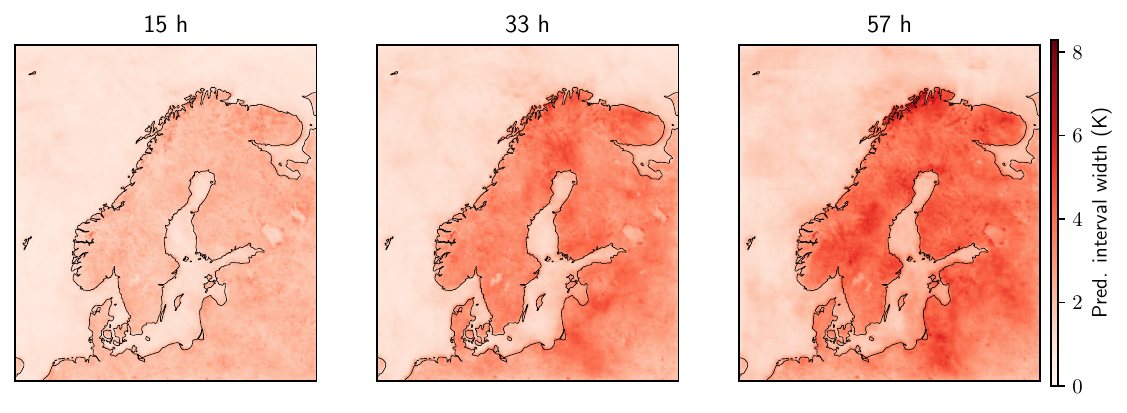}
        \caption{\lammodelmse}
    \end{subfigure}
    \begin{subfigure}{\textwidth}
        \centering
        \includegraphics[width=\linewidth]{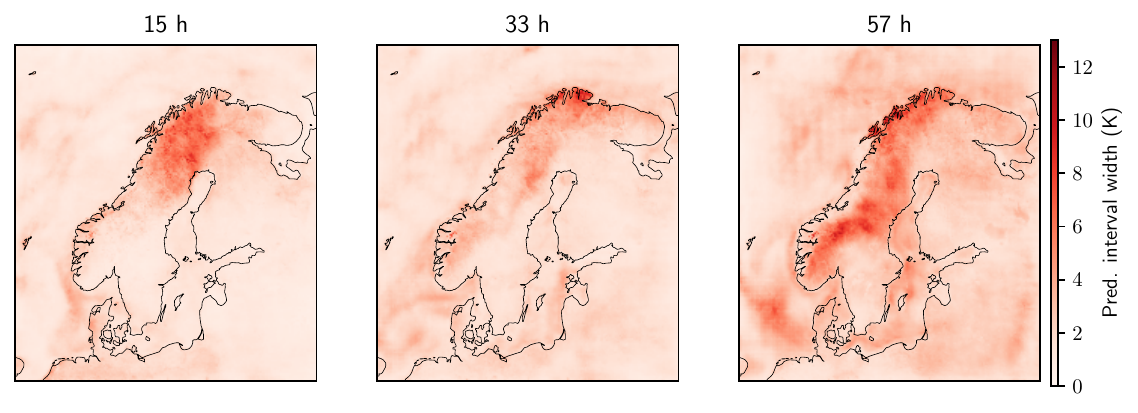}
        \caption{\lammodelnll}
    \end{subfigure}
    \caption{Width of predictive intervals in an example January forecast from the distributional shift setting. Here $\alpha = 0.05$ and we show results for temperature 2 m above ground (\texttt{2t}).
    }
    \label{fig:neurwp_ood_iwidth_t_2}
\end{figure}

To investigate the setting where there is a distributional shift between the calibration and test datasets, we consider a setting where we shift the month of the year in the weather data.
For this experiment, we calibrate the limited area weather model using predictions for September 2021, and evaluate on January 2023.
The weather in the Nordic region being modelled exhibits strong seasonal dependencies, with the winter month of January being much colder than the September month used for calibration.
Note that while we only use September for calibration, the model has been trained with data from all seasons of the year.
Empirical coverage is shown in \cref{fig:neurwp_ood_emp_cov}, and in \cref{fig:neurwp_ood_iwidth_t_2} we plot the width of predictive intervals for 2 m temperature predictions.
We note that, while the coverage shows larger deviations than in the setting where both calibration and evaluation were performed for the September data, the method overall still provides useful uncertainty estimates. This may be attributed to the reason that the physics regime characterising the weather evolution does not change across the months, but only the initial conditions to which the model is exposed. 

\subsection{Additional Results}

\begin{figure}
    \centering
        \begin{subfigure}{0.5\textwidth}
        \centering
        \includegraphics[width=\linewidth]{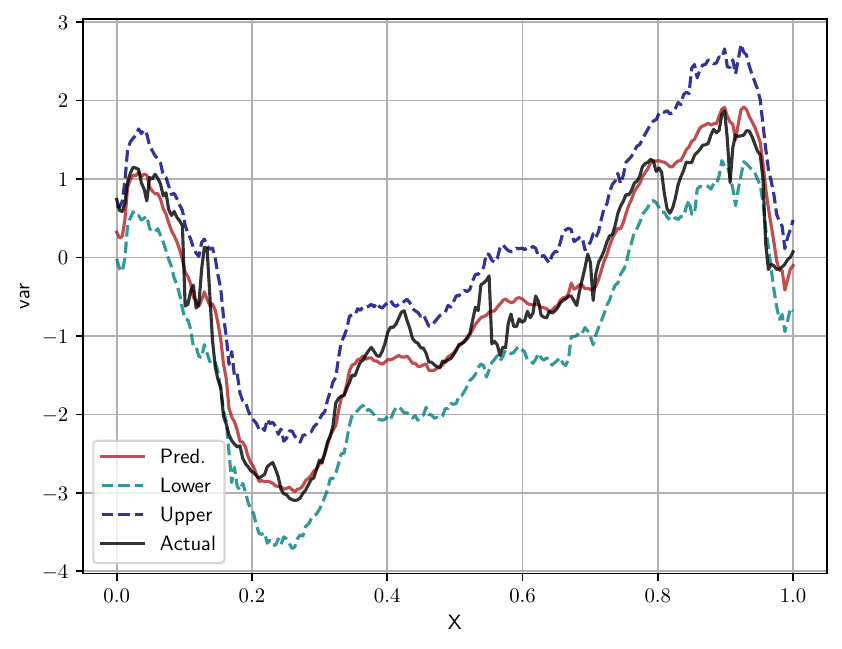}
        \caption{\lammodelmse, $\alpha = 0.05$}
        \end{subfigure}%
    \begin{subfigure}{0.5\textwidth}
        \centering
        \includegraphics[width=\linewidth]{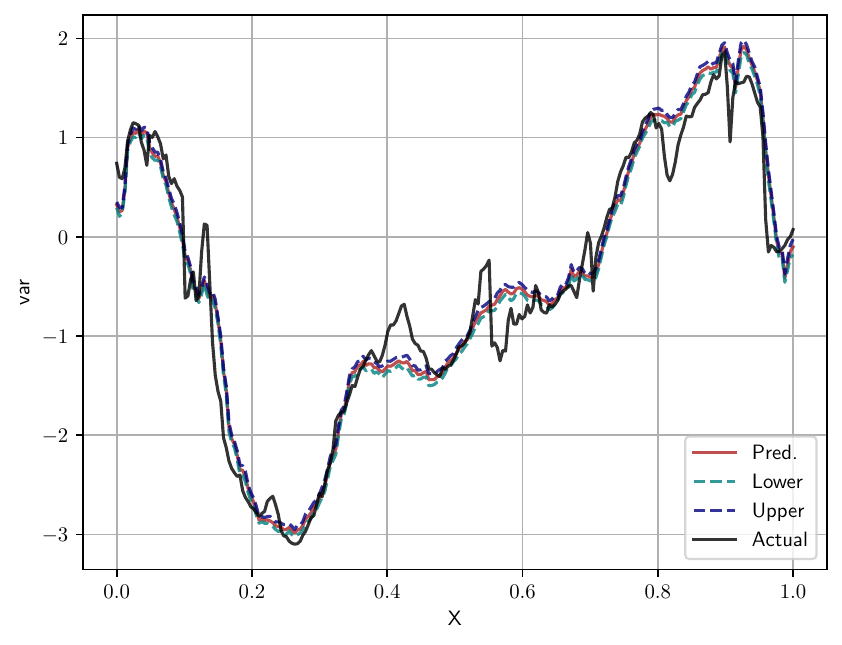}
        \caption{\lammodelmse, $\alpha = 0.85$}
    \end{subfigure}
    \begin{subfigure}{0.5\textwidth}
        \centering
        \includegraphics[width=\linewidth]{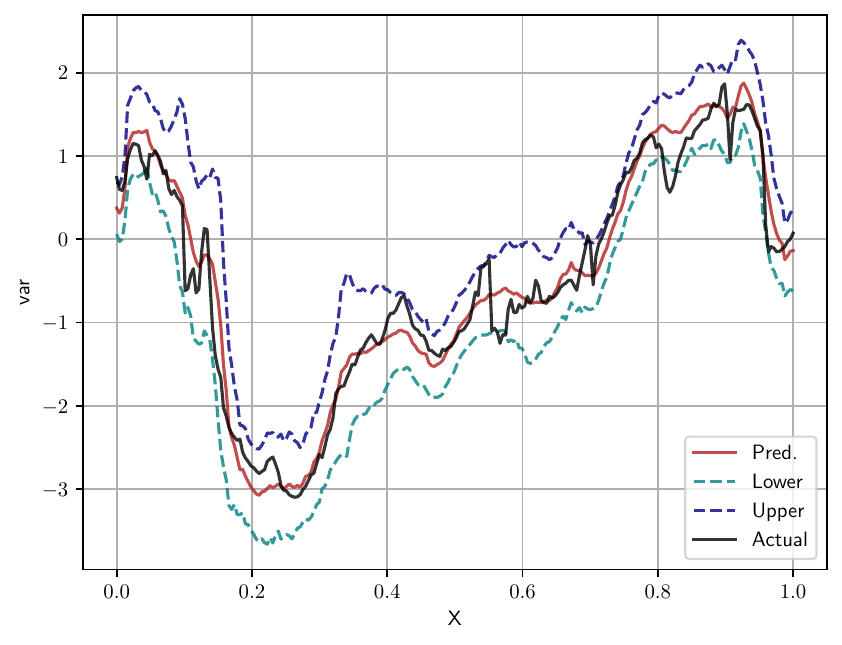}
        \caption{\lammodelnll, $\alpha = 0.05$}
        \end{subfigure}%
    \begin{subfigure}{0.5\textwidth}
        \centering
        \includegraphics[width=\linewidth]{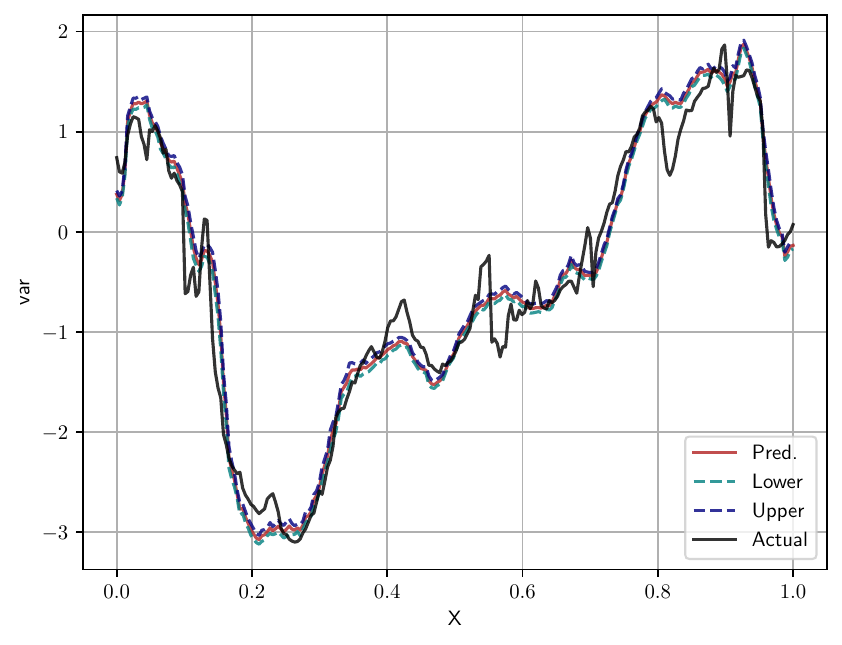}
        \caption{\lammodelnll, $\alpha = 0.85$}
    \end{subfigure}%
    \caption{Slice plots across the $x$-axis showing u-wind component predictions. Panels (a)-(d) show ground truth, prediction, and CP-derived upper/lower bounds for \lammodelmse and \lammodelnll at 95\% ($\alpha=0.05$) and 15\% ($\alpha=0.85$) coverage.}
    \label{fig:neurwp_slices}
\end{figure}

\Cref{fig:neurwp_slices} shows spatial slices illustrating error bars at different coverage levels. The \lammodelnll model produces tighter, spatially variable error bars due to input-dependent uncertainties, while \lammodelmse produces constant-width bars at each lead time. Both achieve target coverage, validating the CP framework.

\begin{figure}
    \centering
    \begin{subfigure}{0.5\textwidth}
        \centering
        \includegraphics[width=\linewidth]{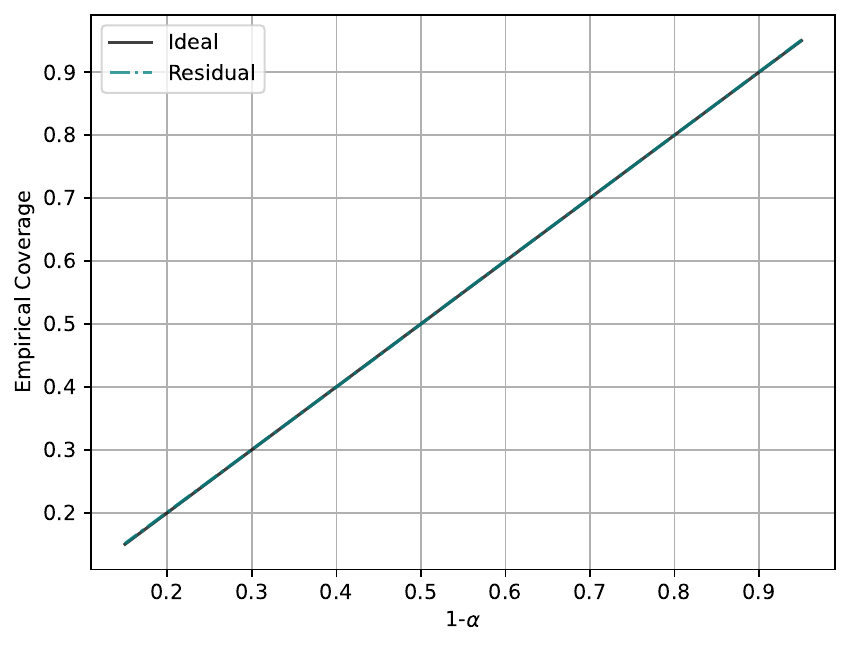}
        \caption{\lammodelmse, \texttt{t700}}
    \end{subfigure}%
    \begin{subfigure}{0.5\textwidth}
        \centering
        \includegraphics[width=\linewidth]{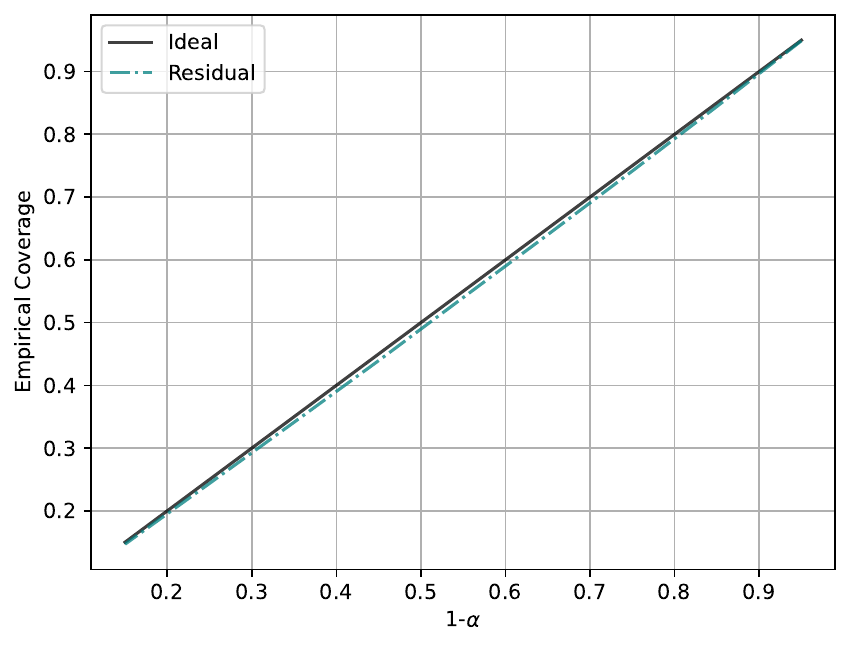}
        \caption{\lammodelmse, \texttt{q700}}
    \end{subfigure}
    \begin{subfigure}{0.5\textwidth}
        \centering
        \includegraphics[width=\linewidth]{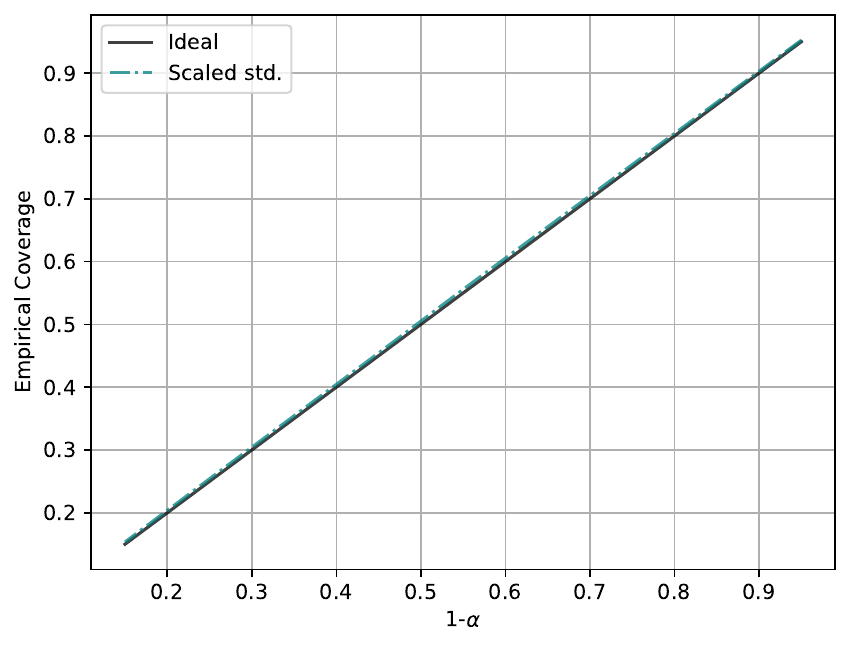}
        \caption{\lammodelnll, \texttt{t700}}
    \end{subfigure}%
    \begin{subfigure}{0.5\textwidth}
        \centering
        \includegraphics[width=\linewidth]{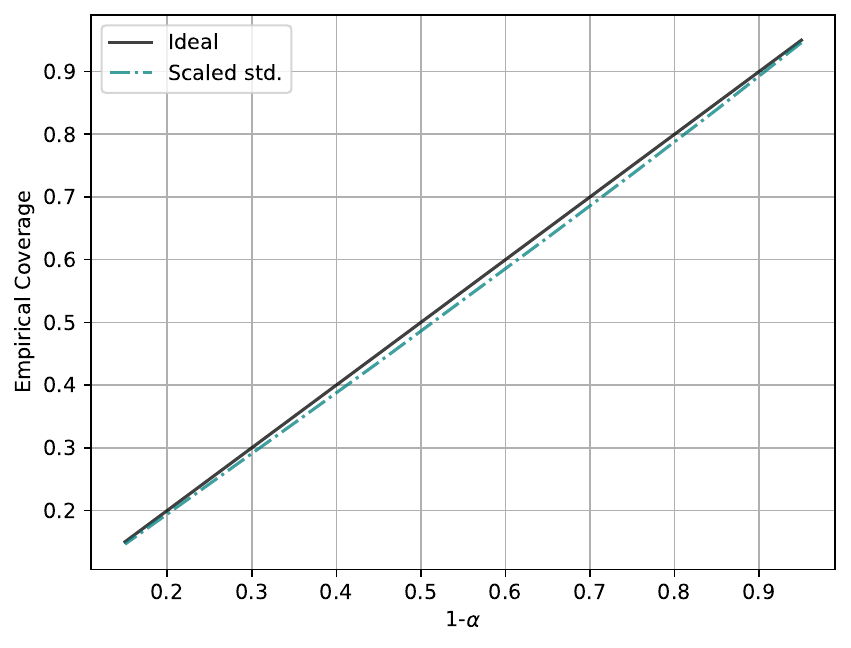}
        \caption{\lammodelnll, \texttt{q700}}
    \end{subfigure}%
    \caption{Empirical coverage for global weather forecasting models showing temperature (\texttt{t700}) and geopotential (\texttt{q700}) at 700 hPa. Both models achieve near-ideal coverage across all $\alpha$ levels.}
    \label{fig:global_neurwp_emp_cov}
\end{figure}

\Cref{fig:global_neurwp_emp_cov} demonstrates empirical coverage for global forecasting (temperature and geopotential at 700 hPa). Both model types closely follow the ideal diagonal, confirming guaranteed marginal coverage across $1{,}161{,}600$ spatial-temporal points per variable.

\subsubsection{Deterministic vs. Probabilistic Models}

\textbf{Deterministic Models (AER):} Error bars have constant width at each lead time, determined purely by calibration performance. Computationally efficient but potentially overconservative in predictable regions.

\textbf{Probabilistic Models (STD):} Error bars adapt to local forecast uncertainty, providing tighter bounds in high-confidence regions. Requires NLL training but benefits from CP calibration of potentially miscalibrated model uncertainties.

Both approaches achieve guaranteed coverage, with probabilistic models generally providing more informative bounds at the cost of additional complexity.

\newpage
\section{Camera Diagnostic on a Tokamak}
\label{appendix_camera}

\subsection{Camera Configuration and Data Processing}

\subsubsection{Camera Specifications}
The Mega-Ampere Spherical Tokamak (MAST) was equipped with Photron fast-visible cameras capturing plasma evolution at an average temporal resolution of 1.2 ms. We focus on the RBB camera that prioritises a central solenoid view of the plasma. It provides a panoramic view of the reactor, showing the poloidal cross-sectional layout on both sides of the central solenoid with spatial resolution of $448 \times 640$ pixels.

This camera is tuned to visible wavelengths and primarily captures Balmer D$_\alpha$ light emitted from the plasma edge, enabling visualisation of complex MHD dynamics, including edge-localised modes (ELMs).

\subsubsection{Data Selection Criteria}
Shot selection for training and testing followed stringent criteria to ensure data quality and consistency:

\begin{enumerate}
    \item \textbf{Plasma Presence:} Only shots containing actual confined plasma were selected, excluding commissioning pulses and dummy shots.
    \item \textbf{Temporal Duration:} Shots must contain more than 100 time steps to capture sufficient plasma evolution dynamics.
    \item \textbf{Camera Calibration Consistency:} All selected shots must share identical camera calibration parameters. Using CalCam \citep{scott_silburn_2022_6891504} combined with 3D CAD models of MAST, the domain range covered by each camera calibration was determined. The upper and lower limits of the domain range were mapped onto uniform grids in both $R$ and $Z$ axes.
    \item \textbf{Spatial Resolution Uniformity:} All camera images must have identical spatial resolution to maintain consistency in the training dataset.
\end{enumerate}

55 shots from the range 30,250--30,431 in the M9 campaign were selected (50 for training, 5 for testing). The diversity of plasma scenarios in the training data is illustrated in Figure 19 in \citet{Gopakumar_2024}, which shows the temporal evolution of plasma current and total heating power across the selected shots.

\subsection{Training Regime: Sequential Time Window Approach}

\subsubsection{Rationale for Non-Autoregressive Architecture}
Unlike Markovian simulations, camera data presents fundamentally different characteristics that necessitate an alternative training approach:

\begin{itemize}
    \item \textbf{Non-Markovian Dynamics:} The plasma evolution captured by cameras represents a non-Markovian process with inherent noise and partial system information.
    \item \textbf{Absence of Causal Information:} The dataset does not include explicit information about control inputs (coil currents, heating parameters) that influence plasma evolution.
    \item \textbf{Continuous Data Generation:} Unlike simulations, where initial conditions fully determine evolution, experimental data is continuously generated during the plasma shot, providing new information at each time step.
    \item \textbf{Error Accumulation:} Autoregressive rollout would lead to rapid error accumulation over longer time horizons, limiting prediction capability.
\end{itemize}

\subsubsection{Sequential Window Mapping}
Instead of autoregressive prediction, we employ a sliding time window approach. The FNO maps a fixed-length input sequence to a fixed-length output sequence:

\begin{equation}
\tilde{Y}_{t:t+\text{step}} = \mathcal{F}_{\theta}(X_{t-T_{\text{in}}:t}, \mathcal{G})
\end{equation}

where:
\begin{itemize}
    \item $X_{t-T_{\text{in}}:t} \in \mathbb{R}^{T_{\text{in}} \times N_x \times N_y}$ represents the input sequence of camera frames
    \item $\tilde{Y}_{t:t+\text{step}} \in \mathbb{R}^{\text{step} \times N_x \times N_y}$ represents the predicted future frames
    \item $\mathcal{G} \in \mathbb{R}^{N_x \times N_y \times 2}$ represents the spatial grid discretization
    \item $\mathcal{F}_{\theta}$ denotes the FNO with parameters $\theta$
\end{itemize}

For our experiments, we set $T_{\text{in}} = 10$ and $\text{step} = 10$, allowing the FNO to predict 12~ms of plasma evolution given the previous 12~ms of camera data.

\subsubsection{Training Data Construction}
Each plasma shot containing approximately 200 frames is converted into multiple overlapping input-output pairs through a sliding window mechanism with a stride of 1:

\begin{align}
\text{Pair}_1 &: X_{[1:10]} \rightarrow Y_{[11:20]} \\
\text{Pair}_2 &: X_{[2:11]} \rightarrow Y_{[12:21]} \\
&\vdots \\
\text{Pair}_k &: X_{[k:k+9]} \rightarrow Y_{[k+10:k+19]}
\end{align}

This strategy yields approximately 180 training pairs per shot, significantly increasing the effective dataset size. The FNO learns to predict plasma evolution from any intermediate state within the operational range, rather than solely from initial conditions.

\subsection{Real-Time Deployment Strategy}

\subsubsection{Inference Pipeline}
For real-time operation, the trained FNO is deployed with the following pipeline:

\begin{enumerate}
    \item \textbf{Initial Phase:} Collect the first $T_{\text{in}}=10$ camera frames at plasma startup
    \item \textbf{Prediction:} Generate predictions for the next $\text{step}=10$ frames using the FNO
    \item \textbf{Update:} As new frames become available from the camera, slide the input window forward by one frame
    \item \textbf{Iteration:} Repeat prediction with the updated window throughout the plasma shot duration
\end{enumerate}

Since camera data is acquired at 1.2~ms intervals and the FNO requires only 6~ms for inference, predictions can be generated faster than real-time, providing a 12~ms forecast horizon with 6~ms computational overhead.

\begin{figure}
    \centering
    \includegraphics[width=0.75\linewidth]{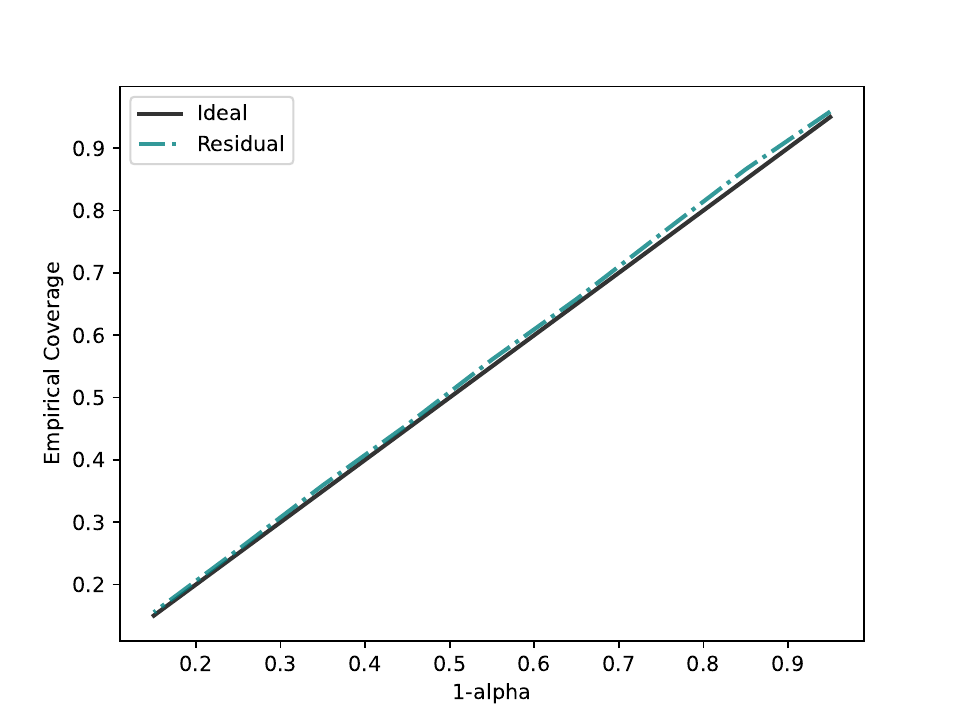}
    \caption{Empirical Coverage guaranteed by applying CP over the FNO forecasting the plasma evolution.}
    \label{fig:camera_coverage}
\end{figure}

\subsection{Exchangeability Considerations for Time Series Data}

\subsubsection{Theoretical Framework}
Traditional conformal prediction requires exchangeability between calibration and test data \citep{vovk2005algorithmic}. For time-series data, this assumption is typically violated. However, by treating each plasma shot as an Initial Boundary Value Problem (IBVP), we maintain exchangeability through the following reasoning:

\begin{itemize}
    \item \textbf{IBVP Structure:} Each input-output pair $(X_i, Y_i)$ represents a self-contained prediction task where the output is fully determined by the input sequence and boundary conditions, if they are there (forcing terms from control systems).
    \item \textbf{Temporal Independence:} Plasma profile predictions starting at different times (e.g., 30 seconds into the shot or 60 seconds into the shot) are independent, conditioned on their respective initial states.
    \item \textbf{Distributional Sampling:} The calibration dataset represents samples from a large distribution characterising the entire operational range of the tokamak under consideration.
\end{itemize}

\subsubsection{Limitations and Assumptions}
The exchangeability assumption relies on several critical conditions:

\begin{equation}
\mathbb{P}(\text{Exchangeable}) = \begin{cases}
\text{High} & \text{if shots share similar plasma profiles} \\
\text{Low} & \text{if shots have dissimilar characteristics}
\end{cases}
\end{equation}

As demonstrated in Figure~\ref{fig:camera_coverage_diff_shots}, when the prediction shot has significantly different plasma discharge profiles from the calibration shots, exchangeability is violated and coverage degrades. This necessitates calibration datasets that adequately span the operational space of interest.

\end{document}